# Deep learning for pedestrians: backpropagation in CNNs

Laurent Boué

SAP Labs

**Abstract**

The goal of this document is to provide a pedagogical introduction to the main concepts underpinning the training of deep neural networks using gradient descent; a process known as backpropagation. Although we focus on a very influential class of architectures called "convolutional neural networks" (CNNs) the approach is generic and useful to the machine learning community as a whole. Motivated by the observation that derivations of backpropagation are often obscured by clumsy index-heavy narratives that appear somewhat mathemagical, we aim to offer a conceptually clear, vectorized description that articulates well the higher level logic. Following the principle of "writing is nature's way of letting you know how sloppy your thinking is", we try to make the calculations meticulous, self-contained and yet as intuitive as possible. Taking nothing for granted, ample illustrations serve as visual guides and an extensive bibliography is provided for further explorations.

For the sake of clarity, long mathematical derivations and visualizations have been broken up into a short "summarized view" and a longer "detailed view". Detailed views are encoded into the PDF as optional content groups that become visible by clicking on the yellow symbol +. In addition, some figures contain animations designed to illustrate important concepts in a more engaging style. For these reasons, we advise to download the document locally and open it using **Adobe Acrobat Reader**. Other viewers were not tested and may not render the detailed views, animations correctly. For completeness, the overall structure of the paper is summarized in a table of contents.

## 1 Supervised machine learning from 20,000 feet...

The general workflow of training supervised machine learning models follows a well-oiled iterative procedure which we illustrate in the context of **convolutional neural networks** (CNNs) for image classification. This field has undergone such rapid development in the last few years that it is sometimes used as the official "success story" of deep learning. Indeed, supervised image classification (and related topics such as object detection, instance segmentation...) is nowadays considered as a kind of commodity software that made its way into countless industrial applications and, consequently, it is worthwile to become familiar with how CNNs work under the hood. Broadly speaking, deep learning models have a highly modular structure where so-called "layers" play the role of elementary building blocks. Although different kinds of layers serve different purposes, it turns out that CNNs are constructed from very generic layers that also appear in a wide range of very diverse neural network architectures. This means that the results presented here are effortlessly portable and remain useful in vast and far-reaching applications of deep learning that go beyond CNNs and even supervised techniques.

As an introduction, let us go over the high-level description of the iterative loop of training a machine learning model. This procedure is illustrated graphically in Fig.1 and the rest of this section is dedicated to a brief review of machine learning basics.

1. The starting point consists in gathering a large set of “training” data along with their accompanying “ground-truth” labels:

   - Each sample $s$ of data can be expressed as a $f$-dimensional feature vector $\mathbf{a}^s \sim \mathbb{R}^f$. What those $f$ features represent in the physical world depends on the modality (time-series, image, text...) of the data considered. Let's assume that we are dealing with a minibatch of $n$ such samples simultaneously. In this case it is convenient to stack together all $n$ training samples vertically into:

$$\mathbf{A} = \begin{pmatrix} \mathbf{a}^1 \sim \mathbb{R}^f \\ \vdots \\ \mathbf{a}^n \sim \mathbb{R}^f \end{pmatrix} = \begin{pmatrix} a_1^1 & \dots & a_f^1 \\ \vdots & \vdots & \vdots \\ a_1^n & \dots & a_f^n \end{pmatrix} \sim \mathbb{R}^{n \times f}$$

     As mentioned above, we will concern ourselves with the task of image classification. This means that each training sample $\mathbf{a}^s$ is actually a color image of height $h$, width $w$ and depth $d$ (number of channels, $d = 3$ for RGB for example). As such, feature vectors can be represented as a 3d structure $f \equiv \mathbb{R}^{d \times h \times w}$. For the sake of simplicity, we will only consider square images and denote by $r \equiv h \equiv w$ the spatial resolution of the image. Note that the dimensionality of $f$ grows quadratically with the spatial resolution of the images meaning that, even for modest sizes $\sim 1000$ pixels, we quickly arrive at very high dimensional input data $f \sim 10^6$ features (pixel values). Stacking together all the images present in a minibatch, the raw input data $\mathbf{A}_0 \sim \mathbb{R}^{n \times d \times r \times r}$ therefore starts as a (1+3)d = 4d array whose shape will evolve (in depth as well as in spatial resolution) as it flows deeper into the network as shown in table 1 and discussed more in detail in point 2 below.

   - In addition, each sample $s$ is also associated with its ground-truth categorical label $\mathbf{y}^s_{\text{gt}}$. Denoting by $n_c$ the number of possible classes, $\mathbf{y}^s_{\text{gt}}$ is generally represented as a “One Hot Encoded” (OHE) vector $\mathbf{y}^s_{\text{gt}} \sim \mathbb{R}^{n_c}$. Stacking the $n$ ground-truth vectors all together, we represent the labels via the following structure:

$$\mathbf{Y}_{\text{gt}} = \begin{pmatrix} \mathbf{y}^1_{\text{gt}} \sim \mathbb{R}^{n_c} \\ \vdots \\ \mathbf{y}^n_{\text{gt}} \sim \mathbb{R}^{n_c} \end{pmatrix} = \begin{pmatrix} y_1^1 & \dots & y_{n_c}^1 \\ \vdots & \vdots & \vdots \\ y_1^n & \dots & y_{n_c}^n \end{pmatrix}_{\text{gt}} \sim \mathbb{R}^{n \times n_c}$$

     If sample $s$ actually belongs to the ground-truth class $c^s_{\text{gt}}$, OHE representation means that there is only a single element $\mathbf{y}^s_{\text{gt}}(c^s_{\text{gt}}) = 1$ which is non-zero and all others are identically null $\mathbf{y}^s_{\text{gt}}(c \neq c^s_{\text{gt}}) = 0$ as illustrated in the top left corner of Fig.2.

   The minibatch size $n$, i.e. number of training samples that we stack together for simultaneous processing, should be considered as a hyper-parameter. Selecting the optimal $n$ remains an active area of research and we come back to it in point 4 when we discuss the learning algorithm. In the following, training data for a specific minibatch b refers to the pair $\mathcal{D}_{\text{b}} = (\mathbf{A}_0, \mathbf{Y}_{\text{gt}})_{\text{b}}$. Assuming that the entire training dataset is divided into $N$ minibatches, we can represent it as a list $\mathcal{D}_{\text{training}} = [\mathcal{D}_1, \cdots, \mathcal{D}_N]$. As we will see shortly, it is implicitly assumed that the training points are independent and identically distributed (i.i.d.) according to an unknown distribution.

2. Define a model architecture for the neural network $\mathcal{N}_{\boldsymbol{\mathcal{P}}}$ by arranging together a number $n_\ell$ of layers. Formally, the model should be thought of as a parametrized function $\mathcal{N}_{\boldsymbol{\mathcal{P}}}$ that takes the input data $\mathbf{A}_0$ and returns a probability distribution $\mathbf{Y}_{\text{pred}}$ over $n_c$ classes:

$$\mathbf{Y}_{\text{pred}} = \mathcal{N}_{\boldsymbol{\mathcal{P}}}(\mathbf{A}_0) \sim \mathbb{R}^{n \times n_c} \tag{1}$$

   Evaluating $\mathcal{N}_{\boldsymbol{\mathcal{P}}}$ given some input data $\mathbf{A}_0$ is referred to as the “forward pass” and the resulting $\mathbf{Y}_{\text{pred}}$ defines the probabilistic prediction of the model. Each training sample $s$ gets a prediction vector $\mathbf{y}^s_{\text{pred}} \sim \mathbb{R}^{n_c}$ indicating how “confident” the network is that this sample belongs to any one of the $n_c$ possible classes as illustrated in the top right corner of Fig.2. The final layer of the network ensures proper normalization of the probability distributions so that $\sum_{c=1}^{n_c} (y_c^s)_{\text{pred}} = 1$ independently for all $n$ samples (see section 3). Denoting by $n_p$ the collective number of parameters contained in the trainable layers of the network, we have $\boldsymbol{\mathcal{P}} \sim \mathbb{R}^{n_p}$.

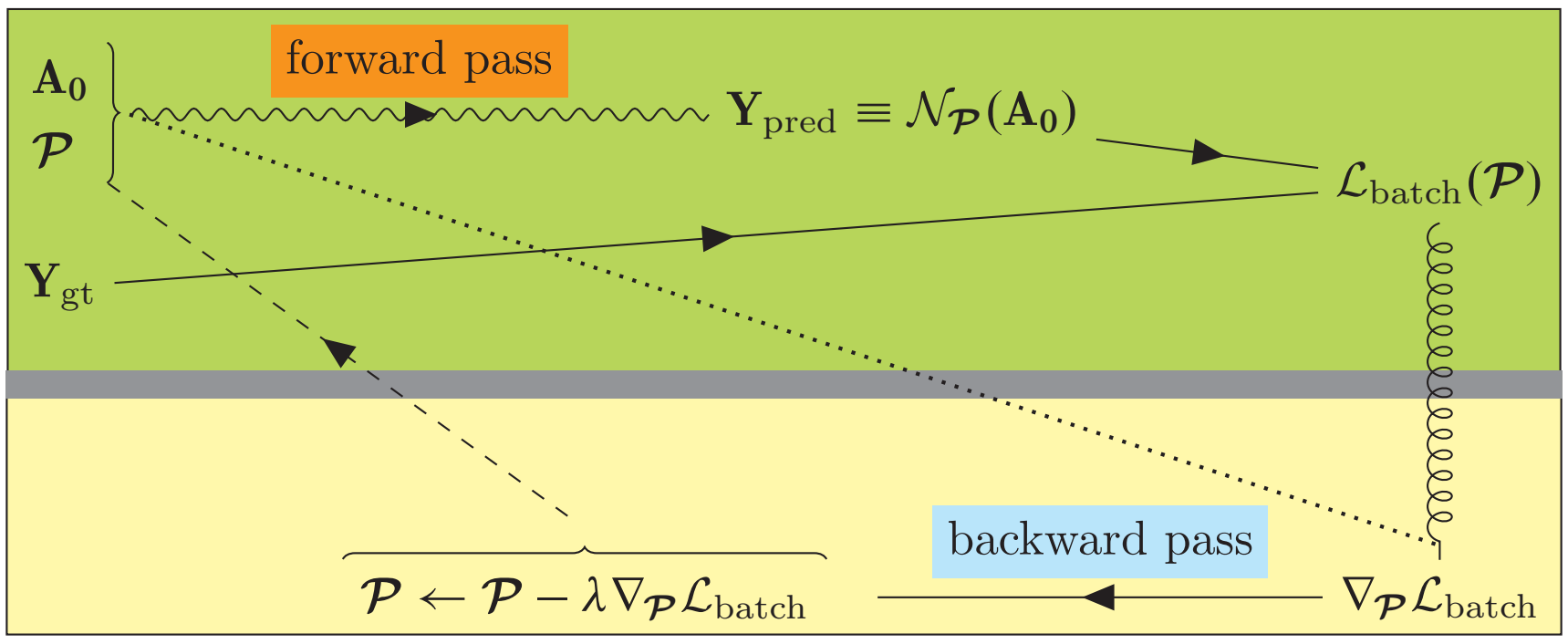

Figure 1: High level cartoon of the iterative training loop. The model $\mathcal{N}_{\mathcal{P}}$ is a probabilistic function parametrized by $\mathcal{P}$ whose purpose is to take in the raw data $\mathbf{A_0}$ and return a probability distribution $\mathbf{Y}_{\text{pred}}$ over a set of $n_c$ classes during the "forward pass". Combining $\mathbf{Y}_{\text{pred}}$ with the ground-truth $\mathbf{Y}_{\text{gt}}$ leads to a scalar $\mathcal{L}_{\text{batch}}(\mathcal{P}) > 0$, known as the loss function, that quantifies the level of mismatch between prediction and ground-truth. The objective of training is to find a better set of parameters $\mathcal{P}$ to minimize the value of the loss over the training data. As discussed in the text, this can be achieved by calculating the gradient $\nabla_{\mathcal{P}} \mathcal{L}_{\text{batch}}$ of the loss during the "backward pass". Calculating this gradient in the context of convolutional neural networks is the focus of this article. Once $\nabla_{\mathcal{P}} \mathcal{L}_{\text{batch}}$ is known, the parameters are updated proportionately to the learning rate $\lambda$. Cycles of forward/backward passes are run iteratively over minibatches of labeled data $\mathcal{D}_{\text{training}}$ until one is satisfied with the overall performance of the model.

In this article, we will consider the following layers:

- **non-trainable**: non-linear activation (4), max-pool (6) & flatten (7)
- **trainable**: fully connected (5), convolution (8) & batch normalization (9)

Inspired by a simple and historically important CNN, we consider a modified version of the famous LeNet-5 model that incorporates a few more modern components (ReLU activation, batch normalization, skip connections...). The architecture of our example network is fully specified in table 1. Its backbone is made up of $n_\ell = 16$ layers comprising of $n_p = 44{,}878$ parameters listed in table 2. Because the architecture does not have loops, it falls under the category of feedforward neural networks which are usually implemented as directed acyclic graphs (DAGs) by deep learning frameworks. Keeping with historical references, we use the MNIST (Modified National Institute of Standards and Technology) dataset as the labeled data $\mathcal{D}_{\text{training}}$. This dataset consists of $n_c = 10$ classes of handwritten digits (0-9) in the form of 70,000 grayscale images (depth $d = 1$ and spatial resolution $r = 28$). A selection of 60,000 samples is assigned to the training set while the remaining 10,000 constitute the test set. With a minibatch of size $n$, the input data is therefore represented as $\mathbf{A_0} \sim \mathbb{R}^{n\times1\times28\times28}$. The "shape" column of table 1 shows how the data is sequentially transformed from pixel space $\mathbf{A_0}$ layer by layer all the way down to "embedding" space $\mathbf{A} \equiv \mathbf{A}_{n_\ell=16}$. The network starts by a series of alternating "convolution — activation — batch normalization — maxpool" layer blocks whose effect is to reduce the spatial resolution of the data while, at the same time, increase its depth. At some point the 3d structure of samples (independent space and depth dimensions of images) are flattened into 1d feature vectors transforming $\mathbf{A_8} \sim \mathbb{R}^{n\times16\times4\times4}$ into a 2d array $\mathbf{A_9} \sim \mathbb{R}^{n\times256}$ which is fed into another series of alternating "fully connected — activation — batch normalization" layer blocks. Note that space and depth information are no longer relevant as interpretable features as soon as data is processed by fully connected layers because of the global connectivity patterns they introduce. The final representation, so-called "embedding", denoted by $\mathbf{A}$ is eventually fed into a softmax layer (section 3) in order to produce a normalized probability distribution $\mathbf{Y}_{\text{pred}} \sim \mathbb{R}^{n\times n_c}$.

The "backward pass" corresponds to an equivalent propagation of error terms $\Delta$'s back up through the layers of $\mathcal{N}_{\mathcal{P}}$ (see section 2). As can be gleaned from table 1, data and error arrays always share the same dimensionality $\mathbf{A}_i \sim \Delta_i$ for all layers.

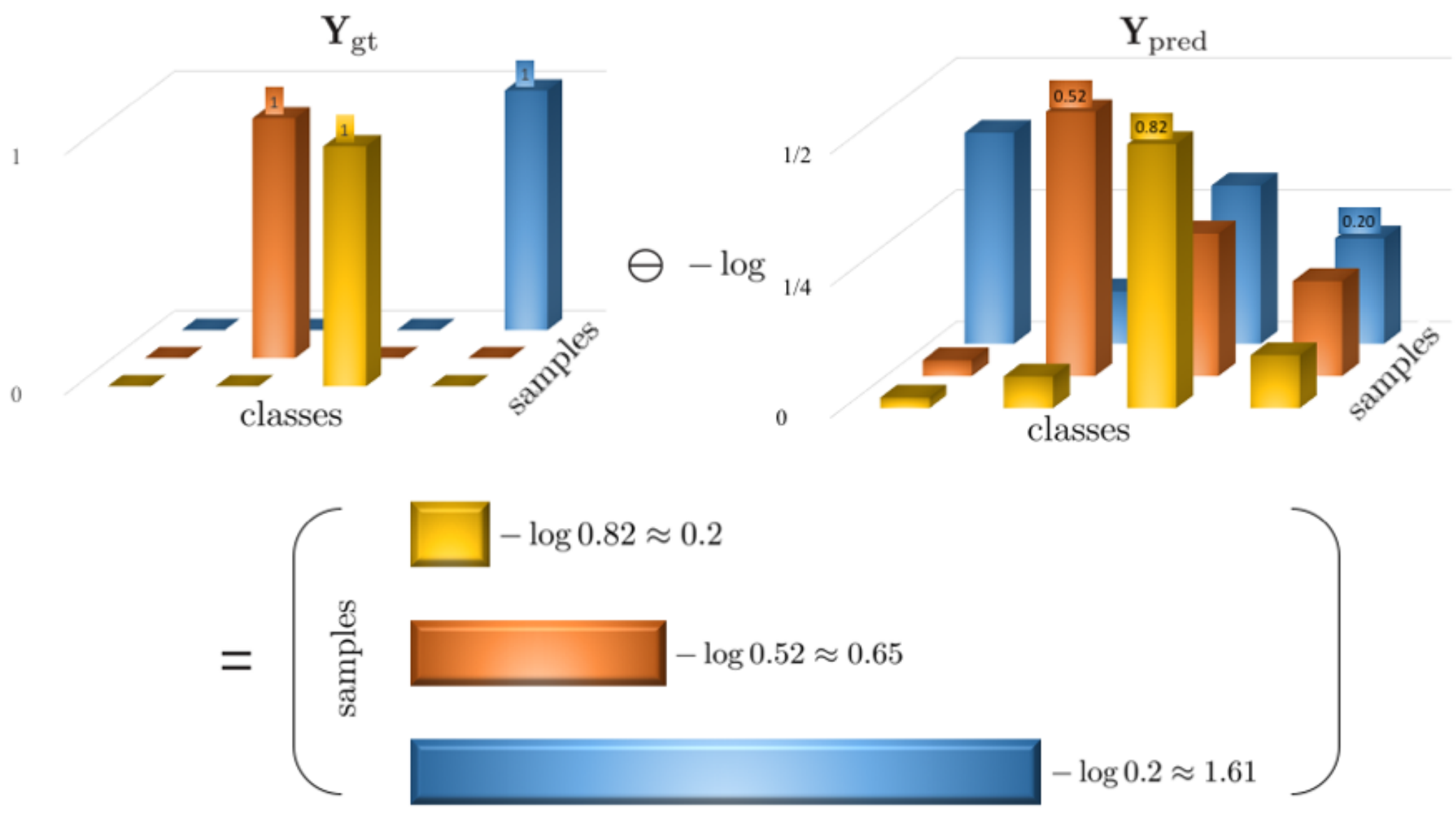

Figure 2: Illustration of the cross-entropy loss function $\ell_{\mathcal{P}}\left(\mathbf{Y}_{\rm gt}, \mathbf{Y}_{\rm pred}\right)$ defining the amount of "mismatch" between the one-hot encoded ground-truth $\mathbf{Y}_{\rm gt}$ and the output probability distribution $\mathbf{Y}_{\rm pred}$ as defined in eq.(2). For clarity, we show only the values of the $\mathbf{y}^s_{\rm pred}(c^s_{\rm gt})$ components for all samples $s$ of $\mathbf{Y}_{\rm pred}$ since they are the only relevant ones as far as the cross-entropy calculation is concerned. (Numerical values are shared with Fig. 5).

3. Define a loss function that measures the amount of disagreement between the predicted $\mathbf{Y}_{\rm pred}$ and the ground-truth $\mathbf{Y}_{\rm gt}$. For classification tasks, it is usual to use the cross-entropy between the predicted probability distribution and the ground-truth distribution:

$$\mathcal{L}_{\rm batch}(\mathcal{P}) = -\mathbf{Y}_{\rm gt} \cdot \log \mathbf{Y}_{\rm pred} \sim \mathbb{R}$$

where the explicit $\mathcal{P}$-dependence of the loss comes its dependence on $\mathbf{Y}_{\rm pred} = \mathcal{N}_{\mathcal{P}}\left(\mathbf{A}_0\right)$ and, recursively, on all the preceding layers of the neural network [1].

In order to gain some insight into how the cross-entropy loss emerges as the natural quantity, let us consider a single training sample $s$ with input data and supervised label pair $\left(\mathbf{a}^s_0, \mathbf{y}^s_{\rm gt}\right)$. Passing this input as an argument to the neural network function $\mathcal{N}_{\mathcal{P}}$ produces a probability distribution vector $\mathbf{y}^s_{\rm pred} = \mathcal{N}_{\mathcal{P}}\left(\mathbf{a}^s_0\right) \sim \mathbb{R}^{n_c}$. Denoting by $c^s_{\rm gt}$ the ground-truth class to which this sample belongs means that all components of $\mathbf{y}^s_{\rm gt} \sim \mathbb{R}^{n_c}$ are identically 0 except for $\mathbf{y}^s_{\rm gt}(c^s_{\rm gt}) \equiv 1$. Because of this OHE representation of $\mathbf{y}^s_{\rm gt}$, its dot-product with $\mathbf{y}^s_{\rm pred}$ produces a single value

$$\mathbf{y}^s_{\rm gt} \cdot \mathbf{y}^s_{\rm pred} = \mathbf{y}^s_{\rm pred}(c^s_{\rm gt}) \sim \mathbb{R}$$

which represents the probability/**likelihood** assigned by $\mathcal{N}_{\mathcal{P}}$ to the actual ground-truth class. Accordingly, a good prediction consists in having a likelihood $0 < \mathbf{y}^s_{\rm pred}(c^s_{\rm gt}) \lessapprox 1$ as high as possible in order to mirror $\mathbf{y}^s_{\rm gt}$. Under the assumption that the $n$ training samples are i.i.d. (as discussed in point 1), the likelihood over the entire minibatch $\mathtt{L}_{\rm batch}$ can be written as a product over the individual likelihoods. The training objective is then formulated as an optimization problem over the parameters $\mathcal{P} \sim \mathbb{R}^{n_p}$ to maximize the minibatch likelihood:

$$\arg\max_{\mathcal{P}} \mathtt{L}_{\rm batch} \; ; \text{ with } \mathtt{L}_{\rm batch} = \prod_{s=1}^{n} \mathbf{y}^s_{\rm pred}(c^s_{\rm gt}) \sim \mathbb{R}$$

[1] Obviously $\mathcal{L}_{\rm batch}$ also depends on the network architecture $\mathcal{N}$ in addition to $\mathcal{P}$ and the training data $(\mathbf{A_0}, \mathbf{Y}_{\rm gt})$ (see also side note). However, as this dependence is usually non-differentiable, we restrict ourselves to static architectures and consider the loss as a function of the parameters and the training data only. We refer the reader to [1] for recent work that formulates architecture search as a gradient-based optimization problem (see point 4) using differentiable losses with respect to $\mathcal{N}$ as an alternative to conventional approaches that use evolution techniques or reinforcement learning over a discrete and non-differentiable search space [2].

Taking the logarithm of the likelihood turns the product into a sum over individual training samples without changing the nature of the optimization objective. Since the log function is strictly monotonic, maximizing the likelihood is equivalent to minimizing the negative log-likelihood:

$$\begin{aligned}
\arg\max_{\mathcal{P}} \mathrm{L}_{\text{batch}} \quad &\Longleftrightarrow \quad \arg\min_{\mathcal{P}} \left(-\log \mathrm{L}_{\text{batch}}\right) \\
&= \quad \arg\min_{\mathcal{P}} \left(-\sum_{s=1}^{n} \log \mathbf{y}^s_{\text{pred}}(c^s_{\text{gt}})\right) \\
&\equiv \quad \arg\min_{\mathcal{P}} \sum_{s=1}^{n} \ell_{\mathcal{P}}\left(\mathbf{y}^s_{\text{gt}}, \mathbf{y}^s_{\text{pred}}\right)
\end{aligned}$$

Thanks to this convenient formulation as a sum over the minibatch, we can identify the amount of mismatch due to an individual training sample $s$ as:

$$\ell_{\mathcal{P}}\left(\mathbf{y}^s_{\text{gt}}, \mathbf{y}^s_{\text{pred}}\right) \equiv -\log \mathbf{y}^s_{\text{pred}}(c^s_{\text{gt}}) \sim \mathbb{R}$$

↓ equivalent to cross-entropy between the distributions $\mathbf{y}^s_{\text{gt}}$ and $\mathbf{y}^s_{\text{pred}}$

(using the OHE representation of $\mathbf{y}^s_{\text{gt}}$)

$$\equiv -\mathbf{y}^s_{\text{gt}} \cdot \log \mathbf{y}^s_{\text{pred}} \sim \mathbb{R}$$

+ illustrative plot

This shows that maximizing the likelihood is equivalent to minimizing the cross-entropy between the ground-truth distribution and the probability distribution vector predicted by the neural network. As can be seen in the illustrative plot, the cross-entropy metric ensures a monotonically decreasing cost from high values when $\mathbf{y}^s_{\text{pred}}(c^s_{\text{gt}}) \ll 1$ (i.e. small likelihood assigned by $\mathcal{N}_{\mathcal{P}}$ to the ground-truth class: bad prediction) down to small cost values as the prediction for the ground-truth class approaches 1, namely $\mathbf{y}^s_{\text{pred}}(c^s_{\text{gt}}) \lessapprox 1$ (i.e. high likelihood assigned by $\mathcal{N}_{\mathcal{P}}$ to the ground-truth class: good prediction).

Going back to the general case of minibatches of training samples, we can dispatch $\ell_{\mathcal{P}}$ to all $n$ samples and express the cross-entropy loss as a vectorized operation:

$$\ell_{\mathcal{P}}\left(\mathbf{Y}_{\text{gt}}, \mathbf{Y}_{\text{pred}}\right) \equiv \begin{pmatrix} \ell_{\mathcal{P}}\left(\mathbf{y}^1_{\text{gt}}, \mathbf{y}^1_{\text{pred}}\right) \sim \mathbb{R} \\ \vdots \\ \ell_{\mathcal{P}}\left(\mathbf{y}^n_{\text{gt}}, \mathbf{y}^n_{\text{pred}}\right) \sim \mathbb{R} \end{pmatrix} = -\mathbf{Y}_{\text{gt}} \ominus \log \mathbf{Y}_{\text{pred}} \sim \mathbb{R}^n \tag{2}$$

where each component corresponds to the loss due to individual samples as illustrated in Fig.2. Using eq.(51) to sum up this loss vector demonstrates that the total loss is indeed given by the cross-entropy between the predicted probability distribution $\mathbf{Y}_{\text{pred}}$ and the ground truth distribution $\mathbf{Y}_{\text{gt}}$ as stated at the beginning of this section:

$$\mathcal{L}_{\text{batch}}(\mathcal{P}) = \sum_{\text{samples}} \ell_{\mathcal{P}}\left(\mathbf{Y}_{\text{gt}}, \mathbf{Y}_{\text{pred}}\right) = -\mathbf{Y}_{\text{gt}} \cdot \log \mathbf{Y}_{\text{pred}} \sim \mathbb{R} \tag{3}$$

In summary, we have shown that the training objective can be formulated as a search for a set of parameters $\mathcal{P} \sim \mathbb{R}^{n_{\mathcal{P}}}$ that minimize the cross-entropy between $\mathbf{Y}_{\text{pred}}$ and $\mathbf{Y}_{\text{gt}}$:

$$\arg\min_{\mathcal{P}} \mathcal{L}_{\text{batch}}(\mathcal{P})$$

Minimizing the training error has the effect of maximizing the similarity between ground-truth and model prediction distributions, i.e. the likelihood that the model is able to produce the correct labels on the training dataset.

4. The purpose of the learning algorithm is to provide a concrete strategy to solve the optimization objective discussed in the previous point. It is known that the inductive biases introduced by different optimization algorithms [3] and proper initializations [4, 5] play a crucial role in the quality of learning and in the generalization ability of the learned models. We focus on the celebrated gradient descent algorithm which, despite its age [6], remains the "workhorse" of deep learning.

   Gradient descent is an intuitive procedure where the parameters $\mathcal{P}$ are iteratively corrected by a small vector $\delta\mathcal{P} \ll \mathcal{P}$ carefully chosen so as to reduce the loss $\mathcal{L}_{\text{batch}}(\mathcal{P}+\delta\mathcal{P}) < \mathcal{L}_{\text{batch}}(\mathcal{P})$. Obviously, a decrease in the loss implies that $\mathbf{Y}_{\text{pred}}$ gradually becomes a little more accurate description of $\mathbf{Y}_{\text{gt}}$ (see point 3). The minimization is implemented by repeating this update many times over different batches of training data $\mathcal{D}_{\text{training}}$. How to find $\delta\mathcal{P}$? Since the parameter update is assumed to be small, we can perform a Taylor expansion of the loss function:

$$\mathcal{L}_{\text{batch}}(\mathcal{P}+\delta\mathcal{P}) \approx \mathcal{L}_{\text{batch}}(\mathcal{P}) + \delta\mathcal{P}^t \cdot \nabla_{\mathcal{P}}\mathcal{L}_{\text{batch}} + \frac{1}{2}\,\delta\mathcal{P}^t\,\nabla^2_{\mathcal{P}}\mathcal{L}_{\text{batch}}\,\delta\mathcal{P} + \cdots$$

   Restricting ourselves to 1$^{\text{st}}$-order effects only, the change in loss values is determined by the dot-product $\delta\mathcal{P}^t \cdot \nabla_{\mathcal{P}}\mathcal{L}_{\text{batch}}$. Clearly, this term depends on the angle between the parameter update $\delta\mathcal{P} \sim \mathbb{R}^{n_p}$ and the gradient $\nabla_{\mathcal{P}}\mathcal{L}_{\text{batch}} \sim \mathbb{R}^{n_p}$. It reaches its maximum when both vectors are aligned with each other showing that the gradient [2] can be interpreted as the direction of steepest ascent. In other words, $\nabla_{\mathcal{P}}\mathcal{L}_{\text{batch}}$ is a vector whose direction in $\mathbb{R}^{n_p}$ space is the one along which $\mathcal{L}_{\text{batch}}$ grows the fastest. Given that our goal is to decrease the loss value, the optimal parameter update consists in choosing a vector $\delta\mathcal{P} \sim \mathbb{R}^{n_p}$ that lies in the opposite direction of the gradient, i.e. the direction of steepest descent. Now that the direction of $\delta\mathcal{P}_{\text{steepest descent}}$ is known, we can fix its magnitude by introducing a "learning rate" $0 < \lambda \ll 1$ such that parameters are updated according to:

$$\mathcal{P} \;\Longleftarrow\; \mathcal{P} + \delta\mathcal{P}_{\text{steepest descent}} \tag{4}$$

$$\delta\mathcal{P}_{\text{steepest descent}} \;\equiv\; -\lambda\nabla_{\mathcal{P}}\mathcal{L}_{\text{batch}} = -\lambda \begin{pmatrix} \partial\mathcal{L}_{\text{batch}}/\partial\mathcal{P}_1 \\ \vdots \\ \partial\mathcal{L}_{\text{batch}}/\partial\mathcal{P}_{n_p} \end{pmatrix} \sim \mathbb{R}^{n_p} \tag{5}$$

   Obviously, this approach requires the explicit calculation of the gradient $\nabla_{\mathcal{P}}\mathcal{L}_{\text{batch}}$ and, for implementation reasons that will introduced in section 2 and be the focus of the rest of the article, the evaluation of $\nabla_{\mathcal{P}}\mathcal{L}_{\text{batch}}$ is referred to as the "backward pass".

   In our case, the input data consists of minibatches representing only a subset of the entire training dataset. As a result, the gradient is calculated based on a limited number $n$ of samples for each update and the learning algorithm is typically referred to as "stochastic gradient descent" (SGD) to reflect the noise introduced by this finite-size estimation of the gradient. (This is in contrast with "batch" gradient descent that uses the entire available dataset.) Choosing the minibatch size $n$ remains a delicate issue which is entangled with the learning rate $\lambda$ and its evolution during training. It is customarily believed [7, 8] that smaller values of $n$ lead to better generalization performance: an effect attributed to the randomness in minibatch sampling inducing an "exploratory" behavior of SGD dynamics. In fact, one can show that the covariance of the minibatch noise is related to the Hessian $\nabla^2_{\mathcal{P}}\mathcal{L}_{\text{batch}}$ of the loss function [9] hinting at a connection between noise and higher-order effects. Overall, noise appears to play a crucial role by implicitly providing a form of regularization that may help escape saddle points and facilitate training. In contrast, a number of studies advocate larger batch sizes in order to reduce the number of parameter updates and allow the use of distributed training without having to sacrifice performance [10, 11, 12]. Obviously, the geometry and training dynamics of loss landscapes remain topics of intense research; fruitful connections have appeared with models and tools coming statistical physics [13].

[2] The gradient $\nabla_{\mathcal{P}}\mathcal{L}_{\text{batch}}$ of a function $\mathcal{L}_{\text{batch}}(\mathcal{P}) : \mathbb{R}^{n_p} \to \mathbb{R}$ is defined as the vector of partial derivatives with respect to each of its parameters: $\nabla_{\mathcal{P}}\mathcal{L}_{\text{batch}} = \left(\partial\mathcal{L}_{\text{batch}}/\partial\mathcal{P}_1, \cdots, \partial\mathcal{L}_{\text{batch}}/\partial\mathcal{P}_{n_\ell}\right)$. Trainable layers typically have more than a single parameter and, denoting by $n_p$ the total number of parameters contained in the $n_\ell$ layers, we have therefore $\nabla_{\mathcal{P}}\mathcal{L}_{\text{batch}} \sim \mathcal{P} \sim \mathbb{R}^{n_p}$.

Deep learning frameworks offer a whole zoo of gradient-based optimization strategies that decorate the "canonical" SGD presented here [14, 15]. Nevertheless, all these methods share one common characteristic which is the necessity to evaluate the gradient $\nabla_{\boldsymbol{\mathcal{P}}} \mathcal{L}_{\text{batch}}$. Complexity-wise, 1[st]-order algorithms are efficient since $\nabla_{\boldsymbol{\mathcal{P}}} \mathcal{L}_{\text{batch}} \sim \mathbb{R}^{n_p}$ is linear with the total number of parameters in the network. Intuitively, one may think that 2[nd] order methods involving the Hessian would improve the search by utilizing information about the curvature of the loss landscape. Unfortunately, the quadratic growth of the Hessian $\nabla_{\boldsymbol{\mathcal{P}}} \mathcal{L}^2_{\text{batch}} \sim \mathbb{R}^{n_p \times n_p}$ coupled with the need for matrix inversion render such approaches prohibitively expensive for large networks. Even higher-order methods suffer from increasingly worse scaling and, additionally, cannot exploit well-established linear algebra tools. In practice, the overwhelming majority of neural networks are trained using simple gradient descent methods based on $\nabla_{\boldsymbol{\mathcal{P}}} \mathcal{L}_{\text{batch}}$.

5. After the parameters have been updated for a minibatch $\mathcal{D}_{\text{b}}$, we can advance to the next minibatch $\mathcal{D}_{\text{b}+1} \in \mathcal{D}_{\text{training}}$, make a forward pass through the network to get the loss and do another parameter update via the backward pass. This loop over minibatches and update of parameters may be repeated until the loss function has decreased enough and the training data is well approximated by the neural network function, i.e. $\mathbf{Y}_{\text{gt}} \approx \mathbf{Y}_{\text{pred}} = \mathcal{N}_{\boldsymbol{\mathcal{P}}}(\mathbf{A_0})$ , $\forall\, \mathcal{D} \in \mathcal{D}_{\text{training}}$.

   One run over the complete $\mathcal{D}_{\text{training}}$ is called an "epoch" and this iterative procedure is summarized in a concise high-level pseudo-code as well as in the cartoon of Fig. 1. Given enough iterations a "good" learning algorithm may lead to a vanishingly small loss $\mathcal{L}(\boldsymbol{\mathcal{P}}_{\text{overfit}}) \approx 0$ meaning that the function $\mathcal{N}_{\boldsymbol{\mathcal{P}}_{\text{overfit}}}$ is capable of perfectly reproducing all the training samples. This is a signal of overfitting and implies that $\mathcal{N}_{\boldsymbol{\mathcal{P}}_{\text{overfit}}}$ has little chance of generalizing to new input data. This is why, in addition to the training loss, it is common practice to also monitor the accuracy on an independent held-out testing set in order to be able to stop training as soon as the testing performance no longer increases; a technique known as early stopping. There exists a large number of "regularization" strategies that aim to address this trade-off between optimization on the training data and generalization to previously unseen datasets. In addition to traditional penalties based on the norm of the parameters such as sparsity inducing L1 (Lasso regression) and L2 regularizations (ridge regression also known as weight decay), other approaches like dropout [16] (a prominent candidate), dataset augmentation techniques, label smoothing, bagging... are frequently used by deep learning practitioners. Nevertheless, controlling the accuracy of a model remains a critical topic of research and training networks that yield state-of-the-art performance still involves many tricks (one may look at [17, 18] for an introduction).

**Closing words on supervised learning from 20,000 feet...** Before moving on to the core of this article which is the description of backpropagation, let us finish this introduction by making a few general observations. First of all, although number of parameters may not the the best way to quantify model complexity [19, 20], let us note that the example CNN discussed in this article (tables 1 and 2) is under-parametrized: the number $n_p = 44{,}878$ of parameters is smaller than the 60,000 samples available for training. This situation is somewhat unusual for modern state-of-the-art deep learning architectures that typically gravitate towards a heavy over-parametrization of the models. For example, the famous AlexNet which propelled deep learning under the spotlight in 2012 after winning the ImageNet ILSVRC challenge (by an unprecedented margin) contained about 60 million parameters trained on a dataset of "only" 1.2 million images [21]. Empirical evidence suggests that models with larger capacity are surprisingly resistant to overfitting and continue to improve their generalization error (see [22] for an extreme example) even when trained without explicit regularization [23].

Putting things in perspective, one may be tempted to think of the task of training the neural network function $\mathcal{N}_{\boldsymbol{\mathcal{P}}}$ as a simple interpolation problem on the dataset $\mathcal{D}_{\text{training}}$. However, the very high-dimensional nature of the input data raises major difficulties. Indeed, it is well known that with increasing dimensionality all data points become equally far away from each other and the notion of nearest neighbors may no longer be relevant [24]; a consequence of the "curse of dimensionality". Dense sampling of a unit hypercube of dimension $d \sim 10^6$ (lower range of real-world data, see point 1 for images) into a grid of (very poor) resolution $\varepsilon \sim 0.1$ would require an unreachable and absolutely absurd number $(1/\varepsilon)^d \sim 10^{1{,}000{,}000}$ of samples. Even if, because of some intrinsic structural constraints, real-world data happens to lie in the vicinity of lower dimensional manifolds [25], training samples are forever condemned to be immensely isolated from each other effectively ruling out naïve interpolation.

**Iterative training: pseudo-code**

Let us mention that the whole training procedure can be summarized in a surprisingly short high-level program mirroring the cartoon of Fig.1. The first step consists in choosing a neural network architecture $\mathcal{N}_\mathcal{P}$, i.e. a parametrized function that, given some input data, returns a probability distribution $\mathbf{Y}_\text{pred}$ over a set of $n_c$ predefined classes. This prediction is then compared to the ground-truth $\mathbf{Y}_\text{gt}$ and the quality of the fit is measured via:

$$\mathcal{L}_\text{batch}\left(\mathbf{Y}_\text{gt}, \mathbf{Y}_\text{pred} = \mathcal{N}_\mathcal{P}(\mathbf{A_0})\right) \equiv \mathcal{L}_\text{batch}\left(\mathcal{N}, \mathcal{P}, \mathcal{D}_\text{batch}\right) \sim \mathbb{R}$$

Conceptually, the loss $\mathcal{L}_\text{batch}$ can be expressed as a scalar function of the architecture $\mathcal{N}$, the current value of the parameters $\mathcal{P} \sim \mathbb{R}^{n_p}$ and the training data $\mathcal{D}_\text{batch} = \left(\mathbf{A_0}, \mathbf{Y}_\text{gt}\right)_\text{batch}$.

*(In our case $\mathcal{N}$ is defined as the functional composition of the layers specified in table 1, parametrized by $\mathcal{P}$ described in table 2 and trained on $\mathcal{D}_{training}$ = MNIST with the cross-entropy loss defined in eq.(3)).*

As explained in point 4, it is necessary to calculate the gradient of $\mathcal{L}_\text{batch}$ with respect to $\mathcal{P}$ in order to perform an SGD update. Restricting ourselves to static network architectures, the gradient can be formally returned by a vector-valued "backpropagation" function $\mathcal{B}$ implicitly parametrized by $\mathcal{N}$:

$$\nabla_{\boldsymbol{\mathcal{P}}}\mathcal{L}_\text{batch} = \mathcal{B}_\mathcal{N}\left(\boldsymbol{\mathcal{P}}, \mathcal{D}_\text{batch}\right) \sim \mathbb{R}^{n_p}$$

In practice, deep learning frameworks expose rich high-level APIs based on automatic differentiation [26] to efficiently implement $\mathcal{B}_\mathcal{N}$ and support complex control flow such as branches/loops as part of the emerging concept of "differentiable programming" [27, 28]. In this article we will calculate the gradient of each layer analytically and define $\mathcal{B}_\mathcal{N}$ as the composition of all the layer-level gradient functions. Once $\mathcal{B}_\mathcal{N}$ is known, training proceeds by i) evaluating the backpropagation function iteratively over the list of supervised observations $\mathcal{D}_\text{training} = [\mathcal{D}_1, \cdots, \mathcal{D}_N]$ and ii) updating the parameters $\boldsymbol{\mathcal{P}}$ each time according to eqs.(4,5). This can be translated in single line of code by defining a training function:

$$\boldsymbol{\mathcal{P}}_\text{trained}\left(\mathcal{N}\right) = \texttt{foldl}\left(\backslash \boldsymbol{\mathcal{P}}, \mathcal{D} \rightarrow \boldsymbol{\mathcal{P}} - \lambda \mathcal{B}_\mathcal{N}\left(\boldsymbol{\mathcal{P}}, \mathcal{D}\right)\right) \boldsymbol{\mathcal{P}}_\text{initial}\ \mathcal{D}_\text{training} \sim \mathbb{R}^{n_p}$$

that returns a set of trained parameters $\boldsymbol{\mathcal{P}}_\text{trained}$ for any given architecture $\mathcal{N}$. Here, `foldl` stands for the left fold function (Haskell-specific syntax; exists under different keywords in other programming languages) that takes 3 arguments: an updating function, an initialized accumulator $\boldsymbol{\mathcal{P}}_\text{initial}$ and a list $\mathcal{D}_\text{training}$ upon which to evaluate the function.

Instead, deep learning is usually presented as a data cascade across multiple levels of abstraction starting with raw unstructured high-dimensional input and finishing with lower dimensional representations called "embeddings". In the case of CNNs, the first few layers are suspected to encode simple features such as edges and colors. Deeper layers pick up the task by trying to detect higher-level motifs that start to resemble more familiar objects. Finally, this hierarchical cascade is believed to produce abstract concepts that are easier to separate into distinct classes [29]. This scenario can be joined with a complimentary interpretation going under the name of "manifold hypothesis" [30]. There, the idea is that the learning process should be seen as a progressive linearization of the topology of the input data that starts as complicated interwoven class manifolds and culminates into embeddings that are separable by simple hyperplanes [31, 32] as illustrated in Fig.3. Giving further support to the importance of these learned representations, it is known that algebraic operations on embeddings can combine high-level concepts into semantically meaningful relationships [33, 34]. Somewhat orthogonally, compression of input data into efficient representations during the training dynamics can also be approached from the point of view of information theory [35].

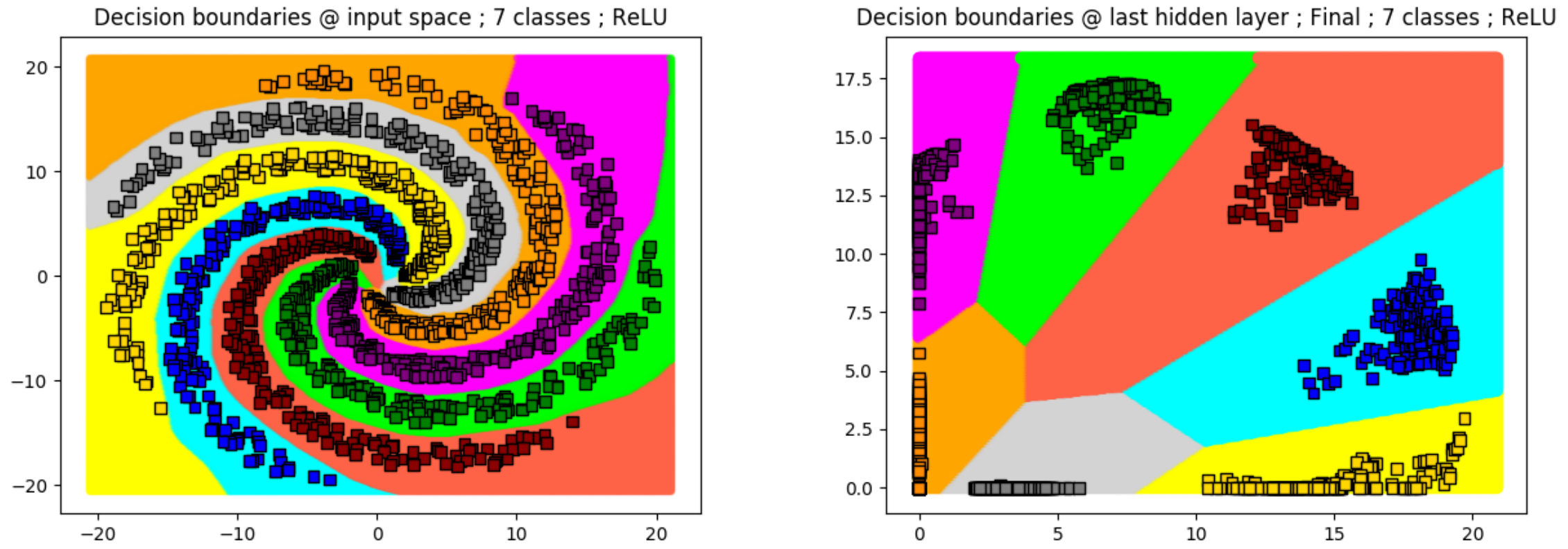

Figure 3: Concrete visualization of the "manifold hypothesis" using a synthetic spiral dataset composed of 7 interleaved classes represented by different colors; see [32] for all details including animations of the training dynamics using different activation functions. **Left)** Raw input data $\equiv \mathbf{A_0} \sim \mathbb{R}^{n\times 2}$. **Right)** Data representation at the level of the "embedding" $\equiv \mathbf{A} \sim \mathbb{R}^{n\times 2}$. Thanks to the 2d geometry of both $\mathbf{A}/\mathbf{A_0}$, one can see how $\mathcal{N}_\mathcal{P}$ has learned to separate the training samples so that classes can be separated by linear boundaries.

This hierarchical and compositional view of deep learning is not specific to CNNs and images but, rather, may be a natural consequence of the generative process which created the data in the first place. Indeed, physical world data is usually modeled in terms of recursive compositions of fundamental building blocks into more and more complex systems. In this case, neural networks can be seen as an attempt to "reverse-engineer" this decomposition of the generative process into a hierarchy of simpler steps [36]. As such, it is probable that successful deep learning architectures implicitly provide very strong priors on the types of statistical patterns that can be discovered from data [37]. Despite empirical evidence that "deeper is better", the actual role played by depth and its impact on the expressiveness of approximating functions remains a topic of active theoretical research [38, 39].

Although this narrative is very widespread and influential [40], it is worth emphasizing that there is still no consensus regarding what it is that the current incarnations of neural networks really do learn; a lot of work is going into understanding and visualizing the predictions of deep learning models [41, 42]. Among many known drawbacks let us mention that neural networks can effortlessly fit data with random labels [43], are alarmingly fooled by minute and imperceptible adversarial perturbations [44], lack a sense of geometric context [45], learn superficial details not related to semantically relevant concepts [46], are brittle against benign distributional shift [47]... Another obvious limitation stems from the implicit static closed-world environment assumption: models can only ever predict classes that belong to the training set. They are forced into making wrong predictions (with uncontrollable confidence) if presented with new objects [48] or with nonsensical input [49]. The practicality (or lack thereof) of deep neural networks in the wild, the role of prior knowledge [50], whether modern architectures are anything more than (somewhat fast and loose [51]) exceptional curve-fitting machines that bear no resemblance whatsoever to human perception let alone reasoning [52]... and many more issues continue [3] to be nebulous and hotly debated topics. Nonetheless, deep learning has undeniably turned out to be very good at discovering intricate structures in high-dimensional data and continues to establish new state-of-the-art performance in many academic fields beyond computer vision. The last few years have also shown how neural networks can have a meaningful impact in wide variety of commercial products. Nevertheless, a healthy dose of skepticism shows that there is still some way to go before they can fully be deployed as trusted components of critical and ever-changing real-world applications.

With this long introduction out of the way, let us now move on to the main purpose of this article which is a pedagogical presentation of backpropagation in the context of CNNs.

[3](as evidenced by the famous, time-tested, quote *"The question of whether machines can think is about as relevant as the question of whether submarines can swim"* [53], these discussions have a long history)

| Layer | Forward pass | Shape | Backward pass |
|---|---|---|---|
| Input data | $\mathbf{A}_0$ | $\mathbb{R}^{n\times1\times28\times28}$ | $n$ grayscale ($d=1$) square images of $28\times28$ resolution |
| Convolution<br>Activation<br>Batch normalization | $\mathbf{A}_1 = \mathbf{w}_0^{\mathfrak{p}_0} \star \mathbf{A}_0 + \widetilde{\mathbf{b}_0}$<br>$\mathfrak{p}_0 = \left(k_0 = 5, s_0 = 1, p_0 = 0\right)$<br>$\mathbf{A}_2 = g(\mathbf{A}_1)$<br>$\mathbf{A}_3 = f_{4d}^t\left(\overline{f_{2d}^t(\mathbf{A}_2)}\,\widetilde{\mathbf{w}_2} + \widetilde{\mathbf{b}_2}\right)$ | $\mathbb{R}^{n\times6\times24\times24}$ | $\Delta_1 = \Delta_2 \circ g'(\mathbf{A}_1)$<br>$\Delta_2 = f_{4d}^t\left[\left(576n f_{2d}^t(\Delta_3)\widetilde{\mathbf{w}_2} - \widetilde{\sum_s f_{2d}^t(\Delta_3)\widetilde{\mathbf{w}_2}} - \overline{f_{2d}^t(\mathbf{A}_2)} \circ \widetilde{\sum_s \overline{f_{2d}^t(\mathbf{A}_2)} \circ f_{2d}^t(\Delta_3)\widetilde{\mathbf{w}_2}}\right)/576n\widetilde{\sigma_2}\right]$<br>$\Delta_3 = \widetilde{\Delta_4} \circ g_{\mathfrak{p}}'(\mathbf{A}_3)$ |
| Maxpool | $\mathbf{A}_4 = \text{maxpool}_{\mathfrak{p}}\, \mathbf{A}_3$<br>$\mathfrak{p} = \left(k = 2\,, s = 2\,, p = 0\right)$ | $\mathbb{R}^{n\times6\times12\times12}$ | $\Delta_4 = \overset{\curvearrowleft t_d}{\mathbf{w}_4}{}^{\mathfrak{p}_4'} \star \Delta_5$<br>$\mathfrak{p}_4' = \left(k_4' = k_4 = 5\,,\ s_4' = 1/s_4 = 1\,,\ p_4' = k_4 - p_4 - 1 = 4\right)$ |
| Convolution<br>Activation<br>Batch normalization | $\mathbf{A}_5 = \mathbf{A}_4 \star \mathbf{w}_4^{\mathfrak{p}_4} + \widetilde{\mathbf{b}_4}$<br>$\mathfrak{p}_4 = \left(k_4 = 5, s_4 = 1, p_4 = 0\right)$<br>$\mathbf{A}_6 = g(\mathbf{A}_5)$<br>$\mathbf{A}_7 = f_{4d}\left(\overline{f_{2d}^t(\mathbf{A}_6)}\,\widetilde{\mathbf{w}_6} + \widetilde{\mathbf{b}_6}\right)$ | $\mathbb{R}^{n\times16\times8\times8}$ | $\Delta_5 = \Delta_6 \circ g'(\mathbf{A}_5)$<br>$\Delta_6 = f_{4d}^t\left[\left(64n f_{2d}^t(\Delta_7)\widetilde{\mathbf{w}_6} - \widetilde{\sum_s f_{2d}^t(\Delta_7)\widetilde{\mathbf{w}_6}} - \overline{f_{2d}^t(\mathbf{A}_6)} \circ \widetilde{\sum_s \overline{f_{2d}^t(\mathbf{A}_6)} \circ f_{2d}^t(\Delta_7)\widetilde{\mathbf{w}_6}}\right)/64n\widetilde{\sigma_6}\right]$<br>$\Delta_7 = \widetilde{\Delta_8} \circ g_{\mathfrak{p}}'(\mathbf{A}_7)$ |
| Maxpool | $\mathbf{A}_8 = \text{maxpool}_{\mathfrak{p}}\, \mathbf{A}_7$<br>$\mathfrak{p} = \left(k = 2, s = 2, p = 0\right)$ | $\mathbb{R}^{n\times16\times4\times4}$ | $\Delta_8 = \text{fold}\,\Delta_9$ |
| Flatten | $\mathbf{A}_9 = \text{flatten}\,\mathbf{A}_8$ | $\mathbb{R}^{n\times256}$ | $\Delta_9 = \Delta_{10}\mathbf{w}_9^t$ |
| Fully connected<br>Activation<br>Batch normalization | $\mathbf{A}_{10} = \mathbf{A}_9\mathbf{w}_9 + \widetilde{\mathbf{b}_9}$<br>$\mathbf{A}_{11} = g(\mathbf{A}_{10})$<br>$\mathbf{A}_{12} = \overline{\mathbf{A}}_{11}\widetilde{\mathbf{w}_{11}} + \widetilde{\mathbf{b}_{11}}$ | $\mathbb{R}^{n\times120}$ | $\Delta_{10} = \Delta_{11} \circ g'(\mathbf{A}_{10})$<br>$\Delta_{11} = \frac{1}{n\widetilde{\sigma_{11}}}\left(n\Delta_{12}\widetilde{\mathbf{w}_{11}} - \widetilde{\sum_s \Delta_{12}\widetilde{\mathbf{w}_{11}}} - \overline{\mathbf{A}}_{11} \circ \widetilde{\sum_s \overline{\mathbf{A}}_{11} \circ \Delta_{12}\widetilde{\mathbf{w}_{11}}}\right)$<br>$\Delta_{12} = \Delta_{13}\mathbf{w}_{12}^t$ |
| Fully connected<br>Activation<br>Batch normalization | $\mathbf{A}_{13} = \mathbf{A}_{12}\mathbf{w}_{12} + \widetilde{\mathbf{b}_{12}}$<br>$\mathbf{A}_{14} = g(\mathbf{A}_{13})$<br>$\mathbf{A}_{15} = \overline{\mathbf{A}}_{14}\widetilde{\mathbf{w}_{14}} + \widetilde{\mathbf{b}_{14}}$ | $\mathbb{R}^{n\times84}$ | $\Delta_{13} = \Delta_{14} \circ g'(\mathbf{A}_{13})$<br>$\Delta_{14} = \frac{1}{n\widetilde{\sigma_{14}}}\left(n\Delta_{15}\widetilde{\mathbf{w}_{14}} - \widetilde{\sum_s \Delta_{15}\widetilde{\mathbf{w}_{14}}} - \overline{\mathbf{A}}_{14} \circ \widetilde{\sum_s \overline{\mathbf{A}}_{14} \circ \Delta_{15}\widetilde{\mathbf{w}_{14}}}\right)$<br>$\Delta_{15} = \Delta_{16}\mathbf{w}_{15}^t$ |
| Fully connected | $\mathbf{A} \equiv \mathbf{A}_{16} = \mathbf{A}_{15}\mathbf{w}_{15} + \widetilde{\mathbf{b}_{15}}$ | $\mathbb{R}^{n\times10}$ | $\Delta_{16} = \mathbf{Y}_{\text{pred}} - \mathbf{Y}_{\text{gt}}$ |
| Softmax | $\mathbf{Y}_{\text{pred}} = \text{softmax}\,\mathbf{A}$ | $\mathbb{R}^{n\times10}$ | probability distribution over $n_c = 10$ classes for all images |

Table 1: Illustration of a typical Convolutional Neural Network (CNN) architecture inspired by the historical LeNet-5. Notice how patterns "convolution/fully connected — activation — batch normalization" are grouped together and repeated. Shortcut connections, another component of modern architectures such as ResNet, are explained in detail even though they are absent from this example network. As usual in CNNs, the shape of the data representations during the **forward pass** (to be read from **top to bottom**) starts by becoming "fatter" (deeper and spatially smaller) early in the network before being flattened into a "thin" feature vector whose dimension is gradually decreased to eventually match the desired number of classes $n_c = 10$ for classification; the final layer before the softmax, sometimes referred to as the "embedding", is denoted as $\mathbf{A} \equiv \mathbf{A}_{16}$. This in contrast to the **backward pass** (to be read from **bottom to top**) where the error arrays, corresponding to the gradient of the loss function with respect to the data, follow the exact opposite dimensionality changes. As can be gleaned from the expressions above and demonstrated in the main text, error backpropagation may be seen as a general function $\Delta_{i-1}\left(\Delta_i, \mathbf{A}_{i-1}, \boldsymbol{\mathcal{P}}_{i-1}\right)$ of the upstream error $\Delta_i$, original data array $\mathbf{A}_{i-1}$ from the forward pass and parameters $\boldsymbol{\mathcal{P}}_{i-1}$; layer-specific implementations of the downstream error terms $\Delta_{i-1}$ can be found in the relevant sections.

*(Note that the dimensionality of convolutional kernels and of arrays that are wrapped by geometrical reshape operations, such as $f_{2d}$ and $f_{4d}$, designed to handle minibatches of image data are provided explicitly in the caption of table 2. More details about the architecture and the dataflow are provided in point 2 of section 1.)*

| Parameters | | | Dimensionality | | Loss derivative |
|---|---|---|---|---|---|
| Fully connected | $\boldsymbol{\mathcal{P}}_{15}$ | $\mathbf{w}_{15}$ | $\mathbb{R}^{84\times 10}$ | 840 | $\partial\mathcal{L}_{\text{batch}}/\partial\mathbf{w}_{15} = \mathbf{A}^t_{15}\Delta_{16}$ |
| | | $\mathbf{b}_{15}$ | $\mathbb{R}^{10}$ | 10 | $\partial\mathcal{L}_{\text{batch}}/\partial\mathbf{b}_{15} = \sum_s \Delta_{16}$ |
| Batch normalization | $\boldsymbol{\mathcal{P}}_{14}$ | $\mathbf{w}_{14}$ | $\mathbb{R}^{84}$ | 84 | $\partial\mathcal{L}_{\text{batch}}/\partial\mathbf{w}_{14} = \text{diag}\left(\overline{\mathbf{A}}^t_{14}\Delta_{15}\right)$ |
| | | $\mathbf{b}_{14}$ | $\mathbb{R}^{84}$ | 84 | $\partial\mathcal{L}_{\text{batch}}/\partial\mathbf{b}_{14} = \sum_s \Delta_{15}$ |
| Fully connected | $\boldsymbol{\mathcal{P}}_{12}$ | $\mathbf{w}_{12}$ | $\mathbb{R}^{120\times 84}$ | 10,080 | $\partial\mathcal{L}_{\text{batch}}/\partial\mathbf{w}_{12} = \mathbf{A}^t_{12}\Delta_{13}$ |
| | | $\mathbf{b}_{12}$ | $\mathbb{R}^{84}$ | 84 | $\partial\mathcal{L}_{\text{batch}}/\partial\mathbf{b}_{12} = \sum_s \Delta_{13}$ |
| Batch normalization | $\boldsymbol{\mathcal{P}}_{11}$ | $\mathbf{w}_{11}$ | $\mathbb{R}^{120}$ | 120 | $\partial\mathcal{L}_{\text{batch}}/\partial\mathbf{w}_{11} = \text{diag}\left(\overline{\mathbf{A}}^t_{11}\Delta_{12}\right)$ |
| | | $\mathbf{b}_{11}$ | $\mathbb{R}^{120}$ | 120 | $\partial\mathcal{L}_{\text{batch}}/\partial\mathbf{b}_{11} = \sum_s \Delta_{12}$ |
| Fully connected | $\boldsymbol{\mathcal{P}}_{9}$ | $\mathbf{w}_{9}$ | $\mathbb{R}^{256\times 120}$ | 30,720 | $\partial\mathcal{L}_{\text{batch}}/\partial\mathbf{w}_{9} = \mathbf{A}^t_{9}\Delta_{10}$ |
| | | $\mathbf{b}_{9}$ | $\mathbb{R}^{120}$ | 120 | $\partial\mathcal{L}_{\text{batch}}/\partial\mathbf{b}_{9} = \sum_s \Delta_{10}$ |
| Batch normalization | $\boldsymbol{\mathcal{P}}_{6}$ | $\mathbf{w}_{6}$ | $\mathbb{R}^{16}$ | 16 | $\partial\mathcal{L}_{\text{batch}}/\partial\mathbf{w}_{6} = \text{diag}\left(\overline{f^t_{2d}(\mathbf{A}_6)}^t f^t_{2d}(\Delta_7)\right)$ |
| | | $\mathbf{b}_{6}$ | $\mathbb{R}^{16}$ | 16 | $\partial\mathcal{L}_{\text{batch}}/\partial\mathbf{b}_{6} = \sum_{s,r} \Delta_{7}$ |
| Convolution | $\boldsymbol{\mathcal{P}}_{4}$ | $\mathbf{w}_{4}$ | $\mathbb{R}^{16\times 6\times 5\times 5}$ | $16\times 150 = 2{,}400$ | $\partial\mathcal{L}_{\text{batch}}/\partial\mathbf{w}_{4} = \text{roll}\left[f_{2d}\left(\Delta_5\right)\phi\left(\mathbf{A}_4\right)^t\right]$ |
| | | $\mathbf{b}_{4}$ | $\mathbb{R}^{16}$ | 16 | $\partial\mathcal{L}_{\text{batch}}/\partial\mathbf{b}_{4} = \sum_{s,r} \Delta_{5}$ |
| Batch normalization | $\boldsymbol{\mathcal{P}}_{2}$ | $\mathbf{w}_{2}$ | $\mathbb{R}^{6}$ | 6 | $\partial\mathcal{L}_{\text{batch}}/\partial\mathbf{w}_{2} = \text{diag}\left(\overline{f^t_{2d}(\mathbf{A}_2)}^t f^t_{2d}(\Delta_3)\right)$ |
| | | $\mathbf{b}_{2}$ | $\mathbb{R}^{6}$ | 6 | $\partial\mathcal{L}_{\text{batch}}/\partial\mathbf{b}_{2} = \sum_{s,r} \Delta_{3}$ |
| Convolution | $\boldsymbol{\mathcal{P}}_{0}$ | $\mathbf{w}_{0}$ | $\mathbb{R}^{6\times 1\times 5\times 5}$ | $6\times 25 = 150$ | $\partial\mathcal{L}_{\text{batch}}/\partial\mathbf{w}_{0} = \text{roll}\left[f_{2d}\left(\Delta_1\right)\phi\left(\mathbf{A}_0\right)^t\right]$ |
| | | $\mathbf{b}_{0}$ | $\mathbb{R}^{6}$ | 6 | $\partial\mathcal{L}_{\text{batch}}/\partial\mathbf{b}_{0} = \sum_{s,r} \Delta_{1}$ |
| Total number of parameters = 44,878 | | | | | |

Table 2: Parameters are presented from top to bottom in the order in which they are updated during the backpropagation algorithm described in section 2. Notice that all gradient components with respect to parameters $\boldsymbol{\mathcal{P}}_{i-1}$ share the same pattern regardless of the type of (linear) layer:

**weights:** $\partial\mathcal{L}_{\text{batch}}/\partial\mathbf{w}_{i-1} \sim \mathbf{A}^t_{i-1}\Delta_i$ Matrix product between the upstream error $\Delta_i$ and the transpose of the data array $\mathbf{A}_{i-1}$ (up to geometrical transformations such as $f_{2d}$, $f_{4d}$, roll, diag...) This shows that intermediate data arrays originating from the forward pass need to be cached in memory to be combined, at a later point, with error arrays during the backward pass; illustration in Fig.4.

**biases:** $\partial\mathcal{L}_{\text{batch}}/\partial\mathbf{b}_{i-1} \sim \sum \Delta_i$ Tensor-contraction of the upstream error. In the case of error arrays associated with image data, the sum runs over the spatial dimensions (indicated by the $r$ subscript) in addition to minibatch samples (indicated by the $s$ subscript) in the summation symbol $\sum_{s,r}$.

*(Details about layer-specific implementations of the* **components of the gradient $\nabla_{\boldsymbol{\mathcal{P}}}\mathcal{L}_{\text{batch}}$** *are provided in the relevant sections of the main text.)*

---

For the sake of completeness, we report here the dimensionality of transformed arrays and of the convolutional kernels relevant both for table 1 as well as for the gradient expressions above:

| | | |
|---|---|---|
| $f^t_{2d}(\mathbf{A}_6) \sim \mathbb{R}^{(8\times 8\times n)\times 16}$ | $f^t_{2d}(\Delta_7) \sim \mathbb{R}^{(8\times 8\times n)\times 16}$ | $\text{diag}\left(\mathbb{R}^{16\times 16}\right) \sim \mathbb{R}^{16}$ |
| $f_{2d}(\Delta_5) \sim \mathbb{R}^{16\times(8\times 8\times n)}$ | $\phi\left(\mathbf{A}_4\right) \sim \mathbb{R}^{(6\times 5\times 5)\times(8\times 8\times n)}$ | $\text{roll}\left(\mathbb{R}^{16\times(6\times 5\times 5)}\right) \sim \mathbb{R}^{16\times 6\times 5\times 5}$ |
| $f^t_{2d}(\mathbf{A}_2) \sim \mathbb{R}^{(24\times 24\times n)\times 6}$ | $f^t_{2d}(\Delta_3) \sim \mathbb{R}^{(24\times 24\times n)\times 6}$ | $\text{diag}\left(\mathbb{R}^{6\times 6}\right) \sim \mathbb{R}^{6}$ |
| $f_{2d}(\Delta_1) \sim \mathbb{R}^{6\times(24\times 24\times n)}$ | $\phi\left(\mathbf{A}_0\right) \sim \mathbb{R}^{(1\times 5\times 5)\times(24\times 24\times n)}$ | $\text{roll}\left(\mathbb{R}^{6\times(1\times 5\times 5)}\right) \sim \mathbb{R}^{6\times 1\times 5\times 5}$ |
| $\mathbf{w}_0^{\text{p}_0} \sim \mathbb{R}^{6\times 1\times k_0\times k_0}$ | $\mathbf{w}_4^{\text{p}_4} \sim \mathbb{R}^{16\times 6\times k_4\times k_4}$ | $\overset{\curvearrowleft t_{\text{d}}\,\text{p}'_4}{\mathbf{w}_4} \sim \mathbb{R}^{6\times 16\times k'_4\times k'_4}$ |

The purely geometrical transformation from $\mathbf{w}_4^{\text{p}_4}$ to $\overset{\curvearrowleft t_{\text{d}}\,\text{p}'_4}{\mathbf{w}_4}$ is explained in a dedicated paragraph and illustrated in an animation of Fig. 12.

## 2 Gradient evaluation via backpropagation

As discussed in the introduction, the key component behind the iterative training loop displayed in Fig 1 consists in being able to provide an explicit expression for the gradient $\nabla_{\mathcal{P}}\mathcal{L}_{\text{batch}}$ of the loss function with respect to the parameters $\mathcal{P}$ of the neural network $\mathcal{N}_{\mathcal{P}}$. This section is dedicated to an in-depth immersion into the fundamental mechanics behind one such implementation of gradient calculation generally referred to as "backpropagation". For typical machine learning datasets, loss functions tend to have signatures of the following type:

$$\mathcal{L}_{\text{batch}}(\mathcal{P}) : \mathbb{R}^{n_p} \to \mathbb{R} \quad \text{with} \quad n_p \gg 1$$

where a very high-dimensional parameter space ($n_p = 44{,}878$ in our example network) is reduced to a scalar value. In this case, backpropagation is computationally efficient [4] since the gradient can be evaluated by a single forward/backward cycle through the neural network, i.e. roughly-speaking with a time complexity on the order of only 2 evaluations of $\mathcal{N}_{\mathcal{P}}$ (at the expense of memory consumption). This section presents the logic of backpropagation in a generic way applicable to any network architecture and layer type as done in tables 1 and 2 for our example network.

**How to start? Bootstrap with loss function & softmax** Let us start by recognizing that, by definition, the total derivative of $\mathcal{L}_{\text{batch}}$ is given by:

$$\begin{aligned} \mathrm{d}\mathcal{L}_{\text{batch}} &= \nabla_{\mathcal{P}}\mathcal{L}_{\text{batch}} \cdot \mathrm{d}\mathcal{P} \\ &\downarrow \quad \text{gradient as a vector of partial derivatives for the } n_\ell \text{ layers; see footnote 2} \\ \mathrm{d}\mathcal{L}_{\text{batch}} &\equiv \sum_{p=1}^{n_\ell} \frac{\partial \mathcal{L}_{\text{batch}}}{\partial \mathcal{P}_p} \cdot \mathrm{d}\mathcal{P}_p \end{aligned} \tag{6}$$

As we will discover in this section, all the components of $\nabla_{\mathcal{P}}\mathcal{L}_{\text{batch}}$ can be extracted by an intuitive pattern matching process that operates by recursively going backwards layer-by-layer through $\mathcal{N}_{\mathcal{P}}$.

Accordingly, let us start with the definition of the cross-entropy loss function in eq.(3) and begin evaluating $\mathrm{d}\mathcal{L}_{\text{batch}}$ at the output level of the neural network:

$$\begin{aligned} \mathrm{d}\mathcal{L}_{\text{batch}} &= \sum_{\text{samples}} \mathrm{d}\ell_{\mathcal{P}}\left(\mathbf{Y}_{\text{gt}}, \mathbf{Y}_{\text{pred}}\right) \\ &\boxed{+} \\ &= \nabla_{\mathcal{P}}\ell_{\mathcal{P}}\left(\mathbf{Y}_{\text{gt}}, \mathbf{Y}_{\text{pred}}\right) \cdot \mathrm{d}\mathbf{Y}_{\text{pred}} \end{aligned} \tag{7}$$

The first thing to notice is that $\mathrm{d}\mathcal{L}_{\text{batch}}$ is expressed as a **Frobenius product between an "error"** [5] **term and the total derivative of a layer**. We will come back to this important observation below but, for now, let us push the calculation one step further. As specified in table 1 and explained in detail in section 3, the predicted probability distribution $\mathbf{Y}_{\text{pred}} = \text{softmax}\,\mathbf{A}$ is determined by applying the softmax function to the "embedded" representation $\mathbf{A}$. In general, this means that its total derivative $\mathrm{d}\mathbf{Y}_{\text{pred}}$ can be formulated in terms of $\mathrm{d}\mathbf{A}$ through the chain rule (its exact expression is provided in eq.(13) as part of the relevant section dedicated to the softmax function).

In order to continue evaluating $\mathrm{d}\mathcal{L}_{\text{batch}}$, let us now consider the error term involving the cross-entropy loss function $\ell_{\mathcal{P}}$ defined in eq.(2):

$$\begin{aligned} \nabla_{\mathcal{P}}\ell_{\mathcal{P}}(\mathbf{Y}_{\text{gt}}, \mathbf{Y}_{\text{pred}}) &= \nabla_{\mathcal{P}}\left(-\mathbf{Y}_{\text{gt}} \ominus \log \mathbf{Y}_{\text{pred}}\right) \\ &\boxed{+} \\ &= -\mathbf{Y}_{\text{gt}} \circ \frac{1}{\mathbf{Y}_{\text{pred}}} \end{aligned}$$

---

[4] To be compared with a straightforward computation of all partial derivatives of $\mathcal{L}_{\text{batch}}(\mathcal{P})$ independently from each which requires $\sim \mathcal{O}(n_p)$ evaluations of $\mathcal{N}_{\mathcal{P}}$. Such "forward mode" implementations of differentiation are efficient only for functions $\mathbb{R}^n \to \mathbb{R}^m$ where the dimensionality of the output space is larger than that of the input space, i.e. $n < m$.

[5] Besides the fact that this term is directly related to the loss function $\ell_{\mathcal{P}}\left(\mathbf{Y}_{\text{gt}}, \mathbf{Y}_{\text{pred}}\right)$, the origin of the naming convention as an "error" term will become evident later.

The next step consists in combining the expression above for $\nabla_{\mathcal{P}} \ell_{\mathcal{P}}(\mathbf{Y}_{\text{gt}}, \mathbf{Y}_{\text{pred}})$ along with that of $\mathrm{d}\mathbf{Y}_{\text{pred}}$ derived in eq.(13) and reproduced here as a reminder:

$$\mathrm{d}\mathbf{Y}_{\text{pred}} = \mathbf{Y}_{\text{pred}} \circ \left( \mathrm{d}\mathbf{A} - \widetilde{\mathbf{Y}_{\text{pred}} \ominus \mathrm{d}\mathbf{A}} \right)$$

together into the Frobenius product of eq.(7) in order to get:

$$\begin{aligned} \mathrm{d}\mathcal{L}_{\text{batch}} &= \left( -\mathbf{Y}_{\text{gt}} \circ \frac{1}{\mathbf{Y}_{\text{pred}}} \right) \cdot \left( \mathbf{Y}_{\text{pred}} \circ \left( \mathrm{d}\mathbf{A} - \widetilde{\mathbf{Y}_{\text{pred}} \ominus \mathrm{d}\mathbf{A}} \right) \right) \\ & \boxed{+} \\ &= \mathbf{Y}_{\text{gt}} \cdot \widetilde{\mathbf{Y}_{\text{pred}} \ominus \mathrm{d}\mathbf{A}} - \mathbf{Y}_{\text{gt}} \cdot \mathrm{d}\mathbf{A} \\ & \boxed{+} \\ &= \left( \mathbf{Y}_{\text{pred}} - \mathbf{Y}_{\text{gt}} \right) \cdot \mathrm{d}\mathbf{A} \end{aligned}$$

Comparing the expression above for $\mathrm{d}\mathcal{L}_{\text{batch}}$ with that of eq.(7), we observe that the structure as a Frobenius product between an error term and the total derivative of a layer is preserved as we go from the level of the predicted probability distribution to that of the embedding layer. Namely:

$$\begin{aligned} \mathrm{d}\mathcal{L}_{\text{batch}} &= \underbrace{\nabla_{\mathcal{P}} \ell_{\mathcal{P}} \left( \mathbf{Y}_{\text{gt}}, \mathbf{Y}_{\text{pred}} \right)}_{\text{upstream error}} \cdot \mathrm{d} \left( \underbrace{\mathbf{Y}_{\text{pred}}}_{\text{current layer}} \right) \\ & \downarrow \quad \text{our } 1^{\text{st}} \text{ backward step through the network} \\ \mathrm{d}\mathcal{L}_{\text{batch}} &= \overbrace{\left( \mathbf{Y}_{\text{pred}} - \mathbf{Y}_{\text{gt}} \right)}^{\text{downstream error}} \cdot \mathrm{d} \left( \overbrace{\mathbf{A}}^{\text{previous layer}} \right) \end{aligned} \tag{8}$$

In other words, this first step in the evaluation of $\mathrm{d}\mathcal{L}_{\text{batch}}$ can be seen as going backwards through one layer of the neural network: we went from an expression involving $\mathbf{Y}_{\text{pred}}$ to a similar expression that now involves the preceding layer $\mathbf{A}$. In this process, the "upstream" error at the level of $\mathbf{Y}_{\text{pred}}$ has been modified into a new "downstream" expression at the level of $\mathbf{A}$.

At this point, it is useful to make a connection with our example network by pattern matching eq.(8) against the downstream error $\Delta_i \equiv \mathbf{Y}_{\text{pred}} - \mathbf{Y}_{\text{gt}}$ and the embedding layer $\mathbf{A} \equiv \mathbf{A}_i$ with $i = 16$ inferred from table 1. As the difference between the predicted probability distribution and the ground-truth, the naming of $\Delta_i$ as an "error" term is self-explanatory. In summary, the backward pass starting at the level of the loss, through the softmax layer and back up to the embedding layer is given by:

**cross-entropy & softmax**: backward pass

$$\begin{aligned} \mathrm{d}\mathcal{L}_{\text{batch}} &= \Delta_i \cdot \mathrm{d}\mathbf{A}_i \\ \Delta_i &= \mathbf{Y}_{\text{pred}} - \mathbf{Y}_{\text{gt}} \end{aligned} \tag{9}$$

More generally, $\Delta_i$ corresponds to the gradient of the loss function with respect to the data array $\mathbf{A}_i$. For consistency, **we will continue to refer to the descendants of $\Delta_i$ as generalized "error" terms**.

**Recursive backwards error flow** Let us now formalize this backwards propagation of the error up through the layers of the network into a high-level generic framework.

As already discussed in the introduction, deep learning models should be understood as the composition $\mathcal{N}_{\mathcal{P}} \equiv f_{n_\ell} \circ \cdots \circ f_1$ of a set of differentiable functions $\{f_1, \cdots, f_{n_\ell}\}$ that define $n_\ell$ layers. For the sake of simplicity [6], let us begin with the assumption that the data $\mathbf{A}_i$ at the $i^{\text{th}}$ layer depends only on its data predecessor $\mathbf{A}_{i-1}$ at the $(i-1)^{\text{th}}$ level and, potentially, a set $\mathcal{P}_{i-1}$ of adjustable parameters. Denoting by $f_i$ the function representing the corresponding layer of $\mathcal{N}_{\mathcal{P}}$, we have:

$$\mathbf{A}_i \equiv f_i(\mathbf{A}_{i-1}, \mathcal{P}_{i-1}) \tag{10}$$

[6] Obviously, this assumption of locality for $\mathbf{A}_i$ neglects the possibility of long-range data dependencies; those can easily be taken into account as explained in a side note dedicated to shortcut connections.

**Small side note about shortcut connections**

Even though the accuracy of statistical learning systems over complex classification tasks has unambiguously benefited from the ever-increasing depths of modern neural networks, very deep architectures (made possible thanks to smart initialization schemes and normalization layers) have empirically revealed the emergence of stubborn degradation effects. For example, it turns out that adding more layers to an already well-trained network leads to a decrease in accuracy (even when measured over the training dataset which suggests that overfitting is not the root cause for the degradation). Paradoxically, constructing the extra layers as identity mappings, one expects that deeper networks should not have a higher training error than shallower ones. Unfortunately, existing solvers often struggle to learn arithmetic concepts even as simple as the identity function [54].

The general idea behind shortcut connections is to decompose the learning task into a fixed identity mapping supplemented by a learned residual function. Accordingly, a shortcut connection linking layer $i-k$ directly down to layer $i$, eliminating the locality restriction of eq.(10), can be implemented via element-wise addition:

$$\mathbf{A}_i \equiv \left( f_i \circ \cdots \circ f_{i-k} \right) \mathbf{A}_{i-k} + \mathbf{A}_{i-k}$$

where $k \geq 2$ stands for the number of skipped layers and where the sequential composition of all the $k$ intermediate layers defines the residual function. This concept was popularized by the successful ResNet architecture [55] which has gone on to inspire the design of many current state-of-the-art networks. During the backward pass, it is the upstream error $\Delta_i$ that is bypassed unchanged straight up to layer $i-k$ where it is combined with the usual error propagation $\Delta_{i-k}$ of the residual function:

$$\Delta_i \cdot \mathrm{d}\mathbf{A}_i = \left( \sum_{p=i-k}^{i} \frac{\partial \mathcal{L}_{\mathrm{batch}}}{\partial \boldsymbol{\mathcal{P}}_p} \cdot \mathrm{d}\boldsymbol{\mathcal{P}}_p \right) + \left( \Delta_{i-k} + \Delta_i \right) \cdot \mathrm{d}\mathbf{A}_{i-k}$$

$+$

In addition to learning decomposition via identity mappings and error backpropagation bypass, the success behind skip connections may also originate from an emerging form of inductive bias that tends to "convexify" loss landscapes [56].

*(In case that $\mathbf{A}_{i-k}$ and $\mathbf{A}_i$ do not have the same dimensionality one needs to perform a linear projection of $\mathbf{A}_{i-k}$. This is usually achieved by $1\times1$ convolutions for 3d data and fully connected layers for 1d data but comes at the cost of introducing new trainable parameters.)*

As emphasized in the previous paragraph, the total derivative of the minibatch loss function $\mathcal{L}_{\mathrm{batch}}$ follows a particular structure as a Frobenius product of the form $\mathrm{d}\mathcal{L}_{\mathrm{batch}} = \Delta_i \cdot \mathrm{d}\mathbf{A}_i$. Using the generic expression for $\mathbf{A}_i$ provided above, we continue the evaluation of $\mathrm{d}\mathcal{L}_{\mathrm{batch}}$ by:

- formally expanding $\mathrm{d}\mathbf{A}_i = \left(\partial f_i / \partial \mathbf{A}_{i-1}\right) \mathrm{d}\mathbf{A}_{i-1} + \left(\partial f_i / \partial \boldsymbol{\mathcal{P}}_{i-1}\right) \mathrm{d}\boldsymbol{\mathcal{P}}_{i-1}$
- plugging this expression back into the Frobenius product defining $\mathrm{d}\mathcal{L}_{\mathrm{batch}}$

Carrying out these steps and gathering the resulting terms into a coherent division between gradient components with respect data, so-called error terms, and those with respect to parameters yields:

$$\mathrm{d}\mathcal{L}_{\mathrm{batch}} = \underbrace{\Delta_i}_{\text{upstream error}} \cdot \, \mathrm{d} \, ( \underbrace{\mathbf{A}_i}_{i^{\text{th}} \text{ layer}} )$$

$+$

$$= \overbrace{\Delta_{i-1}}^{\text{downstream error}} \cdot \, \mathrm{d} \, ( \overbrace{\mathbf{A}_{i-1}}^{(i-1)^{\text{th}} \text{ layer}} ) + \overbrace{\frac{\partial \mathcal{L}_{\mathrm{batch}}}{\partial \boldsymbol{\mathcal{P}}_{i-1}}}^{\text{component of } \nabla_{\boldsymbol{\mathcal{P}}} \mathcal{L}_{\mathrm{batch}}} \cdot \, \mathrm{d} \, \boldsymbol{\mathcal{P}}_{i-1}$$

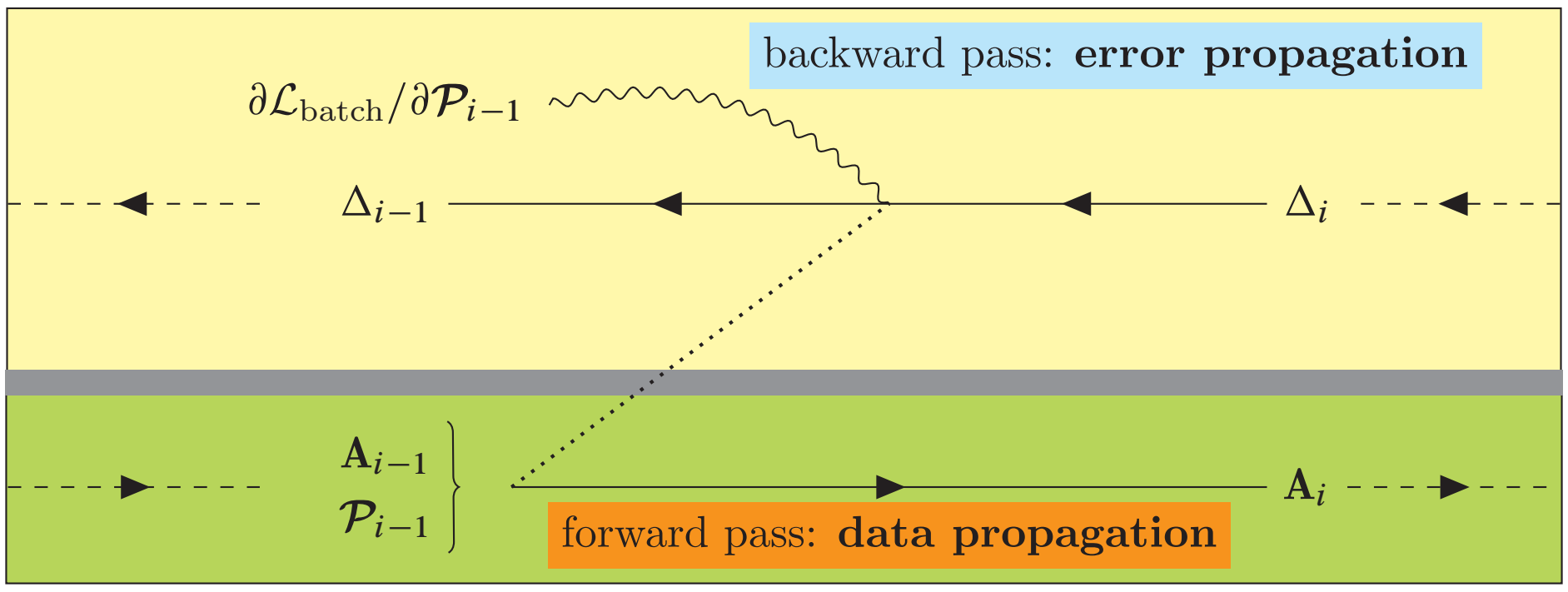

Figure 4: Illustration of how the recursive relation of eq.(11) can be visualized as a backwards error flow from "upstream" $\Delta_i$ into a "downstream" error $\Delta_{i-1}$. This terminology reflects the fact that this error backpropagation is analogous to the forward pass where we follow the propagation of the data the other way around. In case that the layer function $f_i$ during the forward pass has adjustable parameters $\boldsymbol{\mathcal{P}}_{i-1}$, a side-branch away from the error flow evaluates into a gradient component $\partial\mathcal{L}_{\text{batch}}/\partial\boldsymbol{\mathcal{P}}_{i-1}$. The point-dashed line indicates that information from the forward pass needs to be cached in memory and combined with the upstream gradient $\Delta_i$ in order to propagate the error and evaluate the gradient component. The bottom panel simply illustrates the data propagation step from $\mathbf{A}_{i-1}$ to $\mathbf{A}_i$ during the forward pass.

We see that the evaluation of $\mathrm{d}\mathcal{L}_{\text{batch}} = \Delta_i \cdot \mathrm{d}\mathbf{A}_i$ one more step after the embedding layer leads to the emergence of a recursive relation made up of two components:

- $\Delta_{i-1} \cdot \mathrm{d}\mathbf{A}_{i-1}$ In complete analogy with the discussion around eq.(8), this should be interpreted as another backward step through $\mathcal{N}_{\boldsymbol{\mathcal{P}}}$ from the $i^{\text{th}}$ layer back to the $(i-1)^{\text{th}}$ layer. In the process, the upstream error $\Delta_i$ is transformed into a downstream error $\Delta_{i-1}$ which preserves the familiar structure of error backpropagation as a recursive Frobenius product.

- $\dfrac{\partial\mathcal{L}_{\text{batch}}}{\partial\boldsymbol{\mathcal{P}}_{i-1}} \cdot \mathrm{d}\boldsymbol{\mathcal{P}}_{i-1}$ Comparing the latest expression of $\mathrm{d}\mathcal{L}_{\text{batch}}$ derived above against its formal expansion of in eq.(6), we understand that this term should be identified with the gradient of the loss function with respect to parameters $\boldsymbol{\mathcal{P}}_{i-1}$.

This recursive pattern where each step of evaluation of $\mathrm{d}\mathcal{L}_{\text{batch}}$ defines a backwards flow through $\mathcal{N}_{\boldsymbol{\mathcal{P}}}$ from an upstream error $\Delta_i$ into a downstream error $\Delta_{i-1}$ and a side-branch that evaluates to a component $\partial\mathcal{L}_{\text{batch}}/\partial\boldsymbol{\mathcal{P}}_{i-1}$ of the gradient is illustrated in the cartoon of Fig.4.

As revealed in the optional content above, it is important to realize that both $\Delta_{i-1}$ and $\partial\mathcal{L}_{\text{batch}}/\partial\boldsymbol{\mathcal{P}}_{i-1}$ may be seen as general functions that depend on the upstream error $\Delta_i$ as well as on the data and parameters $(\mathbf{A}_{i-1}, \boldsymbol{\mathcal{P}}_{i-1})$ at the $(i-1)^{\text{th}}$ layer that were computed during the forward pass as indicated by the point-dashed line in Fig.4. In other words:

$$\Delta_{i-1} \to \Delta_{i-1}\left(\Delta_i, \mathbf{A}_{i-1}, \boldsymbol{\mathcal{P}}_{i-1}\right) \quad \text{and} \quad \partial\mathcal{L}_{\text{batch}}/\partial\boldsymbol{\mathcal{P}}_{i-1} \to \partial\mathcal{L}_{\text{batch}}/\partial\boldsymbol{\mathcal{P}}_{i-1}\left(\Delta_i, \mathbf{A}_{i-1}, \boldsymbol{\mathcal{P}}_{i-1}\right).$$

This shows that the time efficiency gains obtained by backpropagation (see footnote 4) are mitigated by the **additional memory requirements** necessary to store intermediate data arrays; a manifestation of the classic space-time trade-off.

Recursive backwards **error flow** & **gradient extraction**

$$\Delta_i \cdot \mathrm{d}\mathbf{A}_i = \Delta_{i-1} \cdot \mathrm{d}\mathbf{A}_{i-1} + \frac{\partial\mathcal{L}_{\text{batch}}}{\partial\boldsymbol{\mathcal{P}}_{i-1}} \cdot \mathrm{d}\boldsymbol{\mathcal{P}}_{i-1} \tag{11}$$

Note that if the layer $f_i$ does not have any adjustable parameters, i.e. $\mathbf{A}_i = f_i(\mathbf{A}_{i-1})$, there are no gradient components associated with this layer and the recursion relation of eq.(11) reduces to a simple backward propagation along the error flow only, i.e. $\Delta_i \cdot \mathrm{d}\mathbf{A}_i = \Delta_{i-1} \cdot \mathrm{d}\mathbf{A}_{i-1}$. In case that the network involves shortcut connections, the recursive backwards error flow is easily generalized as shown in the dedicated side note.

**Unrolling backpropagation** Starting from the error at the stage of the embedding layer given by eq.(9) and unrolling the recursion relation all the way back to the input data layer $\mathbf{A}_0$ leads to:

$$\begin{aligned}
\mathrm{d}\mathcal{L}_{\text{batch}} &= \Delta_i \cdot \mathrm{d}\mathbf{A}_i \\
&= \frac{\partial \mathcal{L}_{\text{batch}}}{\partial \boldsymbol{\mathcal{P}}_{i-1}} \cdot \mathrm{d}\boldsymbol{\mathcal{P}}_{i-1} + \Delta_{i-1} \cdot \mathrm{d}\mathbf{A}_{i-1} \\
&= \left( \frac{\partial \mathcal{L}_{\text{batch}}}{\partial \boldsymbol{\mathcal{P}}_{i-1}} \cdot \mathrm{d}\boldsymbol{\mathcal{P}}_{i-1} + \frac{\partial \mathcal{L}_{\text{batch}}}{\partial \boldsymbol{\mathcal{P}}_{i-2}} \cdot \mathrm{d}\boldsymbol{\mathcal{P}}_{i-2} \right) + \Delta_{i-2} \cdot \mathrm{d}\mathbf{A}_{i-2} \\
\downarrow \;& \text{recursively unrolling the backwards error flow and gradient component extraction} \\
&= \left( \sum_{p=1}^{n_\ell - 1} \frac{\partial \mathcal{L}_{\text{batch}}}{\partial \boldsymbol{\mathcal{P}}_p} \cdot \mathrm{d}\boldsymbol{\mathcal{P}}_p \right) + \Delta_1 \cdot \mathrm{d}\mathbf{A}_1 \\
&= \sum_{p=1}^{n_\ell} \frac{\partial \mathcal{L}_{\text{batch}}}{\partial \boldsymbol{\mathcal{P}}_p} \cdot \mathrm{d}\boldsymbol{\mathcal{P}}_p + \underbrace{\Delta_0 \cdot \mathrm{d}\mathbf{A}_0}_{=0}
\end{aligned}$$

This demonstrates that the recursion relation indeed recovers the formal expansion of eq.(6). Therefore, a step-by-step identification allows the extraction of all the components $\partial \mathcal{L}_{\text{batch}} / \partial \boldsymbol{\mathcal{P}}_p$ of the gradient of the loss function $\nabla \mathcal{L}_{\text{batch}}(\boldsymbol{\mathcal{P}}) = \left( \partial \mathcal{L}_{\text{batch}} / \partial \boldsymbol{\mathcal{P}}_1, \cdots, \partial \mathcal{L}_{\text{batch}} / \partial \boldsymbol{\mathcal{P}}_{n_\ell} \right)$. (Note that we terminated the recursion by considering the input data to be fixed, i.e. $\mathrm{d}\mathbf{A}_0 = 0$)

## 3 Softmax layer

**Forward pass** In neural network architectures designed for classification tasks, the final layer is typically chosen to be the softmax function. Its purpose is to transform the final embedded representation $\mathbf{A} \sim \mathbb{R}^{n \times n_c}$, known as "logits", into a probability distribution $\mathbf{Y}_{\text{pred}} \sim \mathbb{R}^{n \times n_c}$ over the $n_c$ classes. This can be accomplished by a multi-dimensional generalization of the logistic function:

**Softmax: forward pass**

$$\mathbf{Y}_{\text{pred}} \equiv \operatorname{softmax} \mathbf{A} = \frac{e^{\mathbf{A}}}{\sum_c e^{\mathbf{A}}} \sim \mathbb{R}^{n \times n_c} \tag{12}$$

where the sum runs over the $c = \{1, \cdots, n_c\}$ features which, at this stage, directly correspond to the $n_c$ possible classes as illustrated in Fig 5. In order to get a more explicit formulation of the broadcasting arithmetic going on under the hood in eq.(12), let's introduce the normalizing vector:

$$\alpha \equiv \frac{1}{\sum_c e^{\mathbf{A}}} = \begin{pmatrix} \alpha_1 \equiv 1 / \sum_c e^{a_c^1} \sim \mathbb{R} \\ \vdots \\ \alpha_n \equiv 1 / \sum_c e^{a_c^n} \sim \mathbb{R} \end{pmatrix} \sim \mathbb{R}^n$$

that needs to be broadcast into $\alpha \to \widetilde{\alpha} \sim \mathbb{R}^{n \times n_c}$ for compatibility with $e^{\mathbf{A}}$. Therefore, a more explicit representation of $\mathbf{Y}_{\text{pred}}$ can be presented as:

$$\mathbf{Y}_{\text{pred}} \equiv \operatorname{softmax} \mathbf{A} = \widetilde{\alpha} \circ e^{\mathbf{A}} = \begin{pmatrix} \alpha_1 e^{a_1^1} & \dots & \alpha_1 e^{a_{n_c}^1} \\ \vdots & \vdots & \vdots \\ \alpha_n e^{a_1^n} & \dots & \alpha_n e^{a_{n_c}^n} \end{pmatrix} \sim \mathbb{R}^{n \times n_c}$$

+

From this expression and the definition of $\alpha$, it is clear that the rows (representing minibatch training data samples) of $\mathbf{Y}_{\text{pred}}$ are indeed normalized correctly and can be interpreted as probabilities:

$$\begin{pmatrix} \sum_c \mathbf{y}_{\text{pred}}^1 \\ \vdots \\ \sum_c \mathbf{y}_{\text{pred}}^n \end{pmatrix} = \begin{pmatrix} \alpha_1 \sum_c e^{a_c^1} \\ \vdots \\ \alpha_n \sum_c e^{a_c^n} \end{pmatrix} \Longrightarrow \begin{pmatrix} 1 \\ \vdots \\ 1 \end{pmatrix}$$

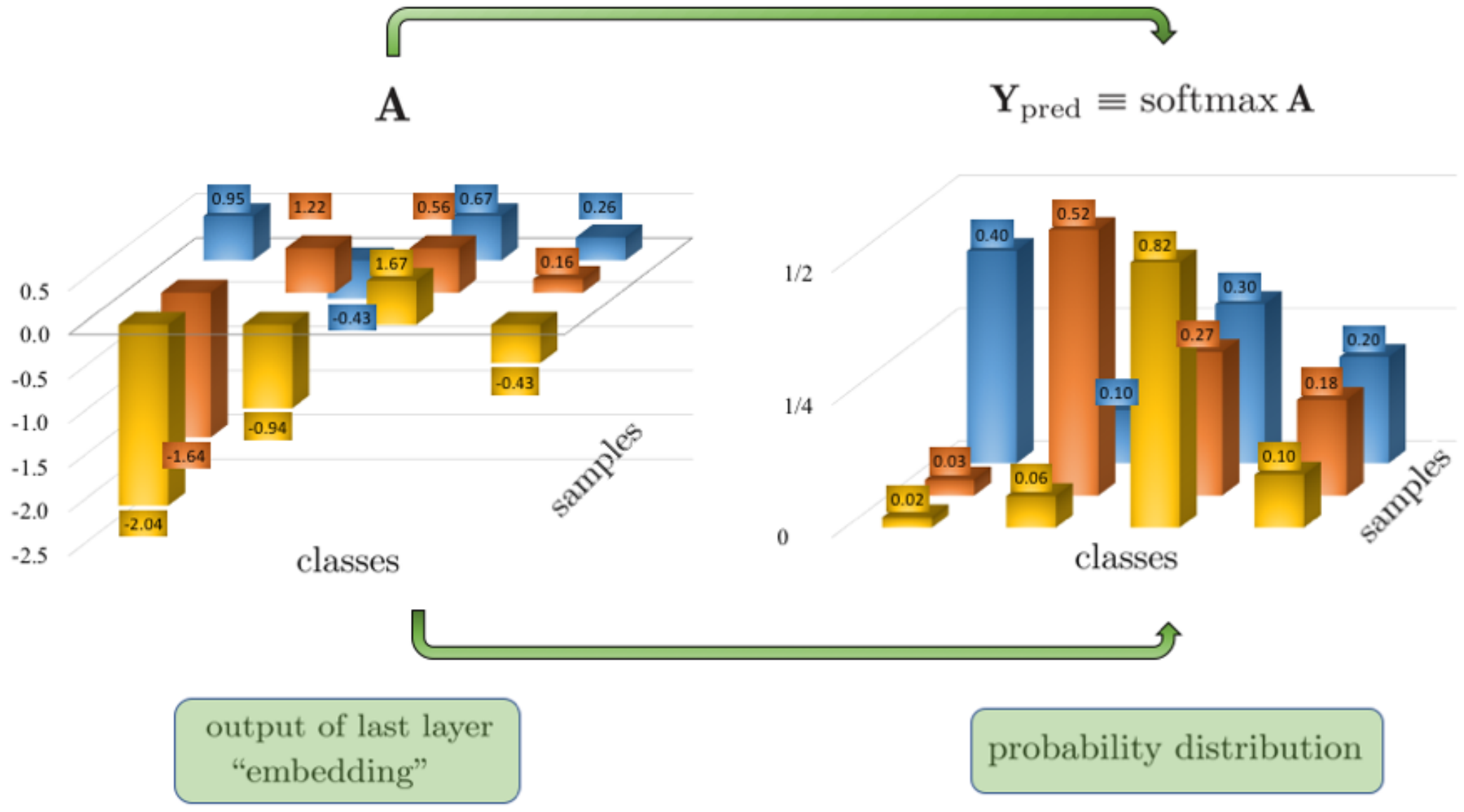

Figure 5: Illustration of the softmax operation. One can see that the samples of $\mathbf{A}$ (last representation layer also known as "embedding") are transformed into $\mathbf{Y}_{\text{pred}}$ whose vectors can be interpreted as probability values since $\sum_c \mathbf{y}^{\mathfrak{s}}_{\text{pred}} = 1$ is correctly normalized for all the samples $\mathfrak{s} \in \{1, \cdots, n\}$ of the training minibatch. Understandably, each component $y^{\mathfrak{s}}_{\text{pred}}(c) \sim \mathbb{R}$ of the prediction vector $\mathbf{y}^{\mathfrak{s}}_{\text{pred}} \sim \mathbb{R}^{n_c}$ quantifies the probability attributed to each class $c \in \{1, \cdots, n_c\}$ by the neural network $\mathcal{N}_{\mathcal{P}}$ for a specific sample $\mathfrak{s}$. (Note that the numerical values used in this illustration are shared with Fig. 2.)

**Small side note about probabilistic interpretation**

Although the softmax layer does indeed turn embeddings into probability distributions, it makes no guarantee that those probabilities reflect reliable confidence estimates. Perfect model calibration means that, given $N$ predictions each with confidence of $y_{\text{pred}}$ we expect exactly $y_{\text{pred}} N$ to be correctly classified. In practice, it turns out that many machine learning models are severely miscalibrated (even if they are accurate) and require post-processing scaling [57]. Generally speaking, the ability to control and estimate model uncertainty is of crucial importance in many applications such as security, medical, safety critical... Going beyond simple point estimates of probabilities in order to try and probe the full distribution of model uncertainty may be better addressed by tools coming from Bayesian machine learning [58, 59].

**Backward pass** As argued in section 2 and demonstrated by eq.(7), it is necessary to evaluate the total derivative of the softmax layer in order to kickstart the recursive error flow at the heart of the backpropagation algorithm. Explicit calculation of $\mathrm{d}\mathbf{Y}_{\text{pred}}$ leads to:

$$\mathrm{d}\mathbf{Y}_{\text{pred}} = \mathrm{d}\,(\text{softmax}\,\mathbf{A})$$

+

$$= \mathrm{d}\widetilde{\alpha} \circ e^{\mathbf{A}} + \mathbf{Y}_{\text{pred}} \circ \mathrm{d}\mathbf{A}$$

+

↓ replacing $\mathrm{d}\widetilde{\alpha}$ and using associativity of Hadamard product to substitute in the definition of $\mathbf{Y}_{\text{pred}}$

$$= \mathbf{Y}_{\text{pred}} \circ \left( \mathrm{d}\mathbf{A} - \widetilde{\mathbf{Y}_{\text{pred}} \ominus \mathrm{d}\mathbf{A}} \right) \tag{13}$$

# 4 Non-linear activation layer

Typically, the trainable layers of a neural network are linear maps (fully connected 5, locally connected i.e. convolutions 8, batch normalization 9...) between the input feature map $\mathbf{A}_{i-1}$ and the output $\mathbf{A}_i$. Since the functional composition of a set of linear transformations can always be reduced to another linear transformation, the overall architecture of a linear neural network could be collapsed onto itself effectively reducing its depth to a single layer. Although it is possible, in principle, to exploit peculiarities of floating-point arithmetic as an esoteric form of non-linearity [60], in practice all neural networks contain explicit non-linear activation layers. Their purpose is to "break" the architectures into distinct parts that can no longer be reduced thereby maintaining the concept of depth.

**Forward pass** Conventional activation layers consist of a simple element-wise application of a non-linear function to the input data as illustrated in Fig 6 and formalized by:

**Non-linear activation**: forward pass

$$\mathbf{A}_i = g\left(\mathbf{A}_{i-1}\right) \tag{14}$$

As established in section 2, backpropagation requires certain smoothness guarantees to be satisfied by the layers that compose the architecture of a neural network. In particular, this means that activation functions are constrained to belong to the class of differentiable functions. Although the literature is flush with an ever-growing number of implementations, it is the rectified linear unit (ReLU) [61] activation, defined as:

$$g(x) \equiv \text{ReLU}(x) = \begin{cases} x & x \geq 0 \\ 0 & x < 0 \end{cases}$$

that has taken prominence over other forms of non-linearities in recent years [7]. Because of its definition, the output feature map of a ReLU activation has a sparse representation where, statistically, half of the neurons of $\mathbf{A}_i$ are set to zero (those are said to be "not activated") while keeping the remaining neurons unchanged ("activated") as shown in the top panel of Fig.7.

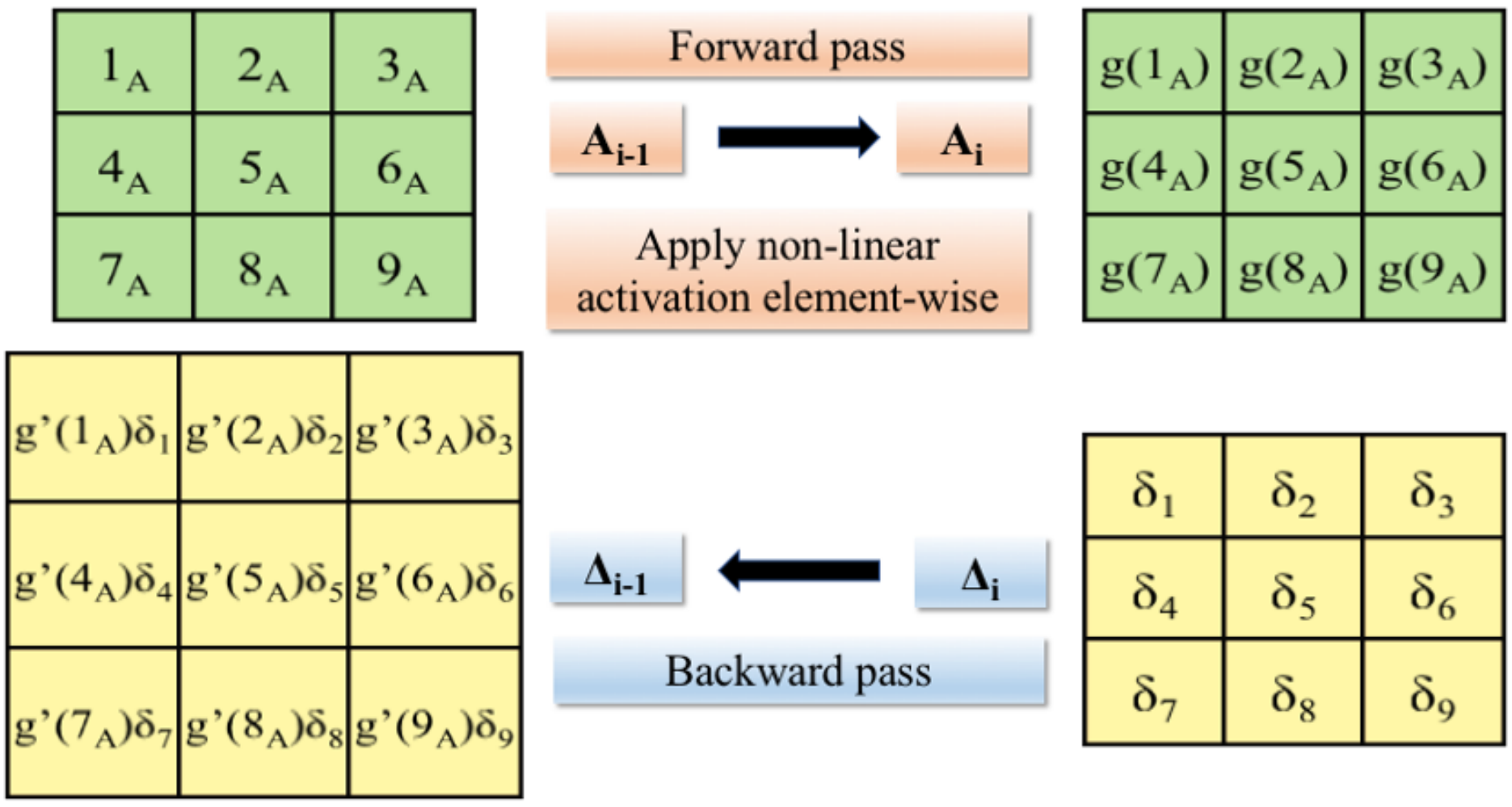

Figure 6: Illustration of the forward/backward passes through a general non-linear activation layer $g$. The forward pass simply consists of an element-wise application of $g$ as shown in eq.(14). The backward propagation of the error downstream to $\Delta_{i-1}$ is given by the Hadamard product between the upstream error $\Delta_i$ and the derivative $g'$ applied to the original input feature map $\mathbf{A}_{i-1}$, see eq.(16).

[7] Confusingly enough, ReLU is continuous everywhere but not differentiable at $x = 0$ making it a function only of class $C^0$ and not $C^1$ as required to ensure continuity of the first derivative. This jump discontinuity is dismissed by deep learning frameworks which ignore this exceptionally rare event by, for example, choosing to return one of the one-sided derivatives instead of raising an exception.

**Backward pass** Since activation layers don't contain trainable parameters, the backward pass only requires us to propagate the upstream error $\Delta_i$. As demonstrated in eq.(11), this is achieved via:

$$\Delta_i \cdot \mathrm{d}\mathbf{A}_i = \Delta_i \cdot \mathrm{d}g\left(\mathbf{A}_{i-1}\right)$$
$$+$$
$$= \underbrace{\Delta_i \circ g'\left(\mathbf{A}_{i-1}\right)}_{\Delta_{i-1}} \cdot \mathrm{d}\mathbf{A}_{i-1} \qquad (15)$$

In the case of ReLU activation, the derivative is given by the (binary) Heaviside function:

$$g'(x) = \begin{cases} 1 & x \geq 0 \\ 0 & x < 0 \end{cases}$$

meaning that neurons that were activated during the forward pass let the corresponding components of the upstream error $\Delta_i$ flow to $\Delta_{i-1}$ without any damping thanks to the constant derivative $g'(x \geq 0) = 1$ thereby helping fight the "vanishing gradient problem". On the other hand, non-activated neurons completely suppress the corresponding components of $\Delta_i$ through $g'(x < 0) = 0$ leading to a sparse representation for $\Delta_{i-1}$ which may act as a form of regularization. For comparison, activation functions that bound the error signal (such as $0 < g'(x) \leq 1/4 \ \forall x \in \mathbb{R}$ for the logistic function) expose the risk of vanishingly small $\Delta_{i-1}$ as one updates weights further and further away from the final layer (i.e. error values that have passed through many bounded activations as they make their way closer and closer to the input layer) leading to a lethargically slow learning process.

In summary, the backward pass of the activation layer consists in a simple propagation of the error given by:

**Non-linear activation**: backward pass

$$\Delta_{i-1} = \Delta_i \circ g'(\mathbf{A}_{i-1}) \qquad (16)$$

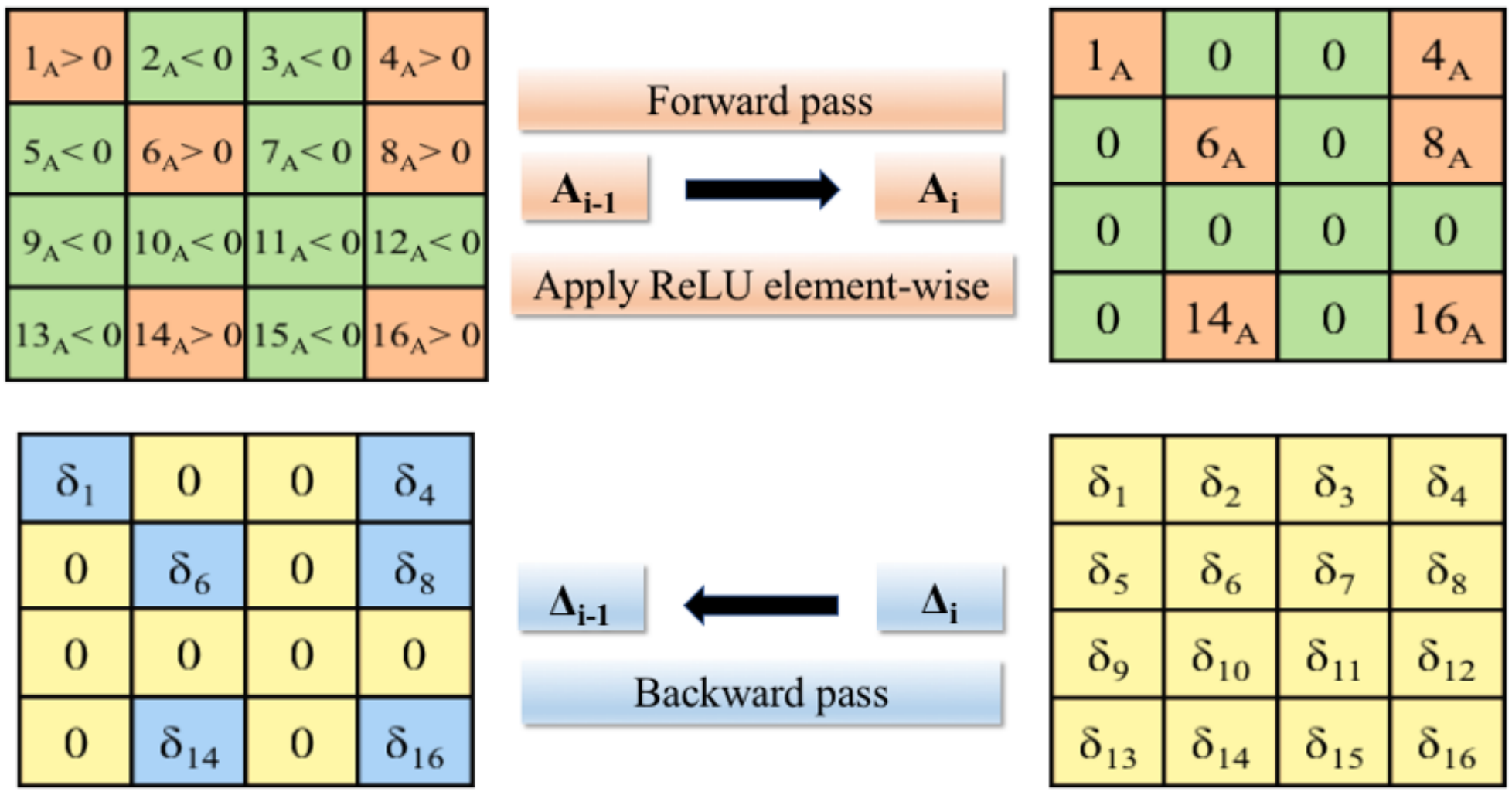

Figure 7: Illustration of the non-linear layer in the special case of the ReLU activation. The orange (resp. greeen) components of $\mathbf{A}_{i-1}$ denote positive (resp. negative) values. The presence of many 0's in $\mathbf{A}_i$ shows how the ReLU function leads to a sparse output during the forward pass. Similarly, the downstream error $\Delta_{i-1}$ also has a sparse representation (blue components). Error values corresponding to the "activated" positions during the forward pass are propagated unchanged from upstream $\Delta_i$ to $\Delta_{i-1}$ because of the identification of $g'$ with the sign function.

## 5 Fully connected layer

**Forward pass** This layer consists of a linear mapping between the input feature map $\mathbf{A}_{i-1} \sim \mathbb{R}^{n \times f_{i-1}}$ and the output $\mathbf{A}_i \sim \mathbb{R}^{n \times f_i}$. Without loss of generality, any linear function $\mathbb{R}^{f_{i-1}} \to \mathbb{R}^{f_i}$ between finite-dimensional vector spaces can be represented by a matrix $\sim \mathbb{R}^{f_{i-1} \times f_i}$. Allowing for the possibility of translation, fully connected layers are defined by a set of 2 trainable parameters:

$$\mathcal{P}_{i-1} \begin{cases} \mathbf{w}_{i-1} \sim \mathbb{R}^{f_{i-1} \times f_i} \,, & \text{“weights”} \\ \mathbf{b}_{i-1} \sim \mathbb{R}^{f_i} \,, & \text{“biases”} . \end{cases}$$

Their purpose is to discover a useful relationship transforming the $f_{i-1}$ input features into a new set of $f_i$ output features which are learned from the training data. The forward pass of this layer can be expressed as a generic affine transformation:

**Fully connected**: forward pass

$$\mathbf{A}_i = \mathbf{A}_{i-1} \mathbf{w}_{i-1} + \widetilde{\mathbf{b}_{i-1}} \tag{17}$$

as illustrated in the top panel of Fig.8. Note that the biases are broadcast $\mathbf{b}_{i-1} \to \widetilde{\mathbf{b}_{i-1}} \sim \mathbb{R}^{n \times f_i}$ over the $+$ operator in order to match the dimensionality of $\mathbf{A}_{i-1}\mathbf{w}_{i-1}$. The name “fully connected” stems from the “neurobiological” interpretation of the matrix multiplication where one can view every component, i.e. “neuron”, of $\mathbf{A}_{i-1}$ as being connected to every other component, i.e. “neuron”, of $\mathbf{A}_i$ via the relevant component of the weights $\mathbf{w}_{i-1}$, i.e. through “synapses”, as illustrated in Fig 9. Note that we have one bias term for each one of the $f_i$ features in the output space (i.e. one bias term per output neuron). Fully connected layers are at the heart of many machine learning algorithms (linear/logistic regression, linear SVM, multi-layer perceptron neural networks...).

**Backward pass** As explained in section 2, the backward pass is carried out by evaluating the recursive relation derived in eq.(11). Specializing $\mathbf{A}_i$ to the case of a fully connected layer and taking its Frobenius product with the upstream error $\Delta_i \sim \mathbb{R}^{n \times f_i}$, we find:

$$\begin{aligned} \Delta_i \cdot \mathrm{d}\mathbf{A}_i &= \Delta_i \cdot \mathrm{d}\left(\mathbf{A}_{i-1}\mathbf{w}_{i-1} + \widetilde{\mathbf{b}_{i-1}}\right) \\ &\boxed{+} \\ &= \underbrace{\mathbf{A}_{i-1}^t \Delta_i}_{\frac{\partial \mathcal{L}_{\text{batch}}}{\partial \mathbf{w}_{i-1}}} \cdot \mathrm{d}\mathbf{w}_{i-1} + \underbrace{\sum_{\text{samples}} \Delta_i}_{\frac{\partial \mathcal{L}_{\text{batch}}}{\partial \mathbf{b}_{i-1}}} \cdot \mathrm{d}\mathbf{b}_{i-1} + \underbrace{\Delta_i \mathbf{w}_{i-1}^t}_{\Delta_{i-1}} \cdot \mathrm{d}\mathbf{A}_{i-1} \end{aligned}$$

Identifying all the terms allows us to extract the components of the gradient with respect to parameters $\mathcal{P}_{i-1} = \{\mathbf{w}_{i-1}, \mathbf{b}_{i-1}\}$ as well as the downstream error $\Delta_{i-1}$. In particular, one can see that $\Delta_{i-1}$ is related to its upstream counterpart $\Delta_i$ in a very similar way as the data flow from $\mathbf{A}_{i-1}$ to $\mathbf{A}_i$ during the forward pass; namely a matrix multiplication using the same weight matrix $\mathbf{w}_{i-1}$ (up to a simple transpose) as illustrated in the bottom panel of Fig.8. Notice also that one needs to keep in memory the original data input $\mathbf{A}_{i-1}$ in order to compute the gradient components during the backward pass.

In summary, the backward pass through a fully connected layer is given by:

**Fully connected**: backward pass

$$\Delta_{i-1} = \Delta_i \mathbf{w}_{i-1}^t \qquad \sim \mathbb{R}^{n \times f_{i-1}} \tag{18}$$

$$\frac{\partial \mathcal{L}_{\text{batch}}}{\partial \mathbf{w}_{i-1}} = \mathbf{A}_{i-1}^t \Delta_i \qquad \sim \mathbb{R}^{f_{i-1} \times f_i} \tag{19}$$

$$\frac{\partial \mathcal{L}_{\text{batch}}}{\partial \mathbf{b}_{i-1}} = \sum_{\text{samples}} \Delta_i \qquad \sim \mathbb{R}^{f_i} \tag{20}$$

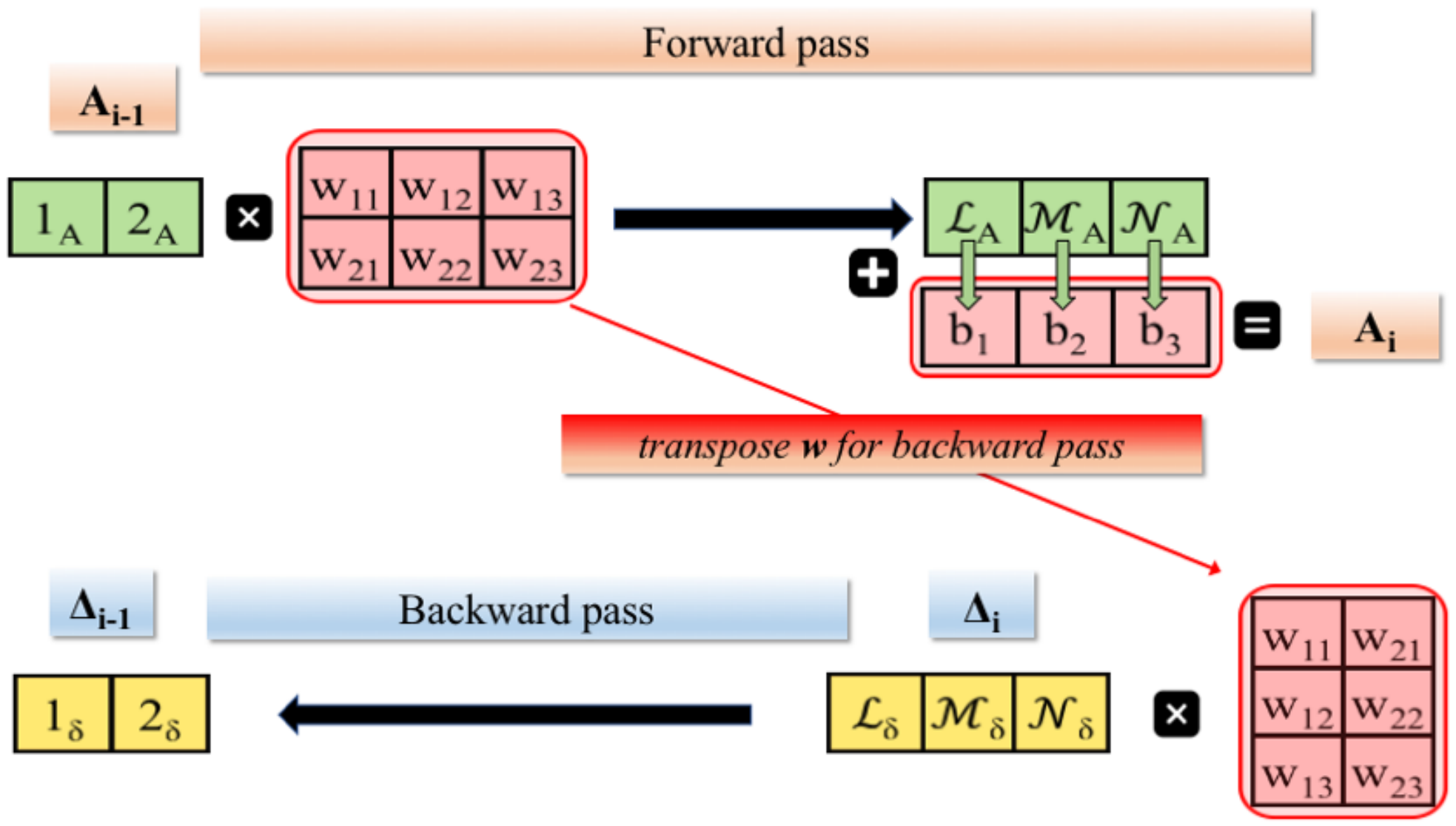

Figure 8: Basic illustration of a fully connected layer for a single sample $n = 1$ and low-dimensional feature maps $\{f_{i-1} = 2, f_i = 3\}$ showing the data flow from $\mathbf{A}_{i-1}$ to $\mathbf{A}_i$ during the forward pass (top panel) and the corresponding error flow from upstream $\Delta_i$ downstream to $\Delta_{i-1}$ during the backward pass (bottom panel). Explicit expansion of the matrix algebra is provided below in order to draw a clear parallel with the "neuron" interpretation displayed in the accompanying Fig.9.

**forward pass**

$$\begin{aligned}
\mathcal{L}_A &= (w_{11} \times 1_A) + (w_{21} \times 2_A) + b_1 \\
\mathcal{M}_A &= (w_{12} \times 1_A) + (w_{22} \times 2_A) + b_2 \\
\mathcal{N}_A &= (w_{13} \times 1_A) + (w_{23} \times 2_A) + b_3
\end{aligned}$$

**backward pass**

$$\begin{aligned}
1_\delta &= (w_{11} \times \mathcal{L}_\delta) + (w_{12} \times \mathcal{M}_\delta) + (w_{13} \times \mathcal{N}_\delta) \\
2_\delta &= (w_{21} \times \mathcal{L}_\delta) + (w_{22} \times \mathcal{M}_\delta) + (w_{23} \times \mathcal{N}_\delta)
\end{aligned}$$

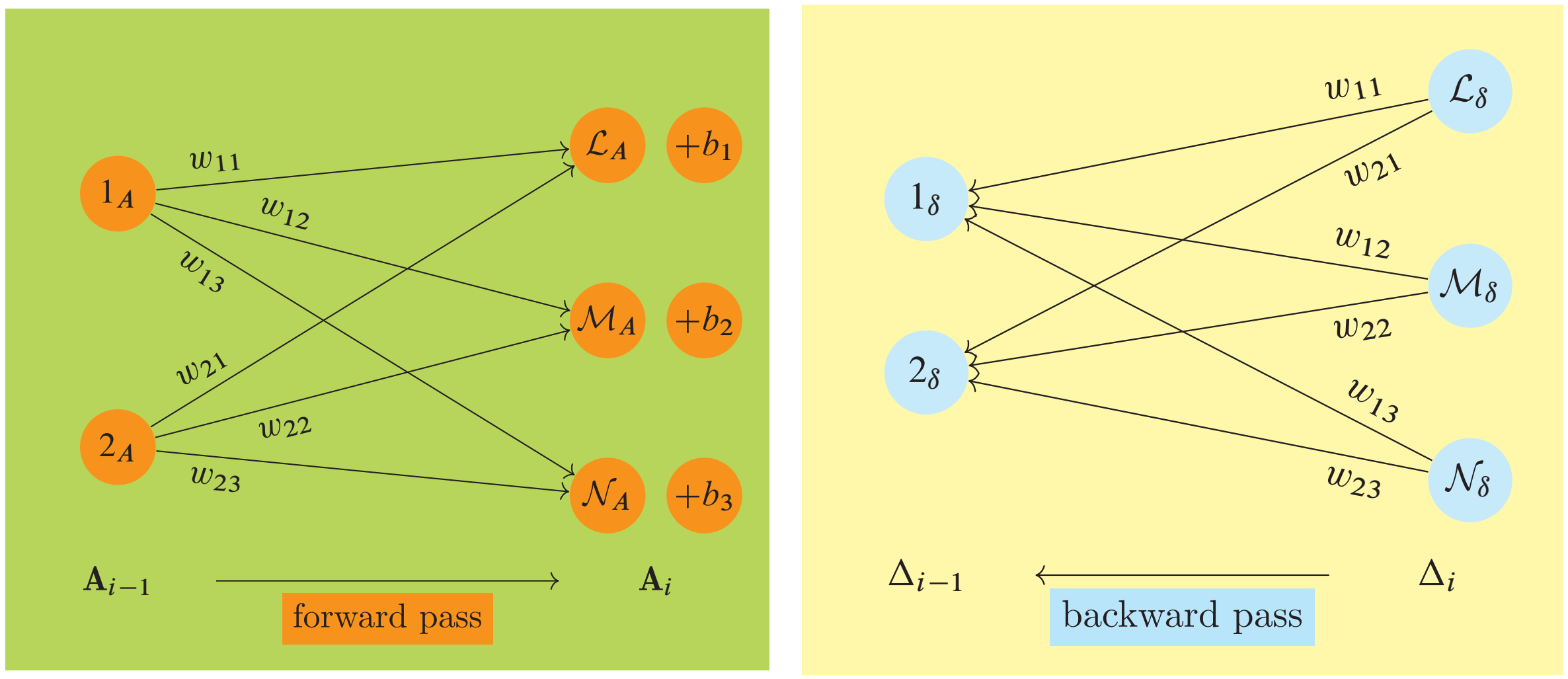

Figure 9: Basic illustration of a fully connected layer according to the "neurobiological" interpretation (same dimensionality as in the equivalent matrix formulation presented in Fig 8). During the forward pass, the "output neurons" $\{\mathcal{L}_A, \mathcal{M}_A, \mathcal{N}_A\}$ are connected to each one of the "input neurons" $\{1_A, 2_A\}$ through some weights $w_{ij}$ playing the role of "synapses". Similarly, the neurons of the downstream error $\{1_\delta, 2_\delta\}$ are connected to each one of the upstream error neurons $\{\mathcal{L}_\delta, \mathcal{M}_\delta, \mathcal{N}_\delta\}$ through the same weights $w_{ij}$ during the backward pass. One can verify that this interpretation of fully connected layers is in complete agreement with the explicit matrix algebra carried out above.

# 6 Maxpool layer

One purpose of pooling layers is to reduce (subsample) the spatial size of feature maps in an effort to make the neural network $\mathcal{N}_{\mathcal{P}}$ less compute/memory intensive. As an added benefit, removing features from the data may also help control overfitting by reducing the number of trainable parameters in subsequent layers of $\mathcal{N}_{\mathcal{P}}$. By far, the most common implementation of pooling layer is the so-called "maxpool" operation which, despite its long-standing popularity in CNN architectures, is not free from criticism [62]. Alternative forms of pooling (average, L2-norm, stochastic [63]...) along with suggestions to replace pooling layers altogether in favor of convolutional subsampling [64] can be found in the literature. Nevertheless, maxpool layers remain widely used by practitioners and are generally considered an important architectural component.

**Forward pass** As all sampling operations (described in appendix A), pooling layers are characterized by the geometrical parameter $\mathfrak{p} \equiv (k, s, p)$. Given an input data $\mathbf{A}_{i-1} \sim \mathbb{R}^{n \times d_{i-1} \times r_{i-1} \times r_{i-1}}$, the idea is to use $\mathfrak{p}$ to partition each one of $N_{\text{activation maps}} = n \times d_{i-1}$ depth and minibatch slices (so-called "activation maps"), individually of spatial size $\sim \mathbb{R}^{r_{i-1} \times r_{i-1}}$, into a set of smaller 2d data patches of kernel size $\mathbf{A}_{i-1}^{\text{patch}} \sim \mathbb{R}^{k \times k}$. The number of patches per activation map, $n_{\text{patches}}$, depends on the stride $s$, padding $p$ and kernel size $k$ defined by $\mathfrak{p}$ and its explicit form is given by eq.(45). Overall, this partitioning of $\mathbf{A}_{i-1}$ generates $N_{\text{patches}} = N_{\text{activation maps}} \times n_{\text{patches}}$ individual data patches.

*(Note that this is unlike convolution layers where one considers data partitions as sliding windows that include depth thereby leading to 3d patches of data $\sim \mathbb{R}^{d_{i-1} \times k \times k}$ as illustrated in Fig. 11 instead of purely spatial 2d patches $\mathbb{R}^{k \times k}$ independent of depth for pooling layers as illustrated in Fig. 10).*

Denoting by $g : \mathbb{R}^{k \times k} \to \mathbb{R}$ a generic pooling function (named so in analogy with activation layers), its role is to summarize the $k^2$ features present in a patch of input data into a single feature denoted $\mathbf{A}_i^{\text{patch}} \sim \mathbb{R} = g(\mathbf{A}_{i-1}^{\text{patch}} \sim \mathbb{R}^{k \times k})$ acting as a kind of local descriptive statistics. Applying $g$ to all the patches of $\mathbf{A}_{i-1}$ defined by $\mathfrak{p}$ allows us to build the output feature map:

**Pooling layer**: forward pass

$$\mathbf{A}_i = \diamondsuit \left\{ g(\mathbf{A}_{i-1}^{\text{patch}}) \mid \text{patch} \in \text{patches of } \mathbf{A}_{i-1} \text{ defined by } \mathfrak{p} \right\} \tag{21}$$

where $\diamondsuit$ stands for the reconstruction of the set $\{\mathbf{A}_i^{\text{patch}} \sim \mathbb{R}\}$ of cardinality $N_{\text{patches}}$ back into a 4d data array of dimensionality consistent with $\mathbf{A}_i$ (minibatch and depth stack of spatial 2d activation maps): i) since pooling layers operate independently of depth, $N_{\text{activation maps}}$ remains unchanged from $\mathbf{A}_{i-1}$ to $\mathbf{A}_i$ and, as a result, depth is conserved i.e. $d_i \equiv d_{i-1}$ ii) the output patch values $\mathbf{A}_i^{\text{patch}}$ are grouped together by their respective activation maps and aggregated into 2d grids of spatial size $\sim \mathbb{R}^{r_i \times r_i}$ where we identify the output size $r_i \equiv \sqrt{n_{\text{patches}}} < r_{i-1}$, finally iii) the complete activation maps are stacked together in order to recover the correct dimensionality $\mathbf{A}_i \sim \mathbb{R}^{n \times d_i \times r_i \times r_i}$.

For maxpool layers, the pooling function is implemented simply as the maximum function $g \equiv \max$. In this case, $\mathbf{A}_i^{\text{patch}}$ corresponds to the value of the largest feature among the $k^2$ components of $\mathbf{A}_{i-1}^{\text{patch}}$. The idea is that since we only care about the maximum value within some spatial neighborhood of size $\sim \mathbb{R}^{k \times k}$, as long as small translations do not remove the largest feature from a pooling region (or bring in another even larger feature into the pooling region) the output representation $\mathbf{A}_i$ is unaffected. As a consequence, maxpool layers are believed to introduce a basic form of translational invariance. In addition to the formalism developed above, the mechanics of maxpool layers is readily understandable as a geometrical procedure illustrated in Fig. 10.

In passing, let us mention that assuming i) the patches defined by $\mathfrak{p}$ do not overlap: specializing eq.(45) leads to $r_i = r_{i-1}/k$ and ii) maxpool layers are the only source of downsampling in the network (often the case since it is common to use the so-called "same padding" in convolution layers), one can see that a network with $n_{\text{pool}}$ maxpool layers experiences an exponential decrease in the size of its final output feature map $r_{\text{final}} = r_{\text{initial}}/k^{n_{\text{pool}}}$. For this reason, it is recommended to adopt less destructive pooling in the form of small kernels with $\mathfrak{p} = (k = 2, s = 2, p = 0)$ being a popular choice illustrated in Fig. 10; an option that nevertheless discards 75% of the input features at each maxpool layer.

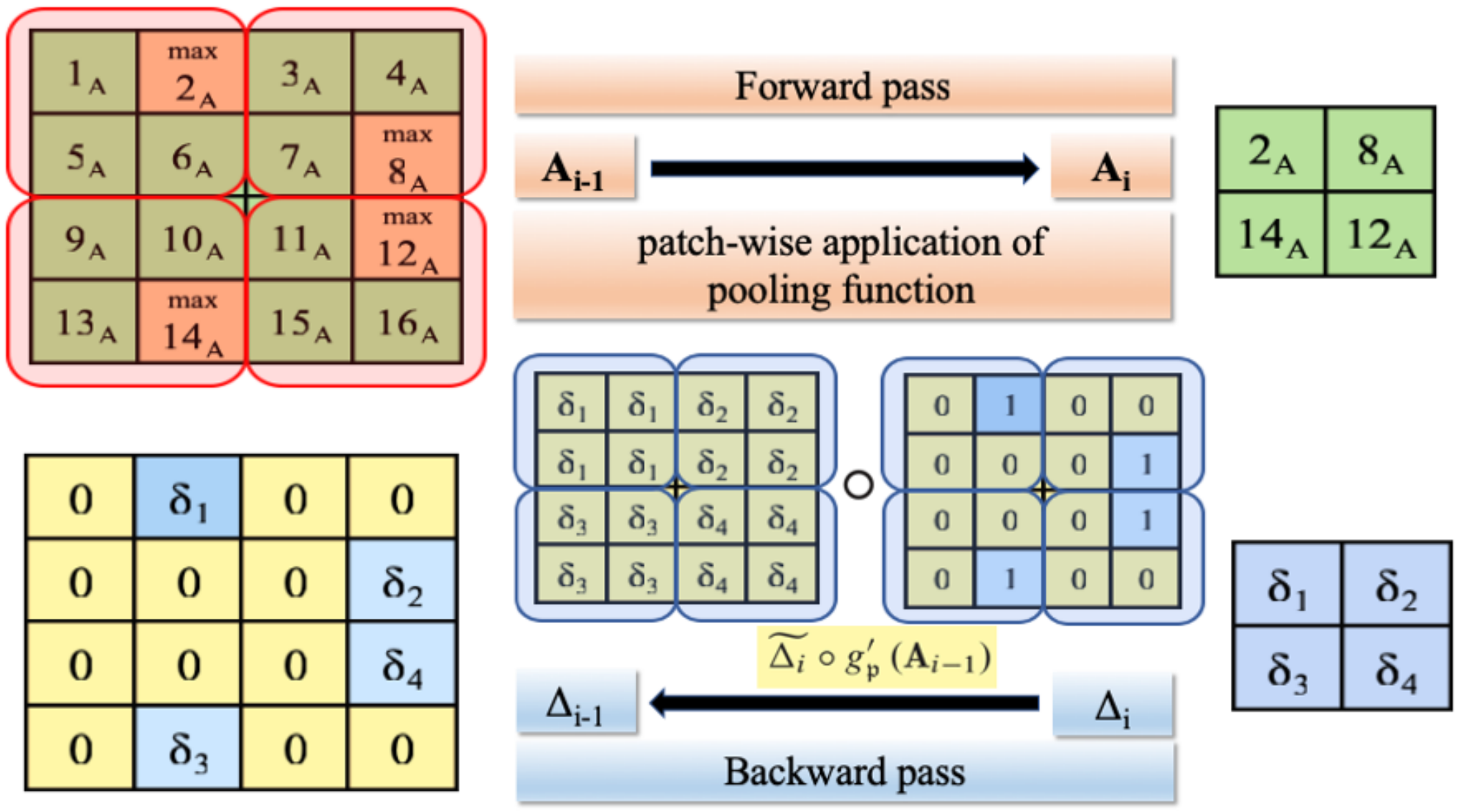

Figure 10: Illustration of the maxpool layer with $\mathfrak{p} = (k = 2, s = 2, p = 0)$ on a small activation map of spatial size $r_{i-1} = 4$. Because pooling layers operate on activation maps independently of depth, it is enough to consider a single activation map. **Forward pass:** Applying the partitioning defined by $\mathfrak{p}$ on $\mathbf{A}_{i-1}$ leads to 4 patches (transparent red squares). The maximum value within each patch (arbitrary cells colored in orange) is returned by the pooling function $g \equiv \max$ thereby building the output $\mathbf{A}_i$ with $r_i = 2$. **Backward pass:** Downstream propagation of the error to $\Delta_{i-1}$ is given by the Hadamard product between the upstream error $\Delta_i$ (up to a patch-wise upsampling broadcast $r_i = 2 \to r_{i-1} = 4$) and the patch-level evaluation of $g'_{\mathfrak{p}}(\mathbf{A}_{i-1})$. In both cases, the patches are defined according to $\mathfrak{p}$ and visualized as transparent blue squares. Accordingly, the backward pass can be implemented in a memory friendly way by routing the components of $\Delta_i$ to the patch location that contributed to $\mathbf{A}_i$ during the forward pass (cells colored in blue) and ignoring others (since their associated derivatives are 0).

**Backward pass** As understood from the forward pass, pooling layers are similar to the non-linear activation layers of section 4. The main difference stems from the fact that while activation functions operate element-wise on the input data (thereby preserving the dimensionality of activation maps), pooling layers are applied at the level of larger input data patches defined by $\mathfrak{p}$ (thereby reducing the spatial size of activation maps). As such, we can combine the error propagation expression previously derived in eq.(15):

$$\Delta_i \cdot \mathrm{d}\mathbf{A}_i = \underbrace{\widetilde{\Delta_i} \circ g'_{\mathfrak{p}}(\mathbf{A}_{i-1})}_{\Delta_{i-1}} \cdot \mathrm{d}\mathbf{A}_{i-1}$$

together with a patch-level evaluation of the derivative of the pooling function $g'_{\mathfrak{p}}(\mathbf{A}_{i-1})$ using the same construction $\Diamond$ that was defined in the forward pass. Note that the upstream error $\Delta_i$ is also broadcast patch-wise in order to match the dimensionality of the original input data $\Delta_i \to \widetilde{\Delta_i} \sim \mathbf{A}_{i-1}$.

**Pooling layer**: backward pass

$$\Delta_{i-1} = \widetilde{\Delta_i} \circ g'_{\mathfrak{p}}(\mathbf{A}_{i-1})$$
$$g'_{\mathfrak{p}}(\mathbf{A}_{i-1}) \equiv \Diamond \left\{ g'(\mathbf{A}_{i-1}^{\text{patch}}) \mid \text{patch} \in \text{patches of } \mathbf{A}_{i-1} \text{ defined by } \mathfrak{p} \right\} \quad (22)$$

In the special case of maxpool activation, the derivative of $g \equiv \max$ with respect to the $k^2$ elements present in a patch $\mathbf{A}_{i-1}^{\text{patch}}$ is given by:

$$g'(\mathbf{A}_{i-1}^{\text{patch}}) \equiv \left\{ \frac{\mathrm{d}\max(a_1, \cdots, a_{k^2})}{\mathrm{d}a_j} \mid a_j \in \mathbf{A}_{i-1}^{\text{patch}} \right\} = \begin{cases} 1 & \text{for the maximum value} \\ 0 & \text{otherwise} \end{cases}$$

This can be implemented in a memory efficient way by caching the position of the maxima within the patches defined by $\mathfrak{p}$ during the forward pass instead of the full $\mathbf{A}_{i-1}$ as shown in Fig 10.

## 7 Flatten layer

In essence, this layer is nothing more than a non-parametric "plumbing" transformation whose sole purpose is to reshape the input data $\mathbf{A}_{i-1}$ during the forward pass and the upstream error $\Delta_i$ during the backward pass in a manner that is appropriate for processing by subsequent layers. Note that in both cases, there is no change in information content besides trivial geometrical re-organizations.

**Forward pass** In typical scenarios (as exemplified in the case study network of table 1), the input data consists of a minibatch $\mathbf{A}_{i-1} \sim \mathbb{R}^{n\times d_{i-1}\times r_{i-1}\times r_{i-1}}$ where each 3d sample (image) is parametrized in terms of meaningful depth and space dimensions. The target is to flatten those dimensions into a 1d feature vector: $\mathbb{R}^{d_{i-1}\times r_{i-1}\times r_{i-1}} \rightarrow \mathbb{R}^{(d_{i-1}\times r_{i-1}\times r_{i-1})} \equiv \mathbb{R}^{f_i}$ so that the minibatch is ultimately represented as a 2d data array $\mathbf{A}_i \sim \mathbb{R}^{n\times f_i}$.

**Flatten**: forward pass

$$\mathbf{A}_i \sim \mathbb{R}^{n\times f_i} = \text{flatten}\left(\mathbf{A}_{i-1} \sim \mathbb{R}^{n\times d_{i-1}\times r_{i-1}\times r_{i-1}}\right) \tag{23}$$

**Backward pass** In this case, the input consists of a 2d upstream error $\Delta_i \sim \mathbb{R}^{n\times f_i}$ and the objective is to restore depth and space as meaningful independent dimensions back in to the downstream error $\Delta_{i-1} \sim \mathbb{R}^{n\times d_{i-1}\times r_{i-1}\times r_{i-1}}$. We formally denote by "fold": $\mathbb{R}^{f_i} \rightarrow \mathbb{R}^{d_{i-1}\times r_{i-1}\times r_{i-1}}$, this inverse of the flattening operation. Applying the general recursion relation eq.(11), we get:

$$\Delta_i \cdot \mathrm{d}\mathbf{A}_i = \Delta_i \cdot \mathrm{d}\left(\text{flatten}\,\mathbf{A}_{i-1}\right) \rightarrow \underbrace{\text{fold}\,\Delta_i}_{\Delta_{i-1}} \cdot \mathrm{d}\mathbf{A}_{i-1}$$

**Flatten**: backward pass

$$\Delta_{i-1} \sim \mathbb{R}^{n\times d_{i-1}\times r_{i-1}\times r_{i-1}} = \text{fold}\left(\Delta_i \sim \mathbb{R}^{n\times f_i}\right) \tag{24}$$

## 8 Convolution layer

**Forward pass** Unlike fully connected layers where samples are represented as a 1d feature vectors (with input $\sim \mathbb{R}^{f_{i-1}}$ and output $\sim \mathbb{R}^{f_i}$; see section 5), we are now considering data arrays (images) that have a 3d structure composed of independent depth and space dimensions as introduced in section 1 and illustrated in Fig.11. Otherwise and similarly to fully connected layers, a convolution layer is another kind of linear mapping between the input data $\mathbf{A}_{i-1} \sim \mathbb{R}^{n\times d_{i-1}\times r_{i-1}\times r_{i-1}}$ and the corresponding output $\mathbf{A}_i \sim \mathbb{R}^{n\times d_i\times r_i\times r_i}$ parametrized by:

$$\mathcal{P}_{i-1} \begin{cases} \mathbf{w}^{\mathfrak{p}}_{i-1} \sim \mathbb{R}^{d_i\times d_{i-1}\times k\times k}\,, & \text{"weights" / convolutional kernels} \\ \mathbf{b}_{i-1} \sim \mathbb{R}^{d_i}\,, & \text{"biases".} \end{cases}$$

The kernels can be seen as a collection of $d_i$ independent 3d "neurons" of dimensionality $\sim \mathbb{R}^{d_{i-1}\times k\times k}$. Stacking all the individual kernels together leads to a 4d structure for $\mathbf{w}^{\mathfrak{p}}_{i-1}$. Like the biases of fully connected layers, each one of $d_i$ output convolutional neurons gets their own bias term $\mathbf{b}_{i-1} \sim \mathbb{R}^{d_i}$. Accordingly, the forward pass of convolution layers is expressed as an affine transformation:

**Convolution**: forward pass

$$\mathbf{A}_i = \mathbf{w}^{\mathfrak{p}}_{i-1} \star \mathbf{A}_{i-1} + \widetilde{\mathbf{b}_{i-1}} \tag{25}$$

where the biases are appropriately broadcast $\mathbf{b}_{i-1} \rightarrow \widetilde{\mathbf{b}_{i-1}} \sim \mathbb{R}^{n\times d_i\times r_i\times r_i}$ (see appendix B). The crucial difference between straightforward matrix multiplication (fully connected layer) and convolution as symbolized by $\star$ comes from the fact that the geometrical factor $\mathfrak{p} = (k, s, p)$ imposes a spatial local-connectivity pattern between the weights and the data with which they are being convolved. This is unlike fully connected layers where each neuron is connected to all of the input features. In other words, convolutional neurons are only connected to a local spatial neighborhood of the data.

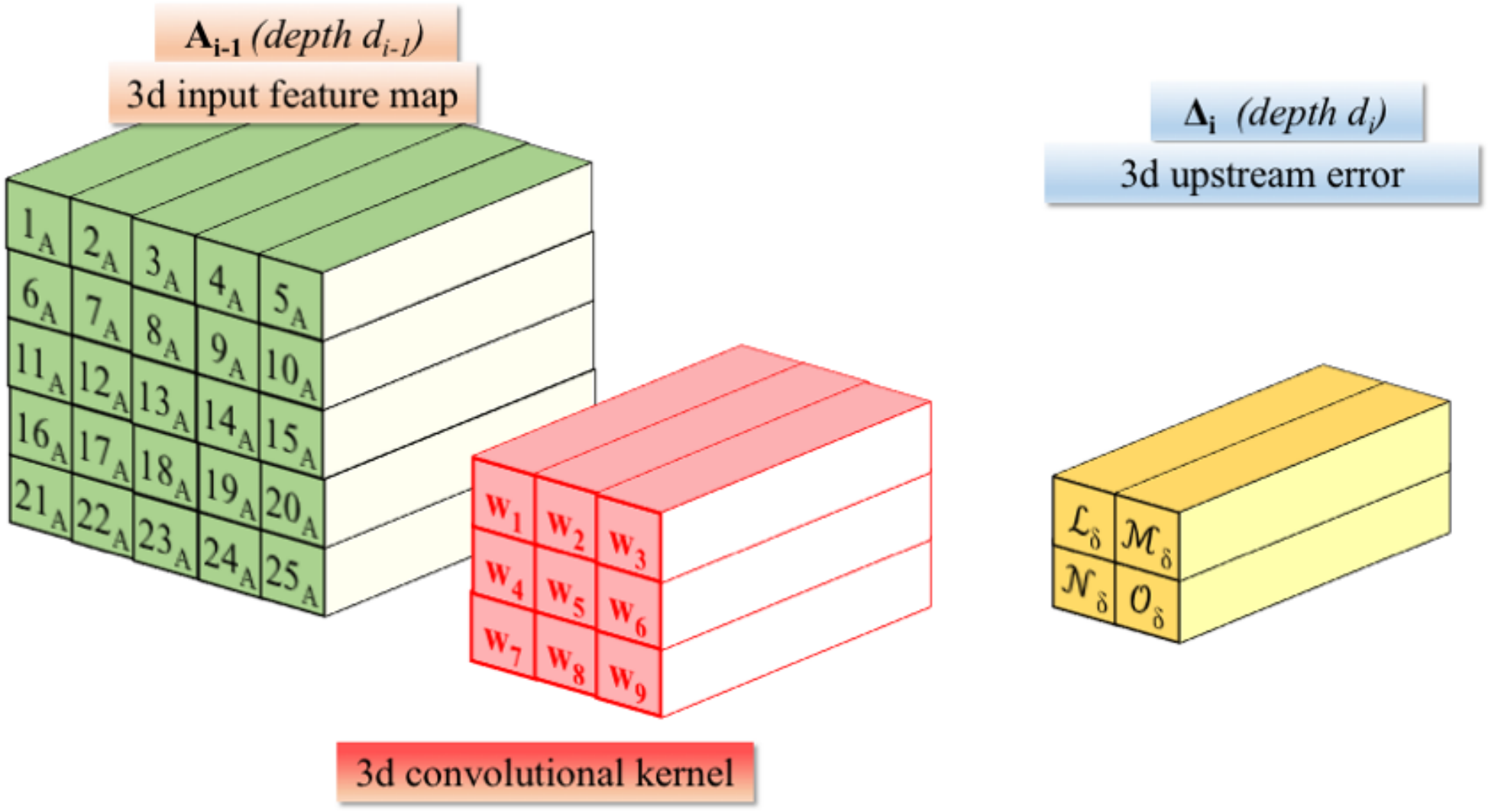

Figure 11: Illustration of the general 3d nature of the data (**with animations**). Complete minibatches of size $n > 1$ are built by stacking 3d samples into 4d structures as discussed in section 1. The purpose of this figure is only to emphasize the fact that the planar representations of Fig. 12 (where the mechanics of convolution layers is explained in detail) correspond to top views of the actual underlying 3d data structures. For example, it should be clear that individual cells in Fig. 12 actually correspond to 1d vectors $\sim \mathbb{R}^d$ across the depth. Therefore, generic products between individual cells of $\mathbf{A}_{i-1}/\Delta_i$ with individual cells of the kernel, as performed in Fig 12, really stand for complete dot-products $\mathbb{R}^d \cdot \mathbb{R}^d \sim \mathbb{R}$. Note that the same kernel is used both with $\mathbf{A}_{i-1}$ (data propagation during the forward pass) and $\Delta_i$ (error propagation during the backward pass) modulo simple geometrical transformations discussed in the text. For example, the depth dimension $d$ is transposed from $\mathbb{R}^{d_{i-1}}$ in the forward pass to $\mathbb{R}^{d_i}$ in the backward pass making the dot-products discussed above well-defined. Even though we are displaying here only a single convolutional kernel, real convolution layers may be made up of a large number of such filters by stacking individual 3d filters into 4d banks of filters. The animation of the upstream error $\Delta_i$ shows how **fractionally strided convolutions** are implemented by a combination of both internal (gray) and external (white) padding; more details in the main text and in Fig.12.

A natural way to look at convolutions consists in imagining that the weights are sliding across different data windows of $\mathbf{A}_{i-1}$: Spatially small, $k < r_{i-1}$, patches of 3d data $\sim \mathbb{R}^{d_{i-1} \times k \times k}$ are paired via tensor contraction with individual 3d kernels of matching dimensionality $\sim \mathbb{R}^{d_{i-1} \times k \times k}$ in order to produce, step-by-step, the scalar components of the output feature map $\mathbf{A}_i$ as illustrated in detail in Fig.12. Note that the filters remain fully connected to the data depth-wise as one can see that $\mathbf{w}_{i-1}^{\mathfrak{p}}$ adopts the same value $d_{i-1}$ along this dimension to match the depth of $\mathbf{A}_{i-1}$. Therefore, $d_{i-1}$ is contracted away by the depth-wise dot-products that build up $\mathbf{A}_i$ and choosing the number $d_i$ of filters prescribes the depth of the output. The precise way in which the sliding windows are constructed is defined by the sampling triplet $\mathfrak{p} = (k, s, p)$ described in appendix A and determines the spatial resolution $r_i$ of the output. One side-effect of this weight-sharing property of $\mathbf{w}_{i-1}^{\mathfrak{p}}$ across different patches of $\mathbf{A}_{i-1}$ is the emergence of a basic (and somewhat debatable [65]) form of translational invariance. In addition, weight-sharing leads to a dramatic decrease in the number of trainable parameters compared to fully connected layers.

Besides this "**sliding-window**" approach, there are many ways to implement convolutions and fast algorithms are still a topic of research since convolutions are the most computationally-intensive layer (Winograd [66], frequency domain [67]...). In this paper, we focus on a very common implementation that uses the so-called "**im2col**" transformation. This intuitive data manipulation scheme [68, 69] allows to re-express the convolution operator $\star$ in terms of a single matrix multiplication that can be handled by existing and highly optimized numerical routines [70, 71]. It turns out that the benefits of expressing convolutions as GEMMs (GEneral Matrix to Matrix Multiplication) generally outweigh the additional memory requirements and this approach is widely used in modern deep learning frameworks.

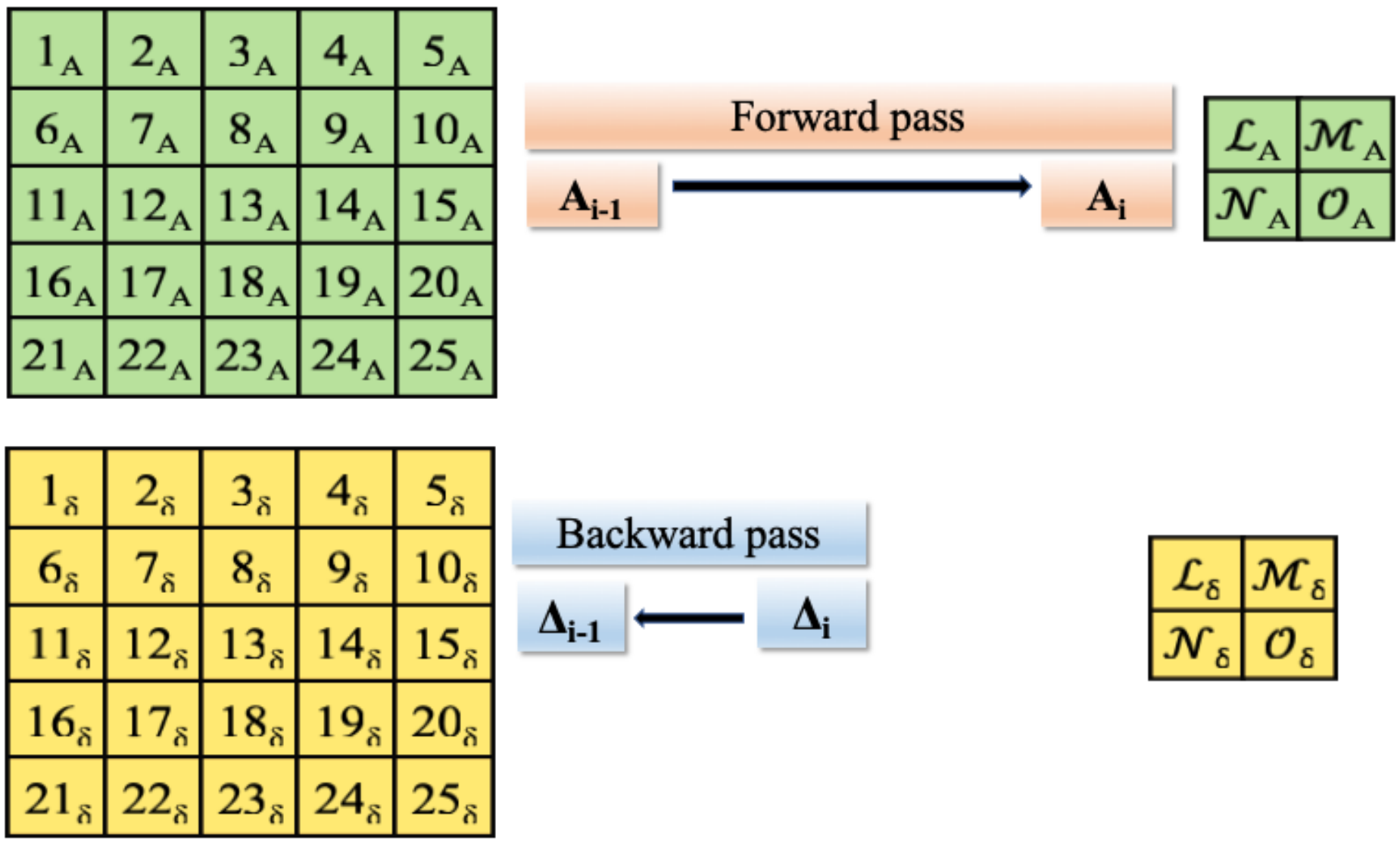

Figure 12: Convolution layer (**with animations**): the "sliding window" approach. For clarity, we consider **top view** visualizations of a single 3d data sample and a single 3d convolutional kernel. The 2d projections of these data structures as squares when viewed from the top should not distract from their true underlying 3d nature: each cell actually represents a 1d vector across the depth dimension as illustrated in Fig.11. The (downsampling) **forward pass** can be thought of as sliding the 3d kernel $\mathbf{w}_{i-1}^{\mathfrak{p}} \sim \mathbb{R}^{d_{i-1} \times k \times k}$ across a set of 3d patches of matching dimensionality $\sim \mathbb{R}^{d_{i-1} \times k \times k}$, depicted as red squares, which partition the input data $\mathbf{A}_{i-1} \sim \mathbb{R}^{d_{i-1} \times r_{i-1} \times r_{i-1}}$. The number of patches is given by eq.(45) and prescribes the spatial resolution $r_i$ of the output activation map $\mathbf{A}_i \sim \mathbb{R}^{r_i \times r_i}$ (note that $d_{i-1}$ has been contracted away since convolutional kernels are fully connected to the input data along the depth dimension). The geometric properties of these patches are characterized (in this example) by the triplet $\mathfrak{p}_{\text{forward}} = (k = 3, s = 2, p = 0)$ and we provide an explicit decomposition of the spatial local connectivity pattern in table 14 as well as its implementation using the im2col transformation in Fig 13. Each component $\{\mathcal{L}_A \sim \mathbb{R}, \mathcal{M}_A \sim \mathbb{R}, \mathcal{N}_A \sim \mathbb{R}, \mathcal{O}_A \sim \mathbb{R}\}$ of the output activation map is determined by the sum of the dot-products (tensor contraction) between the entries $\sim \mathbb{R}^{d_{i-1}}$ of $\mathbf{w}_{i-1}^{\mathfrak{p}}$ and each individual cell $\sim \mathbb{R}^{d_{i-1}}$ of the relevant input data patch (members of the sliding red squares). Obviously, realistic convolution layers are composed of a number $d_i > 1$ of filters and this sliding window approach should be repeated independently for all the filters in order to yield the 3d output feature map $\mathbf{A}_i \sim \mathbb{R}^{d_i \times r_i \times r_i}$. The (upsampling) **backward pass** propagates the upstream error $\Delta_i \sim \mathbb{R}^{d_i \times r_i \times r_i}$ downstream and recovers the dimensionality of the input data $\Delta_{i-1} \sim \mathbf{A}_{i-1}$. This is accomplished by another convolution that uses the same sliding window approach with the same kernel $\mathbf{w}_{i-1}^{\curvearrowleft t_{\text{d}}\, \mathfrak{p}'} \sim \mathbb{R}^{d_i \times k \times k}$ modulo simple geometric transformations such as depth transpose $t_{\text{d}}$ from $d_{i-1}$ to $d_i$ and 180-degree rotation of the weights as explained in the main text. In order to preserve the connectivity pattern that was established during the forward pass, the patches partitioning $\Delta_i$ are now defined by the sampling triplet $\mathfrak{p}' \equiv \mathfrak{p}_{\text{backward}} = (k = 3, s = 1, p = 2)$ coupled with an internal padding on $\Delta_i$ (gray colored cells) effectively mimicking a slow-moving fractional stride $s_{\text{backward}} \sim 1/2$ in inverse proportion to $s_{\text{forward}} = 2$. Indeed, table 14 explicitly verifies that this upsampling convolution shares an identical local connectivity pattern for both forward/backward passes.

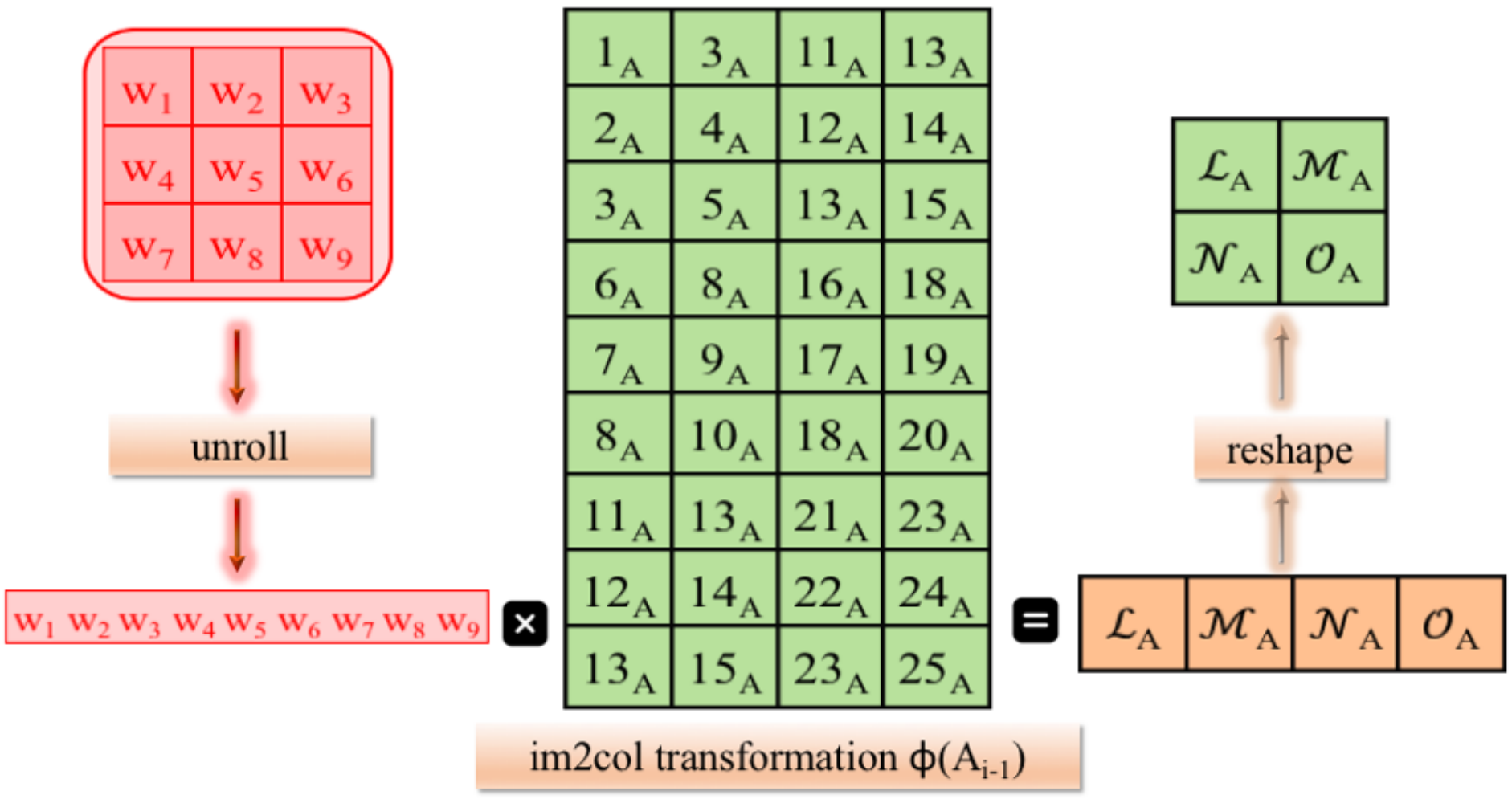

Figure 13: Convolution layer: implementation as a matrix multiplication via **im2col transformation** as shown in eq.(28). For consistency, we use the same notation as the example already discussed in Fig. 12 involving a single data sample ($n = 1$), single convolutional kernel ($d_i = 1$) and the sampling parameter $\mathfrak{p} = (k = 3, s = 2, p = 0)$. Note that this is again a top view (see Fig.11 for the underlying 3d data structures) so that cells actually represent 1d vectors of depth data $\sim \mathbb{R}^{d_{i-1}}$ and products between individual cells correspond to vector dot-products $\sim \mathbb{R}^{d_{i-1}} \cdot \mathbb{R}^{d_{i-1}} \sim \mathbb{R}$ across the depth dimension. First, the kernel is unrolled into an array $\overline{\mathbf{w}}^{\mathfrak{p}}_{i-1}$ of $9d_{i-1}$ components ($k = 3$); see eq.(26). Second, the input data $\mathbf{A}_{i-1}$ is partitioned into 4 patches, according to $\mathfrak{p}$, and the $9d_{i-1}$ components of each patch (members of the sliding red squares in Fig.12) are stacked as matrix columns building up the transformed $\phi(\mathbf{A}_{i-1})$. One can verify that the matrix multiplication $\overline{\mathbf{w}}^{\mathfrak{p}}_{i-1}\,\phi(\mathbf{A}_{i-1})$ of eq.(28) indeed leads to the sequence and sum of dot-products that defined $\{\mathcal{L}_A, \mathcal{M}_A, \mathcal{N}_A, \mathcal{O}_A\}$ in the traditional "sliding window approach" illustrated in Fig.12. Note that the output of this GEMM is appropriately reshaped by $f_{4d}$ in order to restore $\mathbf{A}_i$ with the spatial resolution $r_i = 2$ as prescribed by $\mathfrak{p}$.

**Forward pass (GEMM formulation)** The idea is to perform a series of geometric transformations on the weights $\mathbf{w}^{\mathfrak{p}}_{i-1}$ and the input data $\mathbf{A}_{i-1}$ so as to turn both of them into 2d arrays in order to express their convolution as a GEMM. The first step consists in unrolling the depth ($d_{i-1}$) and space ($k \times k$) dimensions of the $d_i$ kernels into a 1d vector as such:

$$\mathbf{w}^{\mathfrak{p}}_{i-1} \sim \mathbb{R}^{d_i \times d_{i-1} \times k \times k} \rightarrow \overline{\mathbf{w}}^{\mathfrak{p}}_{i-1} \sim \mathbb{R}^{d_i \times (d_{i-1} \times k \times k)} \tag{26}$$

This unrolling operation is illustrated in Fig.13 and can be reversed by rolling back $\overline{\mathbf{w}}^{\mathfrak{p}}_{i-1}$ to restore the depth and space dimensions to the $d_i$ kernels and recover the original 4d structure of $\mathbf{w}^{\mathfrak{p}}_{i-1}$:

$$\text{roll}\left(\overline{\mathbf{w}}^{\mathfrak{p}}_{i-1}\right) \longrightarrow \mathbb{R}^{d_i \times d_{i-1} \times k \times k} \tag{27}$$

Next, the input feature map $\mathbf{A}_{i-1}$ is also re-arranged as a $2d$ array $\mathbf{A}_{i-1} \rightarrow \phi(\mathbf{A}_{i-1})$ using the im2col transformation $\phi$. Keeping in line with the sliding-window approach (as shown in Fig.12) where input data patches are paired with convolutional kernels, we see that the columns of $\phi(\mathbf{A}_{i-1})$ should correspond to the components of the data patches that partition $\mathbf{A}_{i-1}$. The number of patches is determined by $\mathfrak{p}$ through eq.(45) and defines the spatial resolution $r_i$ of the output $\mathbf{A}_i$. With a minibatch size $n$, the matrix $\phi(\mathbf{A}_{i-1})$ therefore has $(r_i \times r_i \times n)$ columns, each of which composed of the particular $(d_{i-1} \times k \times k)$ components of $\mathbf{A}_{i-1}$ present in a specific patch. Dimensionally, the transformation leads to:

$$\mathbf{A}_{i-1} \sim \mathbb{R}^{n \times d_{i-1} \times k \times k} \rightarrow \phi(\mathbf{A}_{i-1}) \sim \mathbb{R}^{(d_{i-1} \times k \times k) \times (r_i \times r_i \times n)}$$

As shown in Fig.13, a simple matrix multiplication between the unrolled weights $\overline{\mathbf{w}}^{\mathfrak{p}}_{i-1}$ and $\phi(\mathbf{A}_{i-1})$ is indeed identical to the tensor contraction discussed in the previous paragraph and illustrated in Fig.12. Putting it all together, the forward pass of the convolution layer can be formulated as:

**Convolution**: forward pass (GEMM formulation)

$$\mathbf{A}_i = f_{4d}\left[\overline{\mathbf{w}}^{\mathfrak{p}}_{i-1}\,\phi(\mathbf{A}_{i-1})\right] + \widetilde{\mathbf{b}_{i-1}} \tag{28}$$

Note that there is a final geometric fold $f_{4d}$ of the resulting GEMM in order to bring the original $4d$ structure of $\mathbf{A}_{i-1}$ back to $\mathbf{A}_i$. Correspondingly, we denote the complimentary procedure which takes in $4d$ representations and folds them to $2d$ arrays as $f_{2d}$:

$$f_{2d}\left[\mathbb{R}^{n\times d\times r\times r}\right] \implies \mathbb{R}^{d\times(r\times r\times n)} \tag{29}$$

$$f_{4d}\left[\mathbb{R}^{d\times(r\times r\times n)}\right] \implies \mathbb{R}^{n\times d\times r\times r} \tag{30}$$

These geometric operations will also prove useful for batch normalization layers; see section 9.

**Small side note about convolution layers**

*(The type of convolution layer discussed in this section is usually referred to as a "2d-convolution". This is due to the fact that only the spatial components (2d) of the kernels are locally connected to the input data whereas the depth dimension is still fully connected. In addition, it may be worth emphasizing that the $\phi$ transformation is not specific to the forward pass but that any convolution operation can be replaced by a matrix multiplication.)*

More generally, convolution layers should be viewed as **universal and powerful feature extractors applicable to all kinds of data modality** by astutely adapting the dimensionality of the kernels. For example, video data is routinely studied with 3d-convolutions where both space (2d) and time (1d) are locally connected to their input. Conversely, sequential data such as time series or text can be treated with the help of 1d-convolutions.

On that note, let us mention that although sequence modeling has long been quasi-synonymous with recurrent neural networks (RNNs), this association is being challenged by recent research indicating that substantial (stateless) memory can be incorporated in to convolutional architectures, allowing CNNs to compete and even outperform canonical recurrent architectures (such as LSTMs and GRUs) in traditional speech recognition and various natural language processing tasks [72, 73]. Arousing further suspicion, the behavior of dynamical systems implemented by RNNs is understood to be somehow reminiscent to that induced by shortcut connections; a type of layer that has become common sight in modern (and more hardware-friendly) feed-forward architectures [74]. Nevertheless, despite these reservations RNNs and variants thereof remain popular in industrial/research projects and are the focus of a separate pedagogical article [75].

**Backward pass (fractionally strided convolutions)** As demonstrated in section 2, the backward pass is determined by specializing the recursive relation derived in eq.(11) to convolutions with:

$$\begin{aligned}\Delta_i \cdot \mathrm{d}\mathbf{A}_i &= \Delta_i \cdot \mathrm{d}\left(\mathbf{w}^{\mathrm{p}}_{i-1} \star \mathbf{A}_{i-1} + \widetilde{\mathbf{b}_{i-1}}\right) \\ &\boxed{+} \\ &= \underbrace{\mathrm{roll}\left[f_{2d}\left(\Delta_i\right)\phi(\mathbf{A}_{i-1})^t\right]}_{\frac{\partial \mathcal{L}_{\mathrm{batch}}}{\partial \mathbf{w}^{\mathrm{p}}_{i-1}}} \cdot \mathrm{d}\mathbf{w}^{\mathrm{p}}_{i-1} + \underbrace{\sum_{\substack{\text{samples}\\ \text{\& space}}} \Delta_i}_{\frac{\partial \mathcal{L}_{\mathrm{batch}}}{\partial \mathbf{b}_{i-1}}} \cdot \mathrm{d}\mathbf{b}_{i-1} + \left[\left(\overline{\mathbf{w}}^{\mathrm{p}}_{i-1}\right)^t f_{2d}\left(\Delta_i\right)\right] \cdot \mathrm{d}\phi(\mathbf{A}_{i-1})\end{aligned}$$

Apart from facilitating the evaluation of the components of the gradient with respect to $\mathcal{P}_{i-1}$, a side-effect of having used the im2col transformation $\phi$ during the forward pass is that it leaves us with an awkward term $\left[\left(\overline{\mathbf{w}}^{\mathrm{p}}_{i-1}\right)^t f_{2d}\left(\Delta_i\right)\right] \cdot \mathrm{d}\phi(\mathbf{A}_{i-1})$ that we would like to invert in order to get back the dimensionality of $\mathbf{A}_{i-1}$ and extract the downstream error $\Delta_{i-1}$. The difficulty stems from the weight-sharing property of convolutions which causes the inverse of the $\phi$ mapping to be "one-to-many" and therefore not a well-defined function. It turns out that reaching an explicit closed-form mathematical expression of the "inverse" procedure $\phi^{-1}$ involves **tedious and heavily index-based manipulations** that are difficult to express elegantly. Nonetheless, efficient implementations of col2im do exist in image processing libraries and they are indeed selected by modern deep learning frameworks.

Since this article attempts to offer a more graceful **vectorized derivation** of backpropagation, let us offer an alternative and more intuitive approach. The key insight consists in recognizing that the upstream error appears in the form of a GEMM $\sim (\overline{\mathbf{w}}^{\mathfrak{p}}_{i-1})^t \, f_{2d}(\Delta_i)$ and, therefore, can be re-expressed as a convolution $\star$ by applying a **formal** inverse protocol $\phi^{-1}$; just like $\phi$ was itself initially used to transform $\star$ to a GEMM. In other words:

$$\left[ (\overline{\mathbf{w}}^{\mathfrak{p}}_{i-1})^t \, f_{2d}(\Delta_i) \right] \cdot \mathrm{d}\phi(\mathbf{A}_{i-1}) \longrightarrow_{\phi^{-1}} \underbrace{\overset{\curvearrowleft t_{\mathrm{d}}\ \mathfrak{p}'}{\mathbf{w}_{i-1}} \star \Delta_i}_{\Delta_{i-1}} \cdot \mathrm{d}\mathbf{A}_{i-1} \tag{31}$$

where the convolution operation $\star$ in the downstream error $\Delta_{i-1}$ must satisfy a number of constraints in order to mimic peculiarities of the non-invertible, one-to-many, $\phi^{-1}$ protocol. This is achieved by introducing a geometrically modified version of the original weights: $\overset{\curvearrowleft t_{\mathrm{d}}\ \mathfrak{p}'}{\mathbf{w}_{i-1}} \leftarrow \mathbf{w}^{\mathfrak{p}}_{i-1}$ which we proceed to describe in detail. First, the sampling triplet $\mathfrak{p}'$ needs to be selected such that the dimensionality of the input feature map will be recovered downstream, namely $\Delta_{i-1} \sim \mathbf{A}_{i-1}$. Second, we must ensure that the connectivity pattern that was established between $\mathbf{A}_{i-1}$ and $\mathbf{w}^{\mathfrak{p}}_{i-1}$ during the forward pass will be respected during the backward pass between $\overset{\curvearrowleft t_{\mathrm{d}}\ \mathfrak{p}'}{\mathbf{w}_{i-1}}$ and $\Delta_i$. As demonstrated in the animated bottom panel of Fig.12 and by exhaustive enumeration in table 14, these constraints are satisfied by executing a series of simple geometric transformations on the original weights:

- transpose the depth dimensions $d_i$ and $d_{i-1}$ of $\mathbf{w}^{\mathfrak{p}}_{i-1} \sim \mathbb{R}^{d_i \times d_{i-1} \times k \times k}$ into $\overset{\curvearrowleft t_{\mathrm{d}}\ \mathfrak{p}'}{\mathbf{w}_{i-1}} \sim \mathbb{R}^{d_{i-1} \times d_i \times k \times k}$ as indicated by the $t_{\mathrm{d}}$ superscript. This guarantees a well-defined convolution that restores the original depth $d_{i-1}$ of $\mathbf{A}_{i-1}$ to the downstream error $\Delta_{i-1}$.
- perform a 180-degree rotation of the weights as indicated by the $\curvearrowleft$ symbol.
- associate the convolution with a new sampling parameter $\mathfrak{p}' = \left(k' = k, s' = 1/s, p' = k - p - 1\right)$. Notice that the stride $s'$ is now fractional and in inverse proportion to that defined by $\mathfrak{p}$ in the forward pass [76]. This fractionally strided convolution is implemented by setting $s' = 1$ and embedding padding zeros in between the spatial components of $\Delta_i$. The effect of this "internal" padding is to slow down the sliding of the kernel in order to establish a connectivity pattern between $\Delta_i$ and $\overset{\curvearrowleft t_{\mathrm{d}}\ \mathfrak{p}'}{\mathbf{w}_{i-1}}$ compatible with the one between $\mathbf{A}_{i-1}$ and $\mathbf{w}^{\mathfrak{p}}_{i-1}$.

Keep in mind that, of course, the actual component values of $\overset{\curvearrowleft t_{\mathrm{d}}\ \mathfrak{p}'}{\mathbf{w}_{i-1}}$ and $\mathbf{w}^{\mathfrak{p}}_{i-1}$ are identical to each other modulo the geometric transformations described above. Incidentally, this fractionally strided convolution during the backward pass should be understood as an upsampling of the upstream error $\Delta_i$ in contrast to the forward convolution that acted as a downsampling of the input feature map $\mathbf{A}_{i-1}$.

Furthermore, one can see that apart from slightly cumbersome geometrical data folding, the backward pass of convolution layers is indeed very similar to that of fully connected layers (section 5). In both cases, the gradient component with respect to the weights involve a matrix product between the transpose of the original data $\mathbf{A}_{i-1}$ and the upstream error $\Delta_i$ while the bias component is obtained by contraction of $\Delta_i$ according to the broadcast rules applied to the bias vector during the forward pass. Although intellectually pleasant, this implementation of error backpropagation via fractionally strided convolutions between weights and upstream error (which mirrors that of fully connected layers, replacing matrix product by $\star$) is not the most computationally efficient and deep learning compilers provide lower level optimizations that break this symmetry.

**Convolution**: backward pass (fractional stride)

$$\Delta_{i-1} = \overset{\curvearrowleft t_{\mathrm{d}}\ \mathfrak{p}'}{\mathbf{w}_{i-1}} \star \Delta_i \qquad \sim \mathbb{R}^{n \times d_{i-1} \times r_{i-1} \times r_{i-1}} \tag{32}$$

$$\frac{\partial \mathcal{L}_{\text{batch}}}{\partial \mathbf{w}^{\mathfrak{p}}_{i-1}} = \text{roll}\left[ f_{2d}(\Delta_i)\, \phi(\mathbf{A}_{i-1})^t \right] \qquad \sim \mathbb{R}^{d_i \times d_{i-1} \times k \times k} \tag{33}$$

$$\frac{\partial \mathcal{L}_{\text{batch}}}{\partial \mathbf{b}_{i-1}} = \sum_{\substack{\text{samples} \\ \&\ \text{space}}} \Delta_i \qquad \sim \mathbb{R}^{d_i} \tag{34}$$

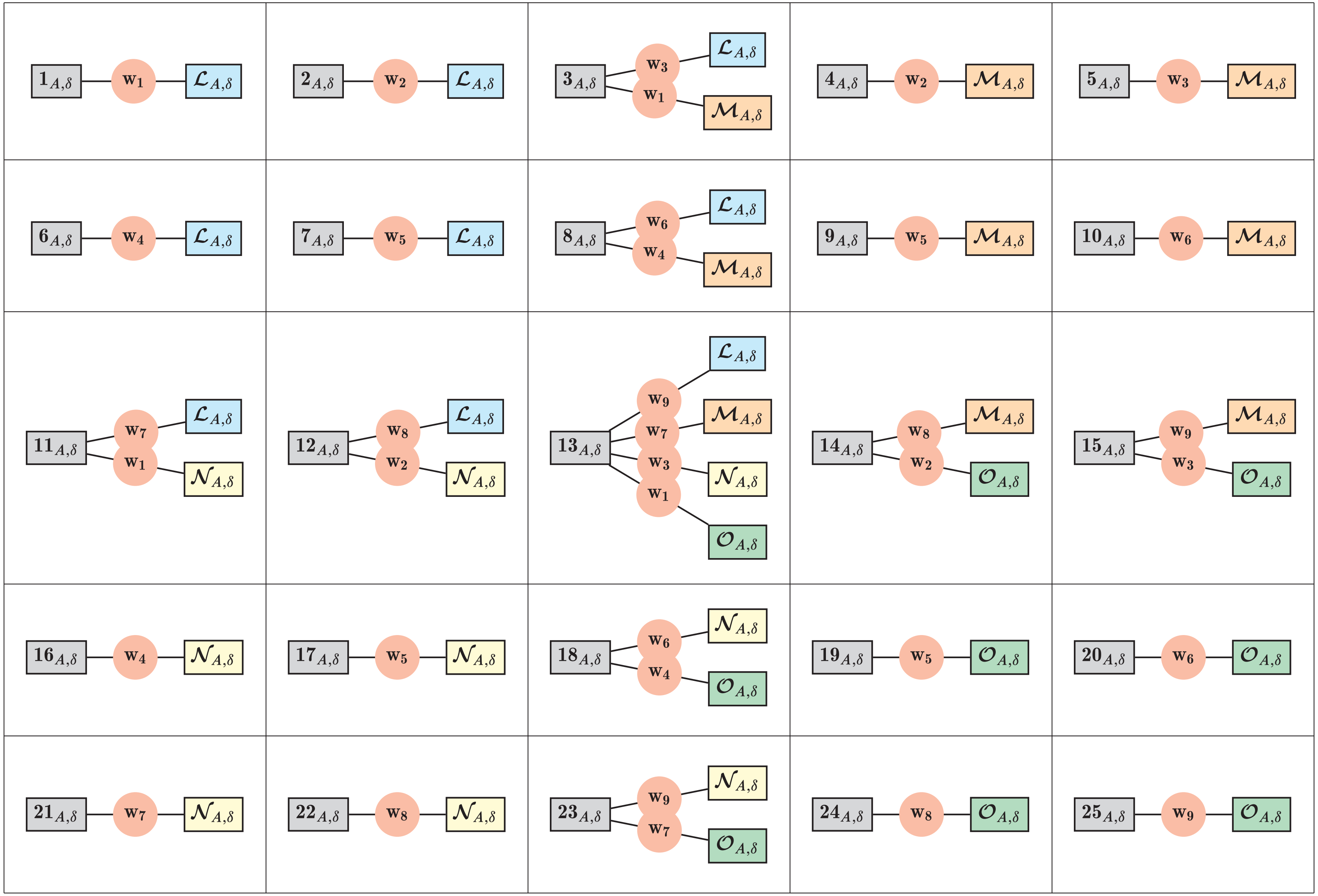

Figure 14: Exhaustive enumeration of the **connectivity pattern** defined by the convolution illustrated in Fig. 12. One can verify that thanks to the geometric transformation $\overset{\curvearrowleft}{\mathbf{w}}{}_{i-1}^{t_d\,p'} \leftarrow \mathbf{w}_{i-1}^{p}$, the local connections between data and error values are indeed identical for both forward/backward passes. *(Colors are intended as a local guide for the eye only and are not related to the general color code adopted in the rest of the article.)*

# 9 Batch normalization (BN) layer

It is considered good-practice among machine learning professionals to include a feature rescaling pre-processing step in their model building workflow. Ordinarily, the idea may be as elementary as to "standardize" the data so that each rescaled feature has zero mean and unit standard deviation. This way, features with different scales and units can be more easily compared with each other and integrated together into a learning algorithm (such as gradient descent) with superior convergence properties. Since the batch normalization layer of modern deep learning architectures is also based on this standardization of the features, let us start by a brief summary of traditional feature normalization.

**Traditional feature normalization** Typically, input data is represented as a 2d array $\mathbf{A} \sim \mathbb{R}^{n \times f}$ by aggregating a minibatch of $n$ independent samples which are, individually, encoded as 1d feature vectors $\sim \mathbb{R}^f$ composed of $f$ scalar features. Let us focus on a specific feature $\mathfrak{f} \in \{1, \cdots, f\}$. Drawing values from across the minibatch samples, $\mathfrak{f}$ can be represented as a vector $\mathbf{a}_\mathfrak{f} = (a_\mathfrak{f}^1, \cdots, a_\mathfrak{f}^n) \sim \mathbb{R}^n$ whose mean and standard deviation are estimated from the minibatch statistics by:

$$\mu_\mathfrak{f} = \frac{1}{n}\left(a_\mathfrak{f}^1 + \cdots + a_\mathfrak{f}^n\right) \sim \mathbb{R} \tag{35}$$

$$\sigma_\mathfrak{f} = \sqrt{\frac{1}{n}\left[\left(a_\mathfrak{f}^1 - \mu_\mathfrak{f}\right)^2 + \cdots + \left(a_\mathfrak{f}^n - \mu_\mathfrak{f}\right)^2\right]} \sim \mathbb{R} \tag{36}$$

These summary statistics are used to transform $\mathbf{a}_\mathfrak{f}$ into a "batch normalized" vector:

$$\overline{\mathbf{a}}_\mathfrak{f} = \frac{\mathbf{a}_\mathfrak{f} - \mu_\mathfrak{f}}{\sigma_\mathfrak{f}} \sim \mathbb{R}^n \tag{37}$$

where the mean $\mu_\mathfrak{f}$ and standard deviation $\sigma_\mathfrak{f}$ are broadcast vector-wise such that basic arithmetic operations with individual components of $\overline{\mathbf{a}}_\mathfrak{f}$ are well-defined. By construction, the batch-normalized vector $\overline{\mathbf{a}}_\mathfrak{f}$ achieves the desired properties of zero mean $\mu_{\overline{\mathbf{a}}_\mathfrak{f}} = 0$ and unit standard deviation $\sigma_{\overline{\mathbf{a}}_\mathfrak{f}} = 1$. Note that this transformation does not change the nature of the underlying statistical distribution that generates feature $\mathfrak{f}$ but simply performs a linear "shift and rescale" operation on its first 2 moments.

Going beyond a single feature, let us now consider the general situation of $f$ features. In this case, we can define a pair of vectors $\mu = (\mu_1, \cdots, \mu_f) \sim \mathbb{R}^f$ and $\sigma = (\sigma_1, \cdots, \sigma_f) \sim \mathbb{R}^f$ made up of the minibatch mean values and standard deviations for all features $\mathfrak{f} \in \{1, \cdots, f\}$. Then, one can assemble together all the batch-normalized features into a single data array:

$$\overline{\mathbf{A}} = \left(\overline{\mathbf{a}}_1, \cdots, \overline{\mathbf{a}}_f\right) = \left(\frac{\mathbf{a}_1 - \mu_1}{\sigma_1}, \cdots, \frac{\mathbf{a}_f - \mu_f}{\sigma_f}\right) \sim \mathbb{R}^{n \times f}$$

This expression may be vectorized further by introducing some broadcasting semantics: i) broadcast the mean value vector $\mu \to \widetilde{\mu} \sim \mathbb{R}^{n \times f}$ additively across the minibatch samples ii) use a multiplicative broadcast to bring the inverse of the standard deviation vector $1/\sigma = (1/\sigma_1, \cdots, 1/\sigma_f) \sim \mathbb{R}^f$ into a square diagonal matrix $1/\sigma \to 1/\widetilde{\sigma} \equiv \text{diag}(1/\sigma) \sim \mathbb{R}^{f \times f}$ as shown in eq.(46). Using these broadcasting rules, the normalized data array can be expressed as:

$$\overline{\mathbf{A}} = \frac{\mathbf{A} - \widetilde{\mu}}{\widetilde{\sigma}} \tag{38}$$

+

Note that we have implicitly introduced the shorthand notation $(\mathbf{A} - \widetilde{\mu}) / \widetilde{\sigma} \equiv (\mathbf{A} - \widetilde{\mu})\,\text{diag}(1/\sigma)$ where division is overloaded to represent the matrix product $\sim \mathbb{R}^{n \times f}\, \mathbb{R}^{f \times f} \sim \mathbb{R}^{n \times f}$ with the multiplicative broadcast of $1/\sigma$. The diagonality of this broadcast is crucial to ensure that each scaling factor $1/\sigma_\mathfrak{f}$ is indeed coupled with its correct companion vector $\mathbf{a}_\mathfrak{f}$ as defined in eq.(37) and derived explicitly in the optional content above.

**Feature definition for 3d image data** How does this concept of traditional feature normalization carry over for 3d image data? In order to answer this, one should look back at what a "feature" is in the context of CNNs. As explained in section 8, convolutional kernels are locally connected in space with the input data patches but they remain fully connected along the depth dimension. Because of this choice of dimensionality, depth is tensor-contracted out as kernels slide across the input data and each kernel generates a spatial feature $\sim \mathbb{R}^{r\times r}$, so-called activation map, which may be unrolled into a vector $\sim \mathbb{R}^{(r\times r)}$. Considering that a general convolution layer comes equipped with a number $f$ of such independent kernels [8], its output should be understood as a set of $f$ features $\sim \mathbb{R}^{(r\times r)\times f}$ where each individual feature is represented by a vector composed of $r \times r = r^2$ unrolled values (instead of just one scalar value per feature as discussed in the previous paragraph for 1d data). Aggregating together the feature vectors for all $n$ samples (images), we get a 2d representation of the data $\sim \mathbb{R}^{(n\times r\times r)\times f}$ where one can think of $n_{\text{eff}} \equiv n \times r \times r$ as an "effective" minibatch size reflecting the intrinsic origin of features as unrolled 2d spatial maps into 1d vectors.

Conveniently enough, geometric reshape operations of data arrays from 4d to 2d and back were already introduced in section 8 and can be directly re-used (up to a simple transpose) for batch normalization. For example, 4d minibatch arrays of images can be converted to their relevant 2d feature vector representations using eq.(29) via $f_{2d}^{t}\left[\mathbb{R}^{n\times f\times r\times r}\right] \implies \mathbb{R}^{n_{\text{eff}}\times f}$ where we denote by $f$ the number of features (i.e. output depth) and make use of the effective minibatch size $n_{\text{eff}}$. Correspondingly, folding a 2d array which is known to have originated from 4d image data is accomplished by applying the transpose of eq.(30) via $f_{4d}^{t}\left[\mathbb{R}^{n_{\text{eff}}\times f}\right] \implies \mathbb{R}^{n\times f\times r\times r}$. These transformations allow us to keep a unified description of batch normalization independent of the underlying dimensionality of the data. When dealing with 4d data, the idea consists in reshaping it into 2d feature vector representations via $f_{2d}$ and use all the formalism described in the following sections, simply replacing $n$ with $n_{\text{eff}}$. Only in the end, one concludes by folding the result back into its relevant 4d structure by invoking $f_{4d}$. For example, the traditional feature normalization defined in eq.(38) can be adapted to 4d image data through the following gymnastics:

$$\overline{\mathbf{A}} = f_{4d}^{t}\left(\frac{f_{2d}^{t}(\mathbf{A}) - \widetilde{\mu}}{\widetilde{\sigma}}\right)$$

**Forward pass** The traditional feature normalization procedure reviewed above has proven itself very successful and maintains its hold as a popular pre-processing step in machine learning pipelines. Nonetheless, introducing it "as is" into the architecture of a neural network is not ideal since restricting activation values to a certain range might limit the representations that can be achieved by subsequent layers. In particular, it would be desirable that any normalization layer inserted into the network would, at least, be able to represent the identity transform. In other words, it should be possible for this layer to learn how to recover the original unnormalized data if that turns out, empirically, to be the optimal thing to do. In order to see how to accomplish this, let us go back to a minibatch standard-normalized feature $\bar{\mathbf{a}}_{\mathfrak{f}} \sim \mathbb{R}^n$ with zero mean and unit standard deviation. Introducing a pair of scalar parameters $w_{\mathfrak{f}} \sim \mathbb{R}$ and $b_{\mathfrak{f}} \sim \mathbb{R}$, one can then apply an affine transformation:

$$\text{BN}\left(\bar{\mathbf{a}}_{\mathfrak{f}}\right) = w_{\mathfrak{f}}\, \bar{\mathbf{a}}_{\mathfrak{f}} + b_{\mathfrak{f}}$$

First of all, let us demonstrate how this weight/bias pair of adjustable parameters trivially determines the summary statistics of batch-normalized $\text{BN}\left(\bar{\mathbf{a}}_{\mathfrak{f}}\right)$ through:

$$\mu_{\text{BN}(\bar{\mathbf{a}}_{\mathfrak{f}})} = b_{\mathfrak{f}}$$

+

$$\sigma_{\text{BN}(\bar{\mathbf{a}}_{\mathfrak{f}})} = w_{\mathfrak{f}}$$

+

Evidently, giving the possibility to the learning algorithm to converge to the special case $w_{\mathfrak{f}} \approx \sigma_{\mathfrak{f}}$ and $b_{\mathfrak{f}} \approx \mu_{\mathfrak{f}}$, as defined in eqs.(35,36), demonstrates how this affine transformation may reclaim the original unnormalized values of the feature vector, namely $\text{BN}\left(\bar{\mathbf{a}}_{\mathfrak{f}}\right)_{w_{\mathfrak{f}}\approx\sigma_{\mathfrak{f}}\,;\,b_{\mathfrak{f}}\approx\mu_{\mathfrak{f}}} \approx \mathbf{a}_{\mathfrak{f}}$.

[8] Notice that the number of kernels is usually denoted by $d$ as it represents the output depth after a convolution layer. We denote it here as $f$ to emphasize that it is also the number of independent features as far as data normalization is concerned.

Going back to the general case of multiple features, we consolidate the newly-introduced adjustable weights $\mathbf{w} = (w_1, \cdots, w_f) \sim \mathbb{R}^f$ and biases $\mathbf{b} = (b_1, \cdots, b_f) \sim \mathbb{R}^f$ into vectors. The batch-normalized version of $\overline{\mathbf{A}}$ can then be expressed as:

$$\mathrm{BN}\left(\overline{\mathbf{A}}\right) = \left(w_1\, \overline{\mathbf{a}}_1 + b_1, \cdots, w_f\, \overline{\mathbf{a}}_f + b_f\right)$$

$$+$$

$$= \overline{\mathbf{A}}\, \widetilde{\mathbf{w}} + \widetilde{\mathbf{b}}$$

where the weights are broadcast diagonally $\mathbf{w} \to \widetilde{\mathbf{w}} \equiv \mathrm{diag}\, \mathbf{w} \sim \mathbb{R}^{f\times f}$ in order to scale the features (columns) of $\overline{\mathbf{A}} \sim \mathbb{R}^{n\times f}$ with the appropriate component of the weight vector; see appendix B. Biases are additively broadcast sample-wise $\mathbf{b} \to \widetilde{\mathbf{b}} \sim \mathbb{R}^{n\times f}$ as usual.

In summary, the batch normalization layer applies an affine transformation parametrized by:

$$\mathcal{P}_{i-1} \begin{cases} \mathbf{w}_{i-1} \sim \mathbb{R}^{f_{i-1}}\,, & \text{"weights"} \\ \mathbf{b}_{i-1} \sim \mathbb{R}^{f_{i-1}}\,, & \text{"biases"} \end{cases}$$

to the $f \equiv f_{i-1}$ features of a data array $\mathbf{A}_{i-1} \sim \mathbb{R}^{n\times f_{i-1}}$ on top of a more traditional feature standardization pre-processing step. The final output of the forward pass is given by:

**Batch normalization:** forward pass

$$\mathbf{A}_i = \overline{\mathbf{A}}_{i-1}\widetilde{\mathbf{w}_{i-1}} + \widetilde{\mathbf{b}_{i-1}} \quad \text{with} \quad \overline{\mathbf{A}}_{i-1} = \frac{\mathbf{A}_{i-1} - \widetilde{\mu}}{\widetilde{\sigma}} \tag{39}$$

The broadcasting rules of the data standardization parameters $\mu$, $\sigma$ and are defined in eq.(38) and those of the affine transformation $\mathbf{w}_{i-1}$ and $\mathbf{b}_{i-1}$ are defined just above. Note that batch normalization layers do not change the dimensionality of the data $\mathbf{A}_i \sim \mathbf{A}_{i-1}$ as they simply provide a parametrized linear shift and rescale of all the features independently from each other.

$+$ *For details about geometric reshape operations in the case of 3d (image) data.*

It is interesting to point out that batch normalization may account for a significant portion of training time since it requires two passes through the input data $\mathbf{A}_{i-1}$. First to compute the batch statistics $\mu$ and $\sigma$, and then to perform the normalization $\overline{\mathbf{A}}_{i-1}$; a process bounded by memory bandwidth that is not easily parallelizable [77]. Although we focus on training, let us mention that one needs to be cautious, engineering-wise, when using batch normalization for inference. Since the concept of minibatch statistics is no longer relevant, it is customary to use population statistics based on the entire training dataset during inference.

**Backward pass** As with all other layers, the backward pass of the batch normalization layer starts by applying the recursive relation eq.(11):

$$\Delta_i \cdot \mathrm{d}\mathbf{A}_i = \Delta_i \cdot \mathrm{d}\left(\overline{\mathbf{A}}_{i-1}\widetilde{\mathbf{w}_{i-1}} + \widetilde{\mathbf{b}_{i-1}}\right)$$

$$+$$

$$= \underbrace{\mathrm{diag}\left(\overline{\mathbf{A}}^t_{i-1}\Delta_i\right)}_{\dfrac{\partial\mathcal{L}_{\text{batch}}}{\partial\mathbf{w}_{i-1}}} \cdot \mathrm{d}\mathbf{w}_{i-1} + \underbrace{\sum_{\text{samples}} \Delta_i}_{\dfrac{\partial\mathcal{L}_{\text{batch}}}{\partial\mathbf{b}_{i-1}}} \cdot \mathrm{d}\mathbf{b}_{i-1} + \left(\Delta_i\,\widetilde{\mathbf{w}_{i-1}}\right)\cdot \mathrm{d}\overline{\mathbf{A}}_{i-1}$$

Although it offers no conceptual challenges, the explicit calculation of $\left(\Delta_i\,\widetilde{\mathbf{w}_{i-1}}\right)\cdot \mathrm{d}\overline{\mathbf{A}}_{i-1}$, which is crucial to extract the downstream error $\Delta_{i-1}$, turns out to be more tedious. Essentially, the calculation boils down to writing the normalized total derivative $\mathrm{d}\overline{\mathbf{A}}_{i-1}$ in terms of the unnormalized $\mathrm{d}\mathbf{A}_{i-1}$. In order to make progress, it is useful to break up the calculation into a few independent steps.

**Small side note about normalization layers**

Since its introduction in [78], batch normalization continues to be credited as a crucial ingredient behind the current deep learning renaissance. Empirically, it has been shown that this normalization mechanism provides tremendous help in stabilizing the training of neural networks by allowing the use of large learning rates with which SGD can reach good data representations faster and more reliably. By preventing the saturation of activation functions, batch normalization also contributes to the elimination of vanishing/exploding gradients; issues that used to poison neural network training. Moreover, it is suspected that the noise introduced by finite-size batch statistics may endow batch normalization with a form of regularization somewhat similar to dropout [16].

Despite widely acknowledged benefits, the jury is nevertheless still out with regards to the underlying reason for the success of batch normalization. Although it was originally argued that batch normalization works by limiting "internal covariate shift" (a process by which input statistical distributions supposedly keep changing thereby slowing down training), recent evidence casts doubts on this view [79]. In general, the idea seems to be that by letting some batch statistics free to be learned by the data instead of being imposed by complex high-order interactions between layers, the dynamics of learning becomes more layer-independent which, in turn, helps stabilize the learning rate [18]. In any case, the literature on normalization layers has expanded enormously and a number of other techniques have recently been proposed in order to fix some shortcomings of batch normalization [80].

*(Technical addendum: it was originally proposed to place the batch normalization layer before the non-linear activation layer, presumably so that the normalized inputs to the non-linearity have a better chance to avoid saturating the activation function; a situation more relevant when dealing with sigmoid/tanh activations for example... However, architectures based on ReLU sometimes place batch normalization after the non-linearity so that negative input values, which are to be eliminated by ReLU, do not contribute to the minibatch statistics; as we implement here in the example network architecture specified in table 1.)*

First of all, let us consider a specific sample $\mathfrak{s}$ and feature $\mathfrak{f}$ extracted from $\overline{\mathbf{A}}_{i-1}$ so as to start by focusing on the calculation of the total derivative of an arbitrary component $\overline{a}_\mathfrak{f}^\mathfrak{s} \sim \mathbb{R}$. Keeping in mind how $\overline{\mathbf{a}}_\mathfrak{f}$ is defined, see eq.(37), promptly yields the total derivative of one of its component as:

$$\mathrm{d}\overline{a}_\mathfrak{f}^\mathfrak{s} = \left(\frac{\partial \overline{a}_\mathfrak{f}^\mathfrak{s}}{\partial a_\mathfrak{f}^\mathfrak{s}}\right) \mathrm{d}a_\mathfrak{f}^\mathfrak{s} + \left(\frac{\partial \overline{a}_\mathfrak{f}^\mathfrak{s}}{\partial \mu_\mathfrak{f}}\right) \mathrm{d}\mu_\mathfrak{f} + \left(\frac{\partial \overline{a}_\mathfrak{f}^\mathfrak{s}}{\partial \sigma_\mathfrak{f}}\right) \mathrm{d}\sigma_\mathfrak{f}$$

+

$$= \frac{1}{\sigma_\mathfrak{f}} \mathrm{d}a_\mathfrak{f}^\mathfrak{s} - \frac{1}{\sigma_\mathfrak{f}} \mathrm{d}\mu_\mathfrak{f} - \frac{\overline{a}_\mathfrak{f}^\mathfrak{s}}{\sigma_\mathfrak{f}} \mathrm{d}\sigma_\mathfrak{f}$$

for $\mathrm{d}\mu_\mathfrak{f}$ +

for $\mathrm{d}\sigma_\mathfrak{f}$ +

$$= \frac{1}{n\sigma_\mathfrak{f}} \left( n\, \mathrm{d}a_\mathfrak{f}^\mathfrak{s} - \sum_\text{samples} \mathrm{d}a_\mathfrak{f}^s - \overline{a}_\mathfrak{f}^\mathfrak{s} \sum_\text{samples} \overline{a}_\mathfrak{f}^s \mathrm{d}a_\mathfrak{f}^s \right)$$

As it will soon become helpful, it is worth mentioning that $\mathrm{d}\overline{a}_\mathfrak{f}^\mathfrak{s}$ can be transformed into a compositional form by factoring out $\mathrm{d}a_\mathfrak{f}^\mathfrak{s}$:

$$\mathrm{d}\overline{a}_\mathfrak{f}^\mathfrak{s} = \frac{1}{n\sigma_\mathfrak{f}} \left( n\,\mathbf{1} - \sum_\text{samples} \mathbf{1} - \overline{a}_\mathfrak{f}^\mathfrak{s} \sum_\text{samples} \overline{a}_\mathfrak{f}^s \right) \mathrm{d}a_\mathfrak{f}^\mathfrak{s} \tag{40}$$

where we introduce the identity mapping $\mathbf{1}$ as an indicator that the terms in between the parentheses should be thought of as operators acting on $\mathrm{d}a_\mathfrak{f}^\mathfrak{s}$.

The next step consists in applying eq.(40), valid for an arbitrary component $d\overline{a}^{\mathfrak{s}}_{\mathfrak{f}}$ of $d\overline{\mathbf{A}}_{i-1}$, to all samples $\mathfrak{s} \in \{1, \cdots, n\}$ and features $\mathfrak{f} \in \{1, \cdots, f \equiv f_{i-1}\}$ thereby allowing us to populate all the components of the total derivative of the input feature map with closed-form expressions:

$$d\overline{\mathbf{A}}_{i-1} = \begin{pmatrix} d\overline{a}^1_1 & \dots & d\overline{a}^1_f \\ \vdots & \vdots & \vdots \\ d\overline{a}^n_1 & \dots & d\overline{a}^n_f \end{pmatrix}$$

+

↓ overloading division as $1/\widetilde{\sigma}$ to represent matrix product with diag$(1/\sigma)$; same as in eq.(38)

$$= \frac{1}{n\widetilde{\sigma}} \left[ n\,(\mathbf{1} \circ d\mathbf{A}_{i-1}) - \underbrace{\begin{pmatrix} \sum_s da^s_1 & \dots & \sum_s da^s_f \\ \vdots & \vdots & \vdots \\ \sum_s da^s_1 & \dots & \sum_s da^s_f \end{pmatrix}}_{\widetilde{\sum_{\text{samples}} \mathbf{1} \circ d\mathbf{A}_{i-1}}} - \underbrace{\begin{pmatrix} \overline{a}^1_1 \sum_s \overline{a}^s_1 d\overline{a}^s_1 & \dots & \overline{a}^1_f \sum_s \overline{a}^s_f d\overline{a}^s_f \\ \vdots & \vdots & \vdots \\ \overline{a}^n_1 \sum_s \overline{a}^s_1 d\overline{a}^s_1 & \dots & \overline{a}^n_f \sum_s \overline{a}^s_f d\overline{a}^s_f \end{pmatrix}}_{\overline{\mathbf{A}}_{i-1} \circ \widetilde{\sum_{\text{samples}} \overline{\mathbf{A}}_{i-1} \circ d\mathbf{A}_{i-1}}} \right]$$

middle term +
last term +

↓ restructure as a Hadamard product

by factoring out $d\mathbf{A}_{i-1}$ and introducing an identity operator $\mathbf{1} \sim \mathbb{R}^{n \times f_{i-1}}$

$$= \frac{1}{n\widetilde{\sigma}} \left( n\,\mathbf{1} - \widetilde{\sum_{\text{samples}} \mathbf{1}} - \overline{\mathbf{A}}_{i-1} \circ \widetilde{\sum_{\text{samples}} \overline{\mathbf{A}}_{i-1}} \right) \circ d\mathbf{A}_{i-1}$$

As expected, this compositional formulation of $d\overline{\mathbf{A}}_{i-1}$ naturally comes out as an elegant and vectorized generalization of eq.(40). Substituting it back into the sought-after term yields an expression from which one can finally identify the downstream error $\Delta_i$:

$$\left(\Delta_i \widetilde{\mathbf{w}_{i-1}}\right) \cdot d\overline{\mathbf{A}}_{i-1} = \left(\Delta_i \widetilde{\mathbf{w}_{i-1}}\right) \cdot \frac{1}{n\widetilde{\sigma}} \left( n\,\mathbf{1} - \widetilde{\sum_{\text{samples}} \mathbf{1}} - \overline{\mathbf{A}}_{i-1} \circ \widetilde{\sum_{\text{samples}} \overline{\mathbf{A}}_{i-1}} \right) \circ d\mathbf{A}_{i-1}$$

+

$$= \underbrace{\frac{1}{n\widetilde{\sigma}} \left( n\Delta_i \widetilde{\mathbf{w}_{i-1}} - \widetilde{\sum_{\text{samples}} \Delta_i \widetilde{\mathbf{w}_{i-1}}} - \overline{\mathbf{A}}_{i-1} \circ \widetilde{\sum_{\text{samples}} \overline{\mathbf{A}}_{i-1} \circ \Delta_i \widetilde{\mathbf{w}_{i-1}}} \right)}_{\Delta_{i-1}} \cdot d\mathbf{A}_{i-1}$$

In summary, the backward pass of the batch normalization layer is given by:

**Batch normalization**: backward pass

$$\Delta_{i-1} = \frac{1}{n\widetilde{\sigma}} \left( n\Delta_i \widetilde{\mathbf{w}_{i-1}} - \widetilde{\sum_{\text{samples}} \Delta_i \widetilde{\mathbf{w}_{i-1}}} - \overline{\mathbf{A}}_{i-1} \circ \widetilde{\sum_{\text{samples}} \overline{\mathbf{A}}_{i-1} \circ \Delta_i \widetilde{\mathbf{w}_{i-1}}} \right) \sim \mathbb{R}^{n \times f_{i-1}} \quad (41)$$

$$\frac{\partial \mathcal{L}_{\text{batch}}}{\partial \mathbf{w}_{i-1}} = \text{diag}\left( \overline{\mathbf{A}}^t_{i-1} \Delta_i \right) \sim \mathbb{R}^{f_{i-1}} \quad (42)$$

$$\frac{\partial \mathcal{L}_{\text{batch}}}{\partial \mathbf{b}_{i-1}} = \sum_{\text{samples}} \Delta_i \sim \mathbb{R}^{f_{i-1}} \quad (43)$$

+ *For details about geometric reshape operations in the case of 3d (image) data.*

*"The Queen propped her up against a tree, and said kindly, You may rest a little now.*

*Alice looked round her in great surprise. Why, I do believe we've been under this tree the whole time! Everything's just as it was!*

*Of course it is, said the Queen, what would you have it?*

*Well, in our country, said Alice, still panting a little, you'd generally get to somewhere else — if you ran very fast for a long time, as we've been doing.*

*A slow sort of country! said the Queen. Now, here, you see, it takes all the running you can do, to keep in the same place.*

*If you want to get somewhere else, you must run at least twice as fast as that!"*

(Lewis Carroll, Through the Looking-Glass, 1871)

# Appendix A Spatial sampling of feature maps

Illustrated examples of spatial sampling of feature maps are provided in:

- Fig. 10 (maxpool) and the top panel of Fig. 12 (convolution): **downsampling**
- bottom panel of Fig. 12 (convolution): **upsampling**

In both cases, the sampling operation can be understood as a sliding window parametrized by a geometric variable made up of 3 components:

$$\mathfrak{p} \equiv (\text{kernel size} = k, \text{stride} = s, \text{padding} = p) \tag{44}$$

The kernel size corresponds to the spatial extent (considered to be square for simplicity) of the sliding window. The stride corresponds to the number of cells to slide the kernel. Notice that fractional strides can be implemented by inserting zero-filled cells in between the actual data of the feature maps (bottom panel of Fig.12). Finally, padding corresponds to the number of zero-padding cells that may be added on the outer edges of feature maps. Note that padding is commonly used in order to control the spatial size of output feature maps. In general, sampling an input feature map $\mathbf{A}_{i-1}$ results in an output $\mathbf{A}_i$ which can be dimensionally summarized as:

$$\mathbf{A}_i \sim \mathbb{R}^{n\times d_i\times r_i\times r_i} = \text{sample}_{\mathfrak{p}}\left(\mathbf{A}_{i-1} \sim \mathbb{R}^{n\times d_{i-1}\times r_{i-1}\times r_{i-1}}\right)$$

The spatial size $r_i$ of the output can be calculated as a function of $r_{i-1}$ and of the geometrical properties $\mathfrak{p}$ of the sliding window via:

$$r_i = \left\lfloor \frac{r_{i-1} + 2p - k}{s} + 1 \right\rfloor \tag{45}$$

The relationship between $d_i$ and $d_{i-1}$ depends on the type of the sampling layer and is independent of $\mathfrak{p}$. For example, $d_i = d_{i-1}$ simply remains unchanged in the case of maxpool layers (section 6). On the other hand, $d_i$ corresponds to the desired number of trainable filters for convolutional layers (see the weights $\mathbf{w}_{i-1}^{\mathfrak{p}} \sim \mathbb{R}^{d_i\times d_{i-1}\times k\times k}$ in section 8).

# Appendix B Broadcasting semantics

Although arithmetic operations require the shape of data arrays to satisfy certain constraints, one would still like to perform vectorized operations (using transparent operator overloading) in case of dimensionality mismatch — as long as the arrays in question are of compatible sizes.

**Additive broadcast** For example, in the case of fully connected layers discussed in section 5, it is very common to add together 2d arrays $\mathbf{A} \sim \mathbb{R}^{n\times f}$ with 1d vectors $\mathbf{b} = (b_1, \cdots, b_f) \sim \mathbb{R}^f$. In this case, the size incompatibility stems from the conflict between our data representation where we simultaneously consider minibatches of $n$ samples and our desire to give each feature its own bias term. Conceptually, this conflict can easily be resolved by explicitly giving all samples their own copy of the original bias vector:

$$\mathbf{b} \Longrightarrow \widetilde{\mathbf{b}} = \begin{pmatrix} \mathbf{b} \\ \vdots \\ \mathbf{b} \end{pmatrix} = \begin{pmatrix} b_1 & \dots & b_f \\ \vdots & \vdots & \vdots \\ b_1 & \dots & b_f \end{pmatrix} \sim \mathbb{R}^{n\times f}$$

so that the bias shift can now be carried out with a well-defined addition $\mathbf{A} + \widetilde{\mathbf{b}}$ between arrays of same sizes. Of course, linear algebra frameworks implement this "broadcasting" process via smart memory-efficient techniques that circumvent the need for data duplication. Furthermore, broadcasting provides a means of vectorizing array operations for more efficient looping instructions. Reverting the broadcasting is performed by contracting out the "duplicated" dimensions. For example, in the case of the bias vector, a simple sum over the samples takes us back to the original 1d representation:

$$\sum_{\text{samples}} \widetilde{\mathbf{b}} \Longrightarrow \mathbf{b} = \left(b_1, \cdots, b_f\right) \sim \mathbb{R}^f$$

As an additional example, let us consider the broadcasting of the biases of convolutional kernels discussed in section 8. The process is very similar to that described above with the only difference that the data with which the bias vector $\mathbf{b} \sim \mathbb{R}^f$ needs to be added is now $\mathbf{A} \sim \mathbb{R}^{n\times f\times r\times r}$ a 4d array [9]. Accordingly, broadcasting is implemented by copying $\mathbf{b}$ into $n$ spatial feature maps of resolution $r$. Similarly, contracting out the newly created dimensions by summing over the spatial dimensions of the feature maps in addition to the samples reverts the broadcasting:

$$\begin{aligned} \mathbf{b} &\implies \widetilde{\mathbf{b}} \sim \mathbb{R}^{n\times f\times r\times r} \\ \sum_{\substack{\text{samples}\\ \text{\& space}}} \widetilde{\mathbf{b}} &\implies \mathbf{b} \sim \mathbb{R}^f \end{aligned}$$

**Multiplicative broadcast** Section 9 introduces another kind of broadcasting semantics. In this case, the idea consists in scaling all the features of a 2d data array $\mathbf{A} \sim \mathbb{R}^{n\times f}$ using a 1d vector $\lambda \sim \mathbb{R}^f$. Distribution of the components of $\lambda$ over to their matching feature can be accomplished by creating a diagonal broadcast:

$$\lambda = (\lambda_1, \cdots, \lambda_f) \sim \mathbb{R}^f \implies \widetilde{\lambda} = \begin{pmatrix} \lambda_1 & & \\ & \ddots & \\ & & \lambda_f \end{pmatrix} \sim \mathbb{R}^{f\times f} \tag{46}$$

so that feature scaling is achieved by a well-defined $\mathbf{A}\widetilde{\lambda}$ matrix multiplication. Accordingly, reverting this multiplicative broadcast is accomplished by picking out the diagonal components:

$$\text{diag}\,\widetilde{\lambda} \implies \lambda \sim \mathbb{R}^f$$

# Appendix C Matrices: a potpourri of what's relevant...

The purpose of this section is to serve a small collection of definitions and basic properties of matrix calculus that are used throughout the article. Keeping in mind the machine learning context, it is common to think of matrices $\mathbf{A} \sim \mathbb{R}^{n\times f}$ as a vertical stack of 1d feature vectors $\mathbf{a} = (a_1, \cdots, a_f) \sim \mathbb{R}^f$ with each row corresponding to an individual sample out of a minibatch of $n$ samples:

$$\mathbf{A} = \begin{pmatrix} \mathbf{a}^1 \sim \mathbb{R}^f \\ \vdots \\ \mathbf{a}^n \sim \mathbb{R}^f \end{pmatrix} = \begin{pmatrix} a_1^1 & \dots & a_f^1 \\ & \dots\dots & \\ a_1^n & \dots & a_f^n \end{pmatrix} \sim \mathbb{R}^{n\times f} \tag{47}$$

Without surprise, we denote the matrix product between $\mathbf{A} \sim \mathbb{R}^{n\times f}$ and $\mathbf{B} \sim \mathbb{R}^{f\times m}$ as $\mathbf{AB} \sim \mathbb{R}^{n\times m}$.

**Feature dot-product** This operation is defined as the dot-products between the feature vectors of a pair of matrices $\mathbf{A} \sim \mathbb{R}^{n\times f}$ and $\mathbf{B} \sim \mathbb{R}^{n\times f}$ as such:

$$\mathbf{A} \ominus \mathbf{B} = \begin{pmatrix} \mathbf{a}^1 \sim \mathbb{R}^f \\ \vdots \\ \mathbf{a}^n \sim \mathbb{R}^f \end{pmatrix} \ominus \begin{pmatrix} \mathbf{b}^1 \sim \mathbb{R}^f \\ \vdots \\ \mathbf{b}^n \sim \mathbb{R}^f \end{pmatrix} = \begin{pmatrix} \mathbf{a}^1 \cdot \mathbf{b}^1 \sim \mathbb{R} \\ \vdots \\ \mathbf{a}^n \cdot \mathbf{b}^n \sim \mathbb{R} \end{pmatrix} \sim \mathbb{R}^n \tag{48}$$

where we denote by $\mathbf{a} \cdot \mathbf{b} = \sum_{\mathfrak{f}=1}^{f} a_{\mathfrak{f}} b_{\mathfrak{f}}$ the regular dot-product between 1d vectors $\mathbf{a} \sim \mathbb{R}^f$ and $\mathbf{b} \sim \mathbb{R}^f$.

**Hadamard product** This operation takes a pair of matrices $\mathbf{A} \sim \mathbb{R}^{n\times f}$ and $\mathbf{B} \sim \mathbb{R}^{n\times f}$ and returns a new matrix $\sim \mathbb{R}^{n\times f}$ where each element is the product of elements of the original two matrices:

$$\mathbf{A} \circ \mathbf{B} = \begin{pmatrix} a_1^1 & \dots & a_f^1 \\ \vdots & \vdots & \vdots \\ a_1^n & \dots & a_f^n \end{pmatrix} \circ \begin{pmatrix} b_1^1 & \dots & b_f^1 \\ \vdots & \vdots & \vdots \\ b_1^n & \dots & b_f^n \end{pmatrix} = \begin{pmatrix} a_1^1 b_1^1 & \dots & a_f^1 b_f^1 \\ \vdots & \vdots & \vdots \\ a_1^n b_1^n & \dots & a_f^n b_f^n \end{pmatrix} \sim \mathbb{R}^{n\times f} \tag{49}$$

[9]Notice that the number of features is usually denoted $d$ for depth. We keep it here as $f$ for consistency with the description of 2d broadcasting of the bias vector and to emphasize that feature maps act as independent features.

**Frobenius product** This operation can be seen as a generalization of vector dot-products to matrices. Taking two matrices of the same size $\mathbf{A} \sim \mathbb{R}^{n\times f}$ and $\mathbf{B} \sim \mathbb{R}^{n\times f}$, the Frobenius product returns a single number $\sim \mathbb{R}$ defined as the sum of the entries of the Hadamard product:

$$\mathbf{A} \cdot \mathbf{B} = \sum_{ij} \mathbf{A} \circ \mathbf{B} = \sum_{s=1}^{n} \sum_{\mathfrak{f}=1}^{f} a_{\mathfrak{f}}^{s} b_{\mathfrak{f}}^{s} \tag{50}$$

Note that this product is also obviously related to the feature dot-product operation through:

$$\mathbf{A} \cdot \mathbf{B} = \sum_{\text{samples}} \mathbf{A} \ominus \mathbf{B} \tag{51}$$

Finally, let us mention some useful identities relating the binary matrix operations defined above with each other:

$$\mathbf{A} \cdot (\mathbf{B}\mathbf{C}) = \left(\mathbf{B}^t \mathbf{A}\right) \cdot \mathbf{C} = \left(\mathbf{A}\mathbf{C}^t\right) \cdot \mathbf{B} \tag{52}$$

$$\mathbf{A} \cdot (\mathbf{B} \circ \mathbf{C}) = (\mathbf{A} \circ \mathbf{B}) \cdot \mathbf{C} \tag{53}$$

# Appendix D Matrix derivatives

**Vector to scalar** As seen in the introduction and in section 2, neural networks involve a supervised loss function:

$$\ell\left(\mathbf{a}^s, \mathbf{y}^s\right) \sim \mathbb{R}$$

that takes a 1d feature vector $\mathbf{a}^s \sim \mathbb{R}^f$ corresponding to a sample $s$ and returns a scalar $\sim \mathbb{R}$ representing the amount of mismatch with a fixed ground-truth label $\mathbf{y}^s$. Typically, backpropagation proceeds by evaluating the total derivative of this loss function over a minibatch of $n$ samples:

$$\mathrm{d}\ell\left(\mathbf{A}, \mathbf{Y}\right) = \begin{pmatrix} \mathrm{d}\ell(\mathbf{a}^1, \mathbf{y}^1) \sim \mathbb{R} \\ \vdots \\ \mathrm{d}\ell(\mathbf{a}^n, \mathbf{y}^n) \sim \mathbb{R} \end{pmatrix} = \begin{pmatrix} \nabla^1 \ell(\mathbf{a}^1, \mathbf{y}^1) \cdot \mathrm{d}\mathbf{a}_1 \\ \vdots \\ \nabla^n \ell(\mathbf{a}^n, \mathbf{y}^n) \cdot \mathrm{d}\mathbf{a}_n \end{pmatrix} = \nabla \ell\left(\mathbf{A}, \mathbf{Y}\right) \ominus \mathrm{d}\mathbf{A} \sim \mathbb{R}^n \tag{54}$$

where the sample-specific $s \in \{1, \cdots, n\}$ nabla operator is defined as:

$$\nabla^s = \frac{\partial}{\partial \mathbf{a}^s} = \left( \frac{\partial}{\partial a_1^s}, \cdots, \frac{\partial}{\partial a_f^s} \right) \sim \mathbb{R}^f$$

Note that we are usually interested in the sensitivity of the loss with respect to the input data $\mathbf{A}$ and not to the ground-truth labels $\mathbf{Y}$. This explains why we considered only the feature vector dependence in the definition of the nabla operator.

**Scalar to scalar** As described in section 4, activation functions typically involve the application of a scalar to scalar function $a_{\mathfrak{f}}^s \sim \mathbb{R} \rightarrow g(a_{\mathfrak{f}}^s) \sim \mathbb{R}$ to all the elements of a data matrix:

$$g(\mathbf{A}) = \begin{pmatrix} g(a_1^1) & \ldots & g(a_f^1) \\ \vdots & \vdots & \vdots \\ g(a_1^n) & \ldots & g(a_f^n) \end{pmatrix} \sim \mathbb{R}^{n\times f}$$

Evaluating its total derivative yields:

$$\mathrm{d}g\left(\mathbf{A}\right) = g'(\mathbf{A}) \circ \mathrm{d}\mathbf{A} \sim \mathbb{R}^{n\times f} \tag{55}$$

# Contents